\pdfoutput=1 

\documentclass{article}

\usepackage{PRIMEarxiv}

\usepackage[utf8]{inputenc} 
\usepackage[T1]{fontenc}    
\usepackage{url}            
\usepackage{booktabs}       
\usepackage{amsfonts}       
\usepackage{nicefrac}       
\usepackage{microtype}      
\usepackage{lipsum}
\usepackage{fancyhdr}       
\usepackage{graphicx}       
\graphicspath{{media/}}     
\usepackage[comma, authoryear]{natbib}
\usepackage{amsmath, mleftright}
\usepackage{framed,multirow}
\usepackage{xparse}
\usepackage{algpseudocode}
\usepackage{mathtools}
\usepackage{graphicx}
\usepackage{caption}
\usepackage{subcaption}
\usepackage{ifthen, color}
\usepackage{lipsum}
\usepackage{booktabs}
\usepackage{tablefootnote}
\usepackage{array}
\usepackage{graphicx}
\usepackage{amssymb}
\usepackage{latexsym}
\usepackage{tablefootnote}
\usepackage{threeparttable}
\usepackage{url}
\usepackage[table]{xcolor}
\usepackage{bm}
\usepackage{tabularx}

\usepackage{hyperref}       

\definecolor{readablegreen}{HTML}{02ad0a}
\definecolor{newcolor}{rgb}{.8,.349,.1}

\DeclareMathOperator{\MS}{\mathcal{M}[\mathcal{S}]}
\DeclareMathOperator{\Ms}{\mathcal{M}_\mathcal{S}}
\newcommand{\M}[2]{\operatorname{\mathcal{M}}_{(#1)}^{(#2)_\star}}
\newcommand{\Mi}[1]{\operatorname{\mathcal{M}[\mathcal{B}}^{(#1)}]}
\newcommand{\D}[2]{\operatorname{\mathcal{D}}(#1, #2)}
\newcommand{\Dsingle}[1]{\operatorname{\mathcal{D}}(#1)}
\newcommand{\Btrain}[1]{\operatorname{\mathcal{B}}^{(#1)}}
\newcommand{\Btest}[1]{\operatorname{\mathcal{B}}^{(#1)}_\star}

\newcommand{\review}[1]{#1}

\definecolor{lightgray}{rgb}{.85,.85,.85}

\NewDocumentCommand{\evalat}{sO{\big}mm}{%
	\IfBooleanTF{#1}
	{\mleft. #3 \mright|_{#4}}
	{#3#2|_{#4}}%
}

\title{Multi-style conversion for semantic segmentation of lesions in fundus images by adversarial attacks
}

\author{
  Clément Playout, Renaud Duval, Marie Carole Boucher \\
  Centre Universitaire d’Ophtalmologie, \\
  Maisonneuve-Rosemont Hospital \\
  Montr\'eal, QC\\
  clement.playout.cemtl@ssss.gouv.qc.ca \\
   \And
  Farida Cheriet \\
  LIV4D, \\
  Polytechnic Montr\'eal \\
  Montr\'eal, QC\\
}

\begin{document}

\maketitle

\newboolean{revisionmode}
\setboolean{revisionmode}{false}

\newboolean{todomode}
\setboolean{todomode}{false}

\newcommand{\revisionFormat}[1]{%
	\ifthenelse{\boolean{revisionmode}}{\textcolor{blue}{\textbf{#1}}}{#1}}
\newcommand{\revisionRemove}[1]{%
	\ifthenelse{\boolean{revisionmode}}{}{}
}
\newcommand{\revisionAdd}[1]{%
	\ifthenelse{\boolean{revisionmode}}{
		\textcolor{readablegreen}{\textbf{#1}}}{#1}
}
\newcommand{\todo}[1]{%
	\ifthenelse{\boolean{todomode}}{\textcolor{orange}{\textbf{#1}}}{#1}}

\begin{abstract}
The diagnosis of diabetic retinopathy, which relies on fundus images, faces challenges in achieving transparency and interpretability when using a global classification approach. However, segmentation-based databases are significantly more expensive to acquire and combining them is often problematic. This paper introduces a novel method, termed adversarial style conversion, to address the lack of standardization in annotation styles across diverse databases. By training a single architecture on  combined databases, the model spontaneously modifies its segmentation style depending on the input, demonstrating the ability to convert among different labeling styles. The proposed methodology adds a linear probe to detect dataset origin based on encoder features and employs adversarial attacks to condition the model's segmentation style. Results indicate significant qualitative and quantitative gains
    through dataset combination, offering avenues for improved model generalization, uncertainty estimation and continuous interpolation between annotation styles. Our approach enables training a segmentation model with diverse databases while controlling and leveraging annotation styles for improved retinopathy diagnosis. 
    \end{abstract}


\section{Introduction}

The identification of anatomical and pathological markers visible in the fundus of the eye is the very first step toward its diagnosis. This observation holds particularly for diabetic retinopathy (DR), which is monitored longitudinally by characterising certain lesions. In contrast, for automatic diagnosis, many studies (\cite{fauwClinicallyApplicableDeep2018a, gulshanPerformanceDeepLearningAlgorithm2019, yangFundusDiseaseImage2021, guClassificationDiabeticRetinopathy2023}) choose a global approach that bypasses the explicit recognition of lesions. Although they achieve impressive performance, these approaches raise several issues frequently discussed in the literature as pointed by \cite{islamDeepLearningAlgorithms2020}. First and foremost, global classification lacks transparency and interpretability for the user (physician or patient), as the diagnosis is not supported by 
elements seen in the image that influenced the algorithm's decision. \revisionAdd{This has motivated others works on joint lesion segmentation and classification, such as the DeepDR system proposed by \cite{daiDeepLearningSystem2021} and recently extended by \cite{daiDeepLearningSystem2024} for prognosis. However, these approaches require a considerable amount of data.} Furthermore, the scale chosen for grading the disease relies on clinical standards that have been constructed according to precise rules for identifying lesions. However, these scales are not universal, and multiple systems coexist (ETDRS, ICDR, Scottish DR GS or Canadian Guideline among others \cite{sunUpdatingStagingSystem2021}), defining more or less compatible rules. These scales are not static and evolve based on clinical understanding of the disease and the imaging modalities (\cite{sunUpdatingStagingSystem2021, yangClassificationDiabeticRetinopathy2022}). These considerations justify the pursuit of research on semantic segmentation of retinal lesions in fundus images alongside the global approach. 

One of the main difficulties is obtaining sufficient annotations from qualified experts. To overcome this barrier, several teams have made their collected and annotated databases publicly available along with their models, thus promoting reproducibility and research in the field. However, despite the growing number of publicly accessible datasets, there is significant variability in the composition of these databases, both in terms of image quality and quantity, as well as the type of annotations provided. 
\revisionAdd{The acquisition itself may induce a distribution shift between different databases: indeed, fundus images can diverge due to differences of field-of-view, resolution, imaging procedure (mydriatic or not), camera type, etc. Some recent works suggest way of dealing with this image variability. \cite{liuTMMNetsTransferredMulti2023} propose a transfer-learning scheme to combine multiple modalities (wide field and regular fundus) to a common representation to diagnose rare retinal diseases. \cite{shenDomaininvariantInterpretableFundus2020} propose a semi-tied 
Adversarial Discriminative Domain Adaptation (ADDA, \cite{tzengAdversarialDiscriminativeDomain2017}) to obtain a domain-invariant quality assessing network. These approaches focus on misalignment in the distributions of images. 
For semantic segmentation, because of the absence of established guidelines, annotation protocols are often overlooked, which leads to very diverse annotation styles. This can be described as a distribution shift in the label space.}

Despite these considerations, research into lesions segmentation rarely addresses the issue of characterisation and comparison between databases. But their differences raise fundamental questions about interoperability: what does a model learn from databases with heterogeneous annotations? Can its behaviour be explicitly controlled? These questions echo, to some extent, the domain adaptation problem, from which we borrow certain ideas. But given that our segmentation work uses fundus images acquired under similar conditions regardless of the database considered, and that we restrict ourselves to a space of classes common to all databases, we prefer the notion of \emph{style conversion}: the same types of lesions will be labelled differently depending on the annotation protocol (which we conflate with the database itself). 
\\
Our work starts by training a single architecture on multiple combined databases, from which we highlight an unexpected 
result: when tested on the different databases' test sets, the model spontaneously converts its segmentation style to match the expected one and thus maximise its performance on a priori non-compatible labelling styles.\revisionAdd{This means that the network learns to recognize the origin of an image (in terms of database) and to adapt its prediction to match the expected style}. To better understand and harness this behaviour, 
we train a probe to identify each image's database using the encoder's features. Following this, our main contributions are based on two considerations:
\begin{itemize}
	\item The probe's ability to detect the image's origin based on the features maps extracted by the segmentation model's encoder \revisionAdd{and decoder}.
	\item The well-known effectiveness of adversarial attacks to fool a classifier into moving in a targeted direction.
\end{itemize}
We propose to use adversarial attack to modify an image toward the distribution of any given training database with a known labelling style. By doing so, we constrain the segmentation style of the model, which provides us with an effective multi-style conversion procedure, including the ability to continuously sample different segmentation hypotheses. 
Notably, our methodology works on any segmentation model based on neural networks and 
trained on multiple databases with a regular segmentation training procedure. The style conversion is done post-training by incorporating the probe, but this operation does not require modifying the segmentation model in any way.
We explore three applications of our method:
\begin{enumerate}
	\item Improving the performance of a model trained with multiples datasets, especially in the case where we only have a small fraction of finely labelled data.
    \item \revisionAdd{Refining a model's performance on an external (previously unseen) database by properly matching the expected style of the database per lesion.}
	\item Generating an uncertainty map for the segmentation produced by a model by sampling through multiples styles. To this end, we introduce the notion of continuous conversion between two styles.
\end{enumerate}

The rest of this paper is organised as follows: the next section situates our work within the existing literature. Section \ref{sec:metho} describes the different stages of our methodology: 
characterizing the different databases considered, constructing an efficient segmentation model, and introducing a formalism describing the proposed approach to condition the model to a specific style of annotations. The details of the experimental protocol are provided in Section \ref{sec:experiments}. 
Section \ref{sec:applications} presents two applications of our method to style distillation and uncertainty estimation. Finally, Sections \ref{sec:discussion} and \ref{sec:conclusion} provide a discussion and conclusion.

\section{Related works}

\subsection{Fundus Segmentation Architecture}
Research on lesion segmentation in the fundus of the eye has a rich history, significantly expanded in recent years. A substantial portion of this research is dedicated to designing new neural network architectures specifically tailored for lesion segmentation. Here, we focus on the most recent works related to multi-lesion segmentation of the four lesions introduced earlier. These architectures commonly emphasise the multi-scale aspect of the problem, as lesions vary greatly in size within an image and depending on their class. \cite{guoLSegEndtoendUnified2019} propose L-Seg, which is based on the multi-scale fusion of features extracted from a VGG network. A similar strategy is adopted by \cite{weiLearnSegmentRetinal2020} for their Lesion-Net, that also adds additional supervision through lesions classification and DR grading, the latter being also experimented in one of our initial work (\cite{playoutNovelWeaklySupervised2019a}).  \cite{heProgressiveMultiscaleConsistent2022} introduce PMCNet, building on the idea of the UNet by \cite{ronnebergerUNetConvolutionalNetworks2015a} but modifying the skipped connections to incorporate multi-scale feature fusion from adjacent encoder layers. A modified UNet is also experimented by \cite{xuFFUNetFeatureFusion2021}.
On the other hand, \cite{yanLearningMutuallyLocalGlobal2019} (Global-Local UNet) and \cite{guoCARNetCascadeAttentive2022a} (CARNet) adopt a different strategy, focusing on the fusion of features extracted at a global scale (entire image at lower resolution) and a local scale (patches of the image extracted at higher resolution).
Designing a novel architecture tackling the specificity of our task is sounded, but in practice, it often hampers reproducibility. The availability of source code is still limited and the complexity of some architectures makes their unambiguous implementation challenging. 
To broaden the spectrum of our results and for the sake of transparency, we have re-implemented and retrained a few of the previously mentioned CNNs as well as more generic ones. The code we built is released as an open-source library alongside this paper.

\subsection{Multi-style conversion}
The conversion to different style of segmentation is a notion rarely covered in the literature, whether for retinal images or other applications. However, it is thematically closely related to the much more covered field of uncertainty assessment, as it also involves predicting multiple plausible segmentation hypotheses from one image. \revisionAdd{The pioneering work of \cite{galDropoutBayesianApproximation2016} introduced an innovative approach to uncertainty estimation in deep models.} It reinterprets Dropout as a Bayesian process over the state of all possible models. Concretely, the network's inner connection are randomly dropped at inference time, the final prediction being obtained by averaging multiple forward passes following a Monte-Carlo-like sampling. To our knowledge, \cite{garifullinDeepBayesianBaseline2021}'s work is the only one aiming at modelling the aleatoric uncertainty in fundus retinal lesions segmentation and it is built upon this latter approach. \revisionFormat{We take inspiration from their work to suggest a similar generation of uncertainty maps from multiple samples.}\\
In style conversion, the hypotheses correspond to various ways of labelling the images, not necessarily due to the uncertainty around the lesion's structure but rather cause by the diversity of annotation protocol proper to each dataset. This observation, at the core of our experiments, has also motivated a recent paper by \cite{zepf2023label}, which distinguishes uncertainties from the style specific to each annotator. In their methodology, the style is explicitly embedded as an input of the prior network and conditions a latent space distribution. \revisionAdd{Their work expands on a rich literature on noisy labels for medical image segmentation  motivated by the difficulty of acquiring (or even defining) an universal groundtruth for many tasks in this field (\cite{kohlProbabilisticUnetSegmentation2018, Kohl2019AHP, Bhat2023EffectOL, qiu2021modal, monteiroStochasticSegmentationNetworks2020}). The Probabilistic U-net by \cite{kohlProbabilisticUnetSegmentation2018} is recognised as an important milestone for the segmentation of ambiguous structures. It integrates the conditional variational autoencoder paradigm with a U-net by broadcasting a latent variable sampled from a learned Gaussian distribution inside the last stage of the decoder. The latent space encompasses the diversity of plausible segmentations given the input image and the annotator's manual labelling. \cite{Kohl2019AHP} extends their previous work by using multiple distributions and integrating different sampled latent variables (one for each distribution) at every steps of the decoder, thereby controlling the hypotheses at different resolutions. More recent papers have explored more complex distributional spaces (Gaussian Mixture by \cite{Bhat2023EffectOL} or discrete variable by \cite{qiu2021modal}).}
\\
\subsection{From Adversarial Domain Adaptation to Conversion}
\revisionAdd{In contrast to these works, our approach does not explicitly model the style distribution. We share the objective of generating multiple segmentation hypotheses from a single model, but we rely on the model's ability to implicitly learn different styles. We introduce a post-training method to manipulate the input images in a way that induces a bias toward a predefined learned style. This approach aligns closely with the field of adversarial domain adaptation.
In adversarial domain adaptation, the typical approach involves a min-max game between a generator and a discriminator. The generator is trained to match a source distribution to a target one, while the discriminator detects the distribution shift in the generator's output. Numerous applications based on this general idea exist, including those involving fundus images. For example, \cite{caoCollaborativeLearningWeaklysupervised2022} uses a Cycle-GAN to improve DR classification performance by combining weak and strong supervision, while \cite{kadambiWGANDomainAdaptation2020, zhouUnsupervisedDomainAdaptation2024} incorporates a Wasserstein-GAN into their architecture to minimize the domain shift between different databases, achieving domain-independent semantic segmentation of the optic disc and cup.
Contrary to these approaches, we do not train a generator, having observed that the regular segmentation model already behaves like one. Instead, we propose to modify the image using adversarial attacks \cite{szegedyIntriguingPropertiesNeural2014}. Adversarial attacks are less commonly applied to segmentation than to classification, due to the unique challenge posed by the large combinatorial space of outcomes (each pixel being a classification problem in itself). Works such as \cite{ronyProximalSplittingAdversarial2023, croceRobustSemanticSegmentation2023} have addressed these challenges, but we adopt a simpler approach by building a proxy linear classification model as the basis of our attack.}
\\
Our methodology follows from an initially counter-intuitive observation: after being trained on multiple datasets simultaneously, a model tends to adopt one style conditionally to the input image. In other words, the image's appearance betrays its origin; \revisionAdd{and since each database is characterized by a labelling style, the network matches the corresponding style to maximize its performance. The tendency of a segmentation model to be very sensitive to biased errors in annotations 
has been observed before by \cite{vorontsovLabelNoiseSegmentation2021}, although not specifically for retinal lesions. They conclude that it is a problem to be mitigated during training, whereas we take advantage of it in a post-training step.} 
Indeed, further analysis of this property has led to a simple but theoretically grounded method to manipulate a model toward a specific style, \revisionAdd{which generalizes to images and databases never seen by the network during training}. As a result, we can sample multiple stylised segmentations from a single conventional model.

\section{Methodology and material}
\label{sec:metho}

\subsection{Clinical elements}
Our clinical framework focuses on four types of lesions, which are the most common manifestations of the first stages of diabetic retinopathy. \textit{Microaneurysms (MA)} are small dilations of the capillaries appearing in very early stages of the disease. Among other causes, the rupture of a microaneurysm can cause a blood leakage, which can take many different shapes (dot, flame-like, pre-retinal, vitreous...) We refer to these as \textit{Hemorrhages (HEM)} indiscriminately. The leakage from damage capillaries can also cause lipoprotein exudations called \textit{Exudates (EX)} that appear as bright lesions with well defined contours. Conversely \textit{Cotton Wool Spots (CWS)}, corresponding to an accumulation of axoplasmic material, tend to have blurrier borders.

\subsection{Datasets characterisation}
\begin{figure}[h]
	\centering
	\includegraphics[width=.5\columnwidth]{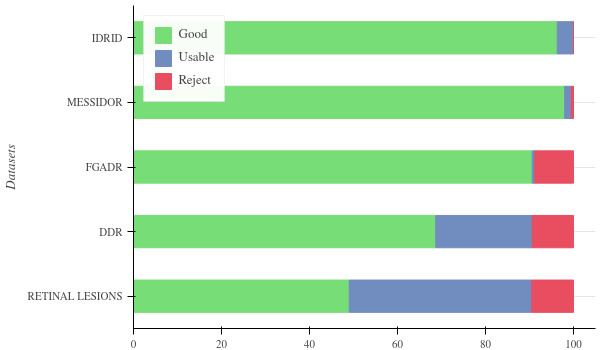}
	\caption{Classification of the images in each dataset into three quality levels, as assessed using MCF-Net.}
	\label{fig:qualityProportion}
\end{figure}

Five distinct and publicly available databases are used throughout our study \revisionAdd{for training and validation. Each one is split into three sets (train, validation and test). We also use a sixth database named TJ-DR, recently introduced by \cite{mao2023tjdr}, for external validation only (this database is never used for training purposes).}
Table \ref{tab:DatasetsCharacteristics} summarises the characteristics of the data we collected, \revisionAdd{and briefly describes the labelling procedures when known. For more details, we refer to the original papers, as the labelling procedures vary greatly between sources.} It should be noted that the heterogeneity of the databases arises from two sources: the images $\mathbf{X}$ on one hand, and the style of the annotations $\mathbf{Y}$ on the other. Characterising the differences between databases within these two distribution spaces is not straightforward. For the images, we restrict our comparison to the quality of the acquisitions. We use the Multiple Color-space Fusion Network (MCF-Net) developed by \cite{fuEvaluationRetinalImage2019} to classify the images into three classes: Good, Usable, Reject (Figure \ref{fig:qualityProportion}). Regarding the annotation style, we characterise it by a pair of variables $(S, Q)$ 
representing the average size and number of annotated structures per image and lesion category. Figure \ref{fig:AnnotationsDistribution} depicts the distributions 
obtained with Kernel Density Estimation for the five databases.

\begin{table}
		\centering
		\begin{threeparttable}
			
		\begin{tabular}{lcccr}
			\toprule
			Dataset & Train & Test & Resolution & \revisionAdd{\# labellers} \\
			\midrule
			IDRiD\tnote{1} & 54 & 27* & 2848x4288 & \revisionAdd{3}\\
			
			MESSIDOR\tnote{2} & 140 & 60* & 1500x1500 & \revisionAdd{7} \\
			
			DDR\tnote{3} & 383+149* & 225* & 1934x1956 & \revisionAdd{6} \\
			
			RET-LES\tnote{4} & 1115 & 478 & 896x896 & \revisionAdd{45} \\
			
			FGADR\tnote{5} & 1290 & 552 & 1280x1280  & \revisionAdd{3}\\

            \midrule
             &  \revisionAdd{Val} & \revisionAdd{Test} & \\
             \midrule
             \revisionAdd{TJ-DR\tnote{6}} & \revisionAdd{448} & \revisionAdd{113} & \revisionAdd{2048x2048} & \revisionAdd{3}\\
			\bottomrule
		\end{tabular}
	\begin{tablenotes}
\footnotesize 
		\item[1] \cite{porwalIDRiDDiabeticRetinopathy2020}; \revisionAdd{one Masters student labelling, revised by two ophthalmologists.}
		\item[2]  \cite{decenciereFEEDBACKPUBLICLYDISTRIBUTED2014a}, \revisionAdd{\cite{lepetit-aimonMAPLESDRMESSIDORAnatomical2024}; one ophthalmologist per biomarker (lesion, anatomical) type.}
		\item[3] \cite{liDiagnosticAssessmentDeep2019}
		\item[4] \cite{weiLearnSegmentRetinal2021}; \revisionAdd{three ophthalmologists per image.}
		\item[5] \cite{zhouBenchmarkStudyingDiabetic2021}; \revisionAdd{two resident ophthalmologists and one physician in charge of revision.}
        \item[6] \cite{mao2023tjdr}; \revisionAdd{three ophthalmologists per image.}
	\end{tablenotes}
\caption{The \review{six} databases used in this study. DDR provides an explicit validation set; for the others, we extract 15\% of the train set for this purpose. Asterisks indicate that the test split was made by the database's authors. Otherwise, we randomly sample 30\% of the whole dataset for the test set.}
\label{tab:DatasetsCharacteristics}
\end{threeparttable}
\end{table}

\def\figSizes{.49\textwidth}
\begin{figure*}[t]
	\centering
	\begin{subfigure}{\figSizes}
		\includegraphics[width=\columnwidth]{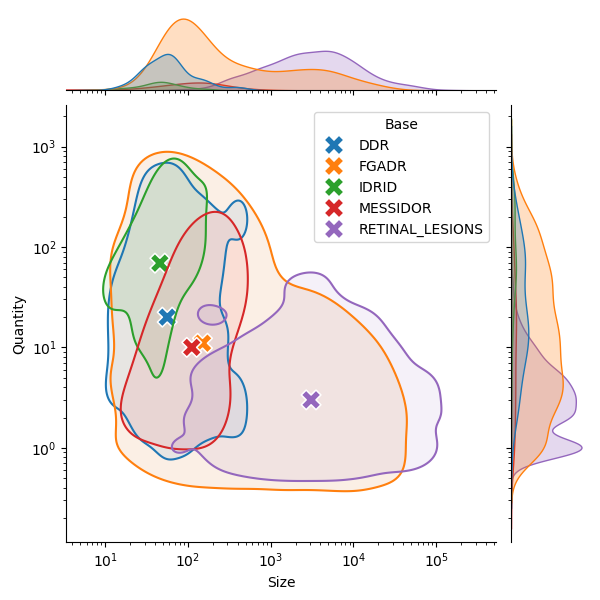}
		\caption{Exudates}
	\end{subfigure}
	\begin{subfigure}{\figSizes}
		\includegraphics[width=\columnwidth]{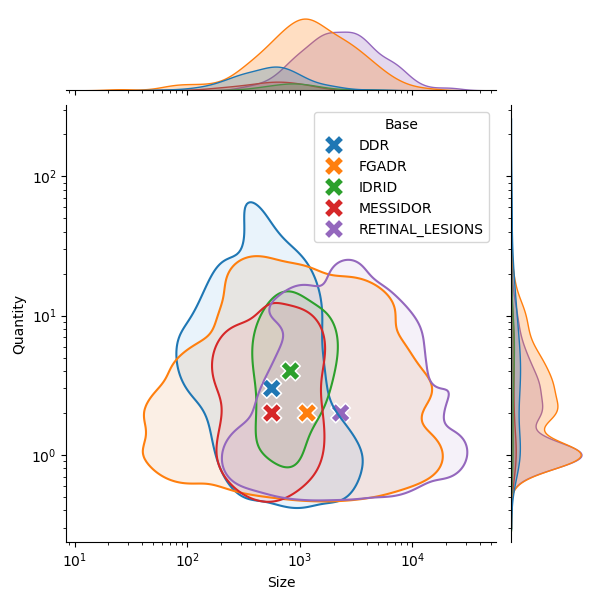}
		\caption{Cotton Wool Spots}
	\end{subfigure}
	\begin{subfigure}{\figSizes}
		\includegraphics[width=\columnwidth]{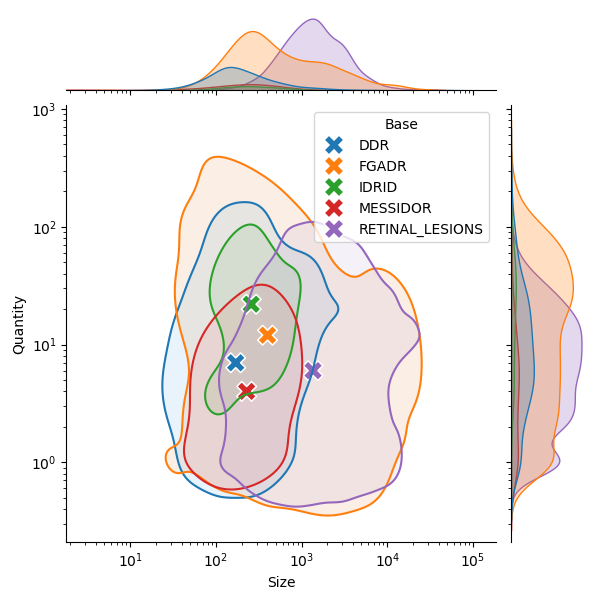}
		\caption{Hemorrhages}
	\end{subfigure}
	\begin{subfigure}{\figSizes}
		\includegraphics[width=\columnwidth]{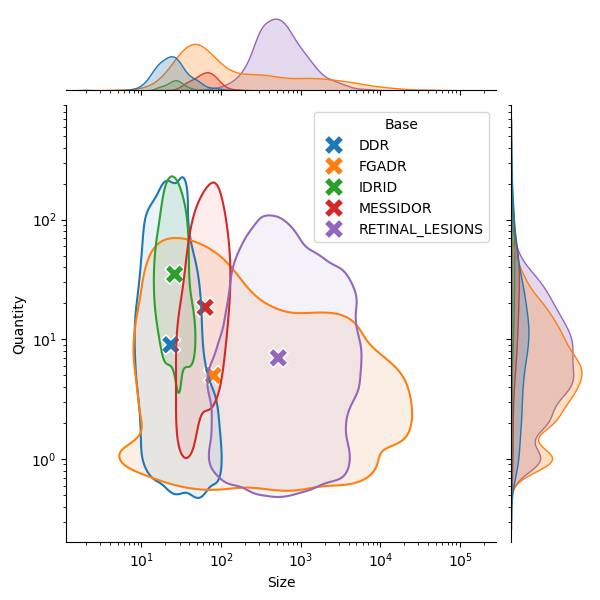}
		\caption{Microaneurysms}
	\end{subfigure}
\caption{Distributions $P^{(i)}(S, Q)$ for each lesion type for the five datasets. The crosses indicates the centroids of each dataset. Note that we use logarithmic scale to fit the distributions on a single graph: several orders of magnitude separate some centroids.}
\label{fig:AnnotationsDistribution}
\end{figure*}

\subsection{Segmentation models}
Our methodology focuses primarily on the interaction between various databases with heterogeneous annotations. In that light, the choice of a particular segmentation architecture is secondary. However, considering the limitations highlighted in our literature review, we have undertaken to provide a standardized re-implementation of several models (specific to retinal lesions or not), 
accessible as a python package structured in the form of a ``model zoo''.
Whenever possible, we have adhered closely to the instructions from the original papers (or the official implementations when available). However, some setups may marginally differ from their authors' original studies (image resolution, data augmentation policy, batch size, number of epochs).

Several standard models (using the implementations provided by \cite{Iakubovskii:2019}) are also trained. The choices of architecture and the training details are reported in Section \ref{sec:TrainingDetails}.

As a segmentation performance metric, it is common practise in the field to use the Area Under the Precision/Recall Curve (AUC), following a convention chosen by the IDRiD competition's organizers (\cite{porwalIDRiDDiabeticRetinopathy2020}).
However, the AUC suffers from being a class-wise metric. To summarise the models' performance globally, we adopt the mean-Intersection-Over-Union (mIoU), which is also widely used in many semantic segmentation tasks. 

\subsection{Training details}
\label{sec:TrainingDetails}
To train the segmentation model, we conducted a Bayesian hyper-parameters tuning over 50 runs by training on the smallest dataset (IDRiD) while monitoring on all datasets' validation sets combined. 
The search space included the cost function (Cross-Entropy with or without balancing and Dice), coefficient for label-smoothing, weight decay, learning rate, optimizer (Stochastic Gradient Descent, Adam, AdamW) and data augmentation regime among three pre-defined configurations: 

\begin{itemize}
	\item \textit{light}: random horizontal flipping, scaling, shifting and rotation;
	\item \textit{medium}: \textit{light} + random vertical flipping and brightness/contrast changes;
	\item \textit{heavy}: \textit{medium} + random gamma transforms and Gaussian blurring.
\end{itemize}

This search converged on using a Dice loss, with a smoothing factor of 0.4, a learning rate of $3 \times 10^{-3}$ and a weight decay of $10^{-5}$. 
These hyperparameters were kept for subsequent training, including for other architectures. 
Regarding the image resolution, we tested a resizing of $1024 \times 1024$ and $1536 \times 1536$. The latter provided a boost in raw performance but was significantly more taxing on hardware resources. Since we found that the results regarding style conversion held for both resolutions, all figures and results presented in this paper were done at $1024 \times 1024$. In either case, the different training runs were done on random crops of the images with a size of $512 \times 512$. The batch size was set to 32 accordingly to the GPU memory at our disposal (48Go on a Nvidia RTX A6000). 

\subsection{Style conversion}

\subsubsection{Notations}
In the interest of clarity, we introduce a set of notations that will be used throughout the rest of the paper. Each train (respectively test) set is referenced as $\mathcal{B}^{(i)}$ (resp. $\mathcal{B}^{(i)}_\star$), where $i$ spans across the set of databases by their initials, i.e $i \in \{I, M, D, R, F\}$. An architecture trained on $\mathcal{B}^{(i)}$ and tested on $\mathcal{B}^{(j)}_\star$ is noted $\mathcal{M}[\mathcal{B}^{(i)}](\mathcal{B}^{(j)}_\star)$ or simply $\mathcal{M}_{(i)}^{(j)^\star}$. It can also be trained on multiple databases $\mathcal{M}[\bigcup_i \mathcal{B}^{(i)}]$. In particular, we note $\mathcal{S} = \bigcup_i^{\{I, M, D, R, F\}} \mathcal{B}^{(i)}$ the union of all the datasets, $\MS$ being the architecture trained on all the training images available. The performance of a model is assessed by similarity score between a prediction and a reference. Most of the time, the latter consists of the annotation of the testing set considered, in which case the similarity measure is written as $\mathcal{D}(\mathcal{M}_{(i)}^{(j)^\star}, \mathcal{B}^{(j)}_\star)$. We also measure the similarity between a pair of models' predictions using a similar notation, $\mathcal{D}(\mathcal{M}_{(m)}^{(j)^\star}, \mathcal{M}_{(n)}^{(j)^\star})$. In Section \ref{sec:AdversarialAttack}, we describe an approach to explicitly modify a model's prediction style so that it adopts the one corresponding to a targeted database.
The modification occurs on the data fed at inference time rather than on the trained model itself. Recall that 
we equate the notion of annotation style with the characteristics proper to each database. The conversion process is marked by an arrow ($\rightarrow$), such that $\mathcal{M}(\mathcal{B}^{(j)}_\star \rightarrow \mathcal{B}^{(T)})$ (or simply $\mathcal{M}(\mathcal{B}^{(j)}_\star \rightarrow T$) represents the prediction of model $\mathcal{M}$ on dataset $j$ that has been modified so that $\mathcal{M}$ adopts the labelling style corresponding to dataset $T$. \revisionAdd{We name this process ``semantic style conversion" as it represents our intended purpose; but in practise, the modification itself is done on the image}. 

\subsubsection{Cross-dataset evaluation}
We investigate the performance obtained by $\Mi{i}$ when tested on $\Btest{j}$ $\forall (i, j) \in \{I, M, D, R, F, \mathcal{S}\}\times \{I, M, D, R, F\}$. This is summarised in matrix form in Table \ref{tab:crossDatasetResults}.

\begin{table}
\centering
	\begin{tabular}{l|ccccc|c}
		\toprule
		Model & $\Btest{I}$ &  $\Btest{M}$ & $\Btest{D}$ & $\Btest{R}$ & $\Btest{F}$ & Average \\
		\midrule
		$\Mi{I}$ & \cellcolor{lightgray}0.555 & 0.375 & 0.339 & 0.247 & 0.298 & 0.330 \\

		$\Mi{M}$ & 0.398 & \cellcolor{lightgray}0.467 & 0.306 & 0.272 & 0.276 & 0.324 \\

		$\Mi{D}$ & 0.520 & 0.353 & \cellcolor{lightgray}0.423 & 0.256 & 0.310 & 0.373 \\

		$\Mi{R}$ & 0.294 & 0.290 & 0.263 & \cellcolor{lightgray}0.480 & 0.292 & 0.344 \\

		$\Mi{F}$ & 0.354 & 0.280 & 0.313 & 0.246 & \cellcolor{lightgray}0.458 & 0.363 \\
		\hline
		\midrule
		
		$\MS$ & \textbf{0.581} & 0.436 & \textbf{0.433} & \textbf{0.496} & \textbf{0.465} & \textbf{0.482} \\
		\bottomrule
	\end{tabular}
\caption{$\text{mIoU}(\mathcal{M}_{(i)}^{(j)^\star}, \mathcal{B}^{(j)}_\star)$ scores computed on the different test sets from the predictions obtained with the same architecture (UNet with a ResNet encoder) trained on the different train sets.}
\label{tab:crossDatasetResults}
\end{table}

The first five rows pertain to models that we identify as ``specialised''. Having been trained on only one database (and therefore a single style), they tend to adopt the style of that particular database, thereby maximising their performance on the corresponding test set. This explains the matrix's diagonal predominance in $\text{mIoU}(\M{i}{j}, \Btest{j})$. It is noteworthy that, on average, all models tend to behave relatively similarly (last column).A column-wise reading of this matrix is also useful: it can serve as a proxy for the similarity between datasets. Expanding on this idea, the standard deviation column-wise provides a compatibility measure between datasets. As reported in Table \ref{tab:StandardDeviation}, it tends to confirm that IDRID and RETINAL-LESIONS are the least compatible with (or the most different from) the other datasets.

\begin{table}
	\centering
	\begin{tabular}{lccccc}
		\toprule
		 & $\Btest{I}$ &  $\Btest{M}$ & $\Btest{D}$ & $\Btest{R}$ & $\Btest{F}$ \\
		\midrule
		$\sigma_i(\mathcal{D}(\Mi{i}))$ & 0.118 &	0.076 & 0.068 &	0.120	& 0.087
		 \\
		\bottomrule
	\end{tabular}
\caption{Standard deviations of the scores obtained by different models $\Mi{i}$ (taken column-wise from Table \ref{tab:crossDatasetResults}).}
\label{tab:StandardDeviation}
\end{table}

\subsubsection{Source identification by feature probing}

\def\imgSizes{.19\textwidth}
\begin{table*}[t]
	\centering
	\setlength\tabcolsep{0pt} 
	\renewcommand{\arraystretch}{0} 
	\begin{tabular}{m{1cm} >{\centering\arraybackslash} m{\imgSizes} >{\centering\arraybackslash} m{\imgSizes} >{\centering\arraybackslash} m{\imgSizes} >{\centering\arraybackslash} m{\imgSizes} >{\centering\arraybackslash} m{\imgSizes} r}

		Model & $\Btest{I}$ &  $\Btest{M}$ & $\Btest{D}$ & $\Btest{R}$ & $\Btest{F}$ & \\
		\midrule
		$\Mi{I}$ 
		& \includegraphics[width=\imgSizes]{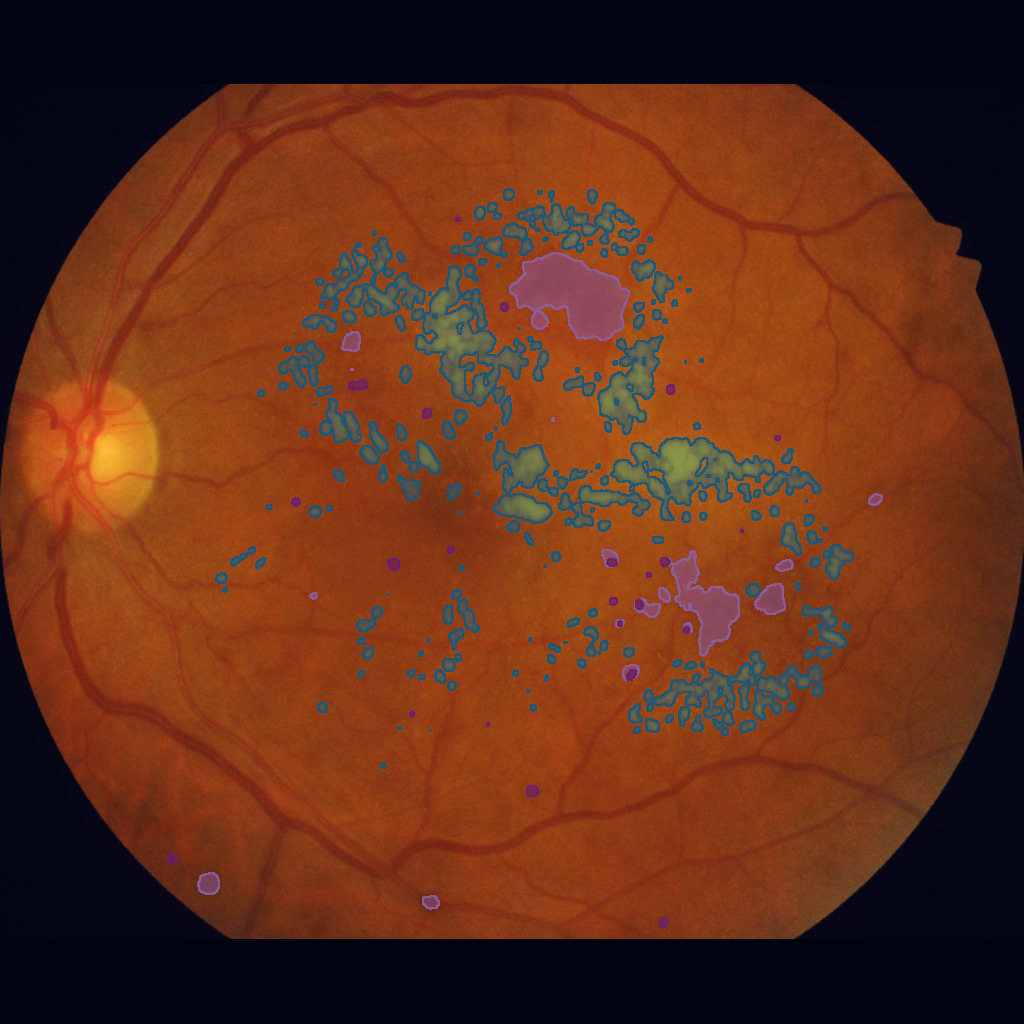} 
		& \includegraphics[width=\imgSizes]{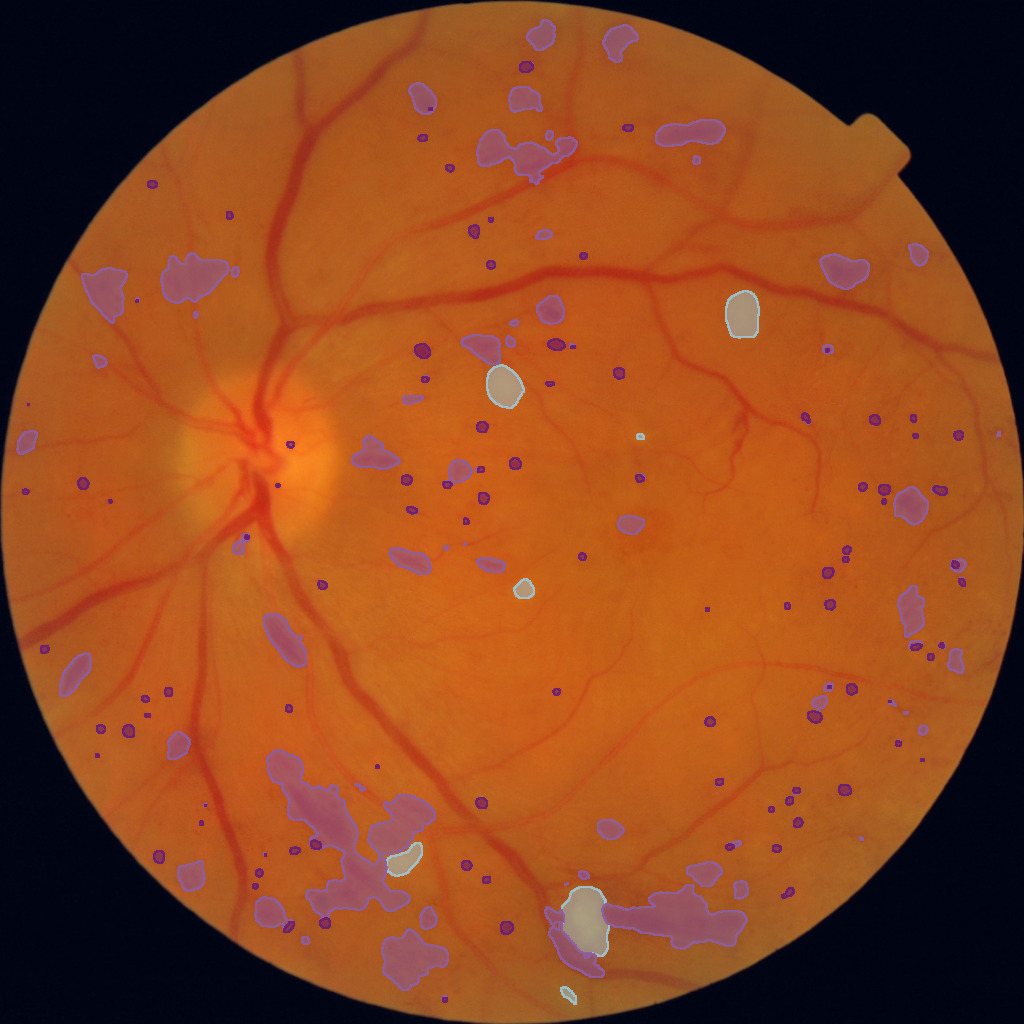} 
		& \includegraphics[width=\imgSizes]{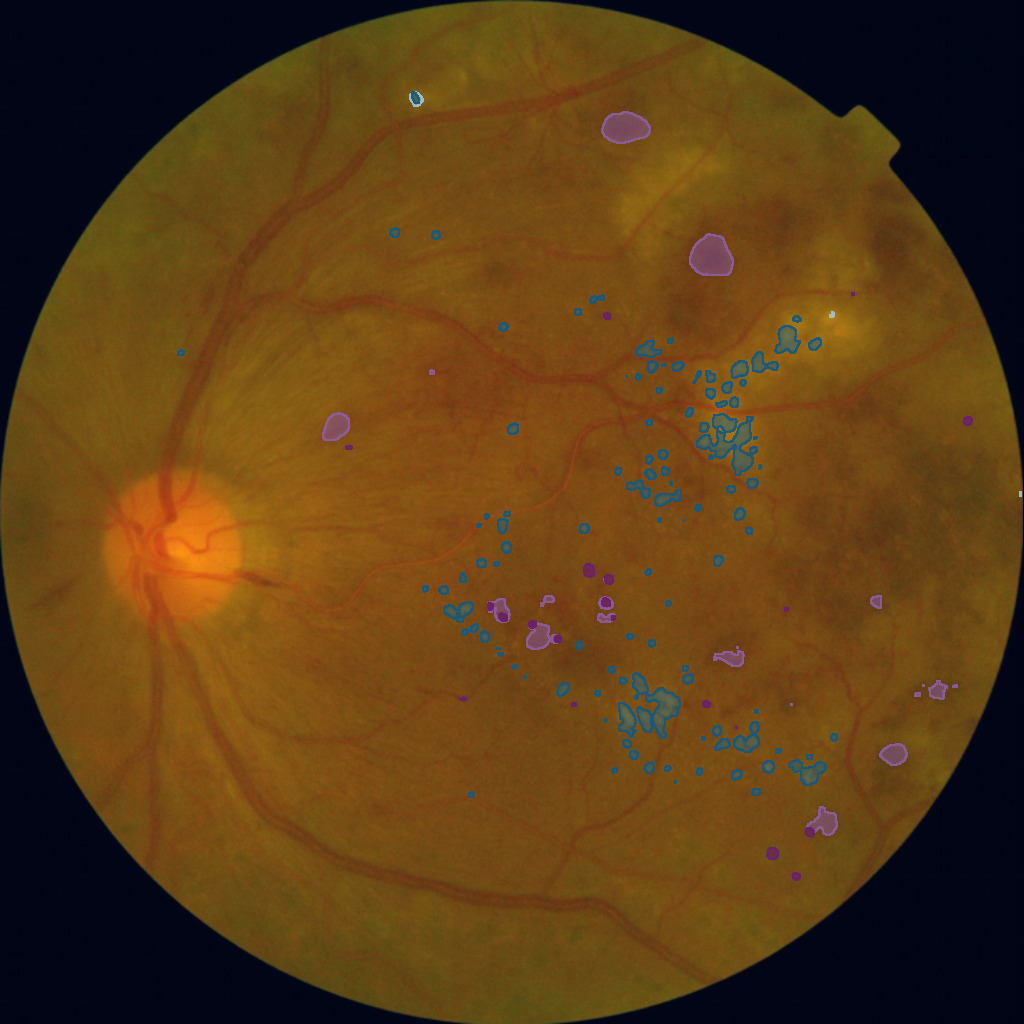} 
		& \includegraphics[width=\imgSizes]{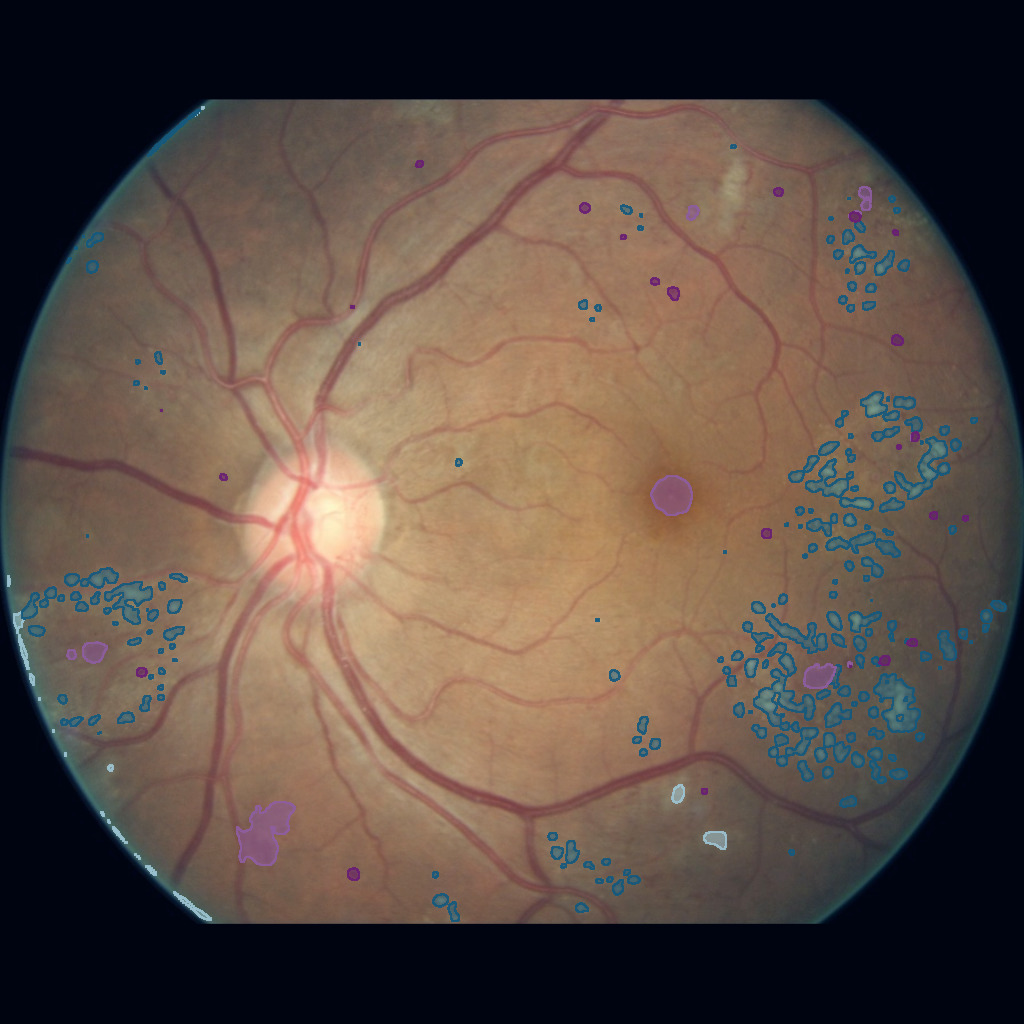} 
		& \includegraphics[width=\imgSizes]{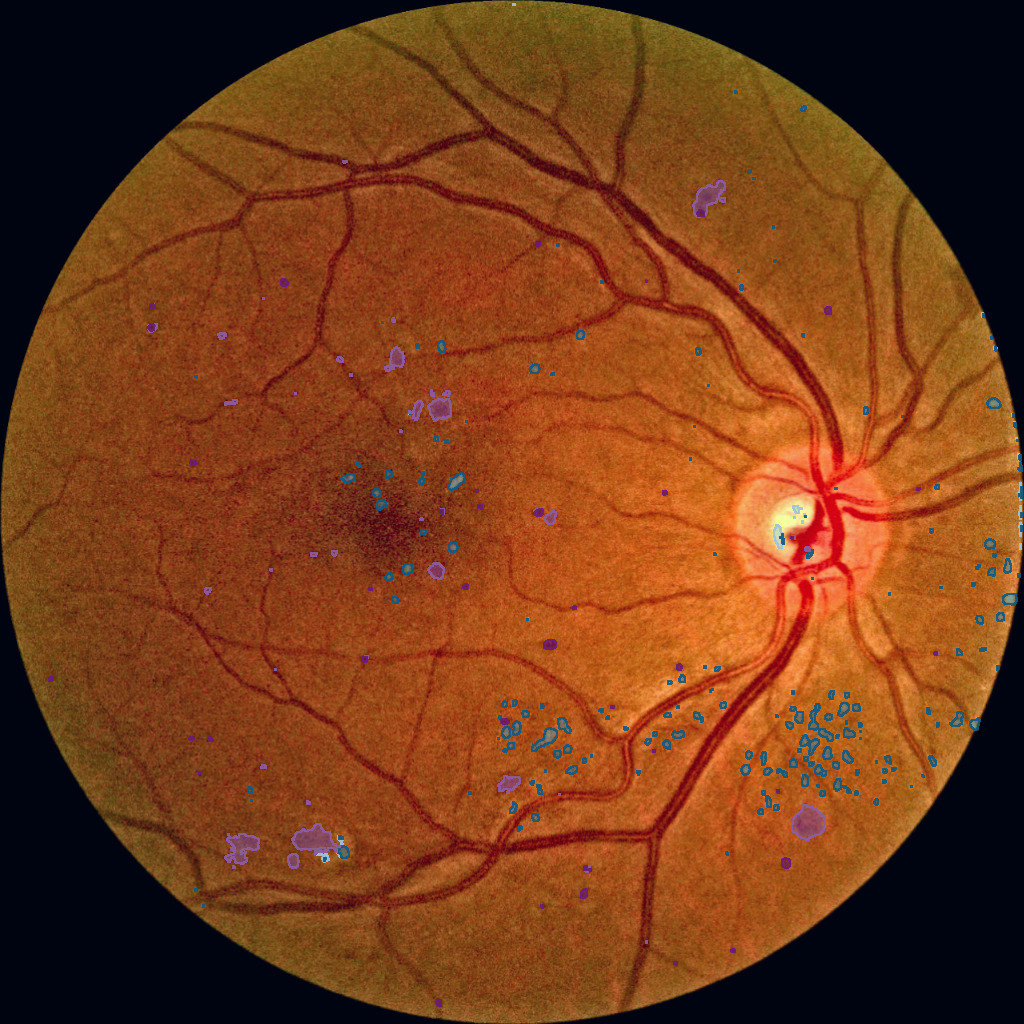} 
		& \multirow{3}{*}[.69in]{\includegraphics[width=.551cm]{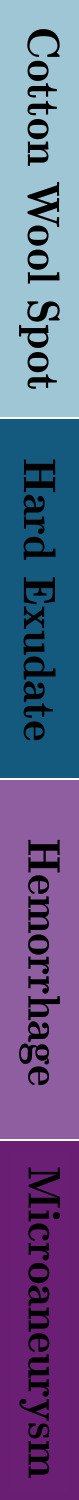}}
		\\
		$\Mi{R}$ 
		& \includegraphics[width=\imgSizes]{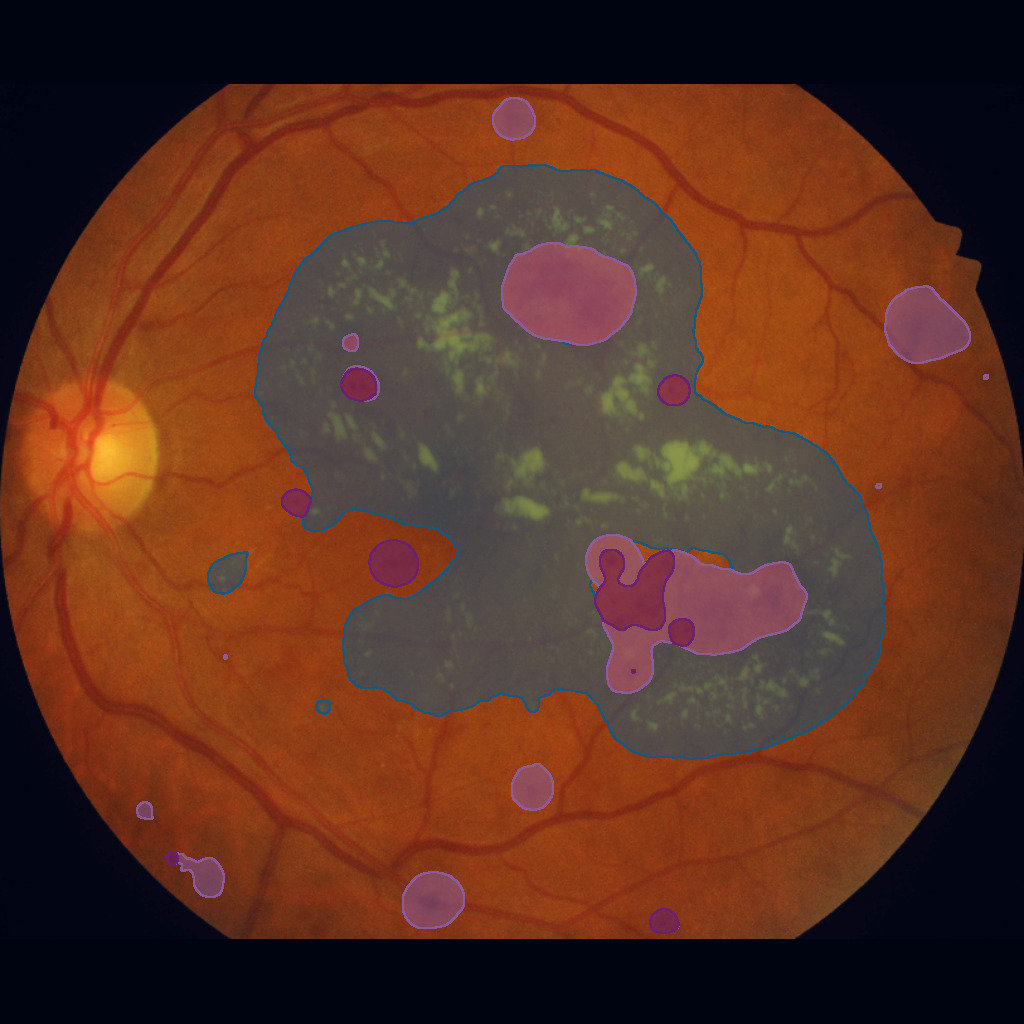} 
		& \includegraphics[width=\imgSizes]{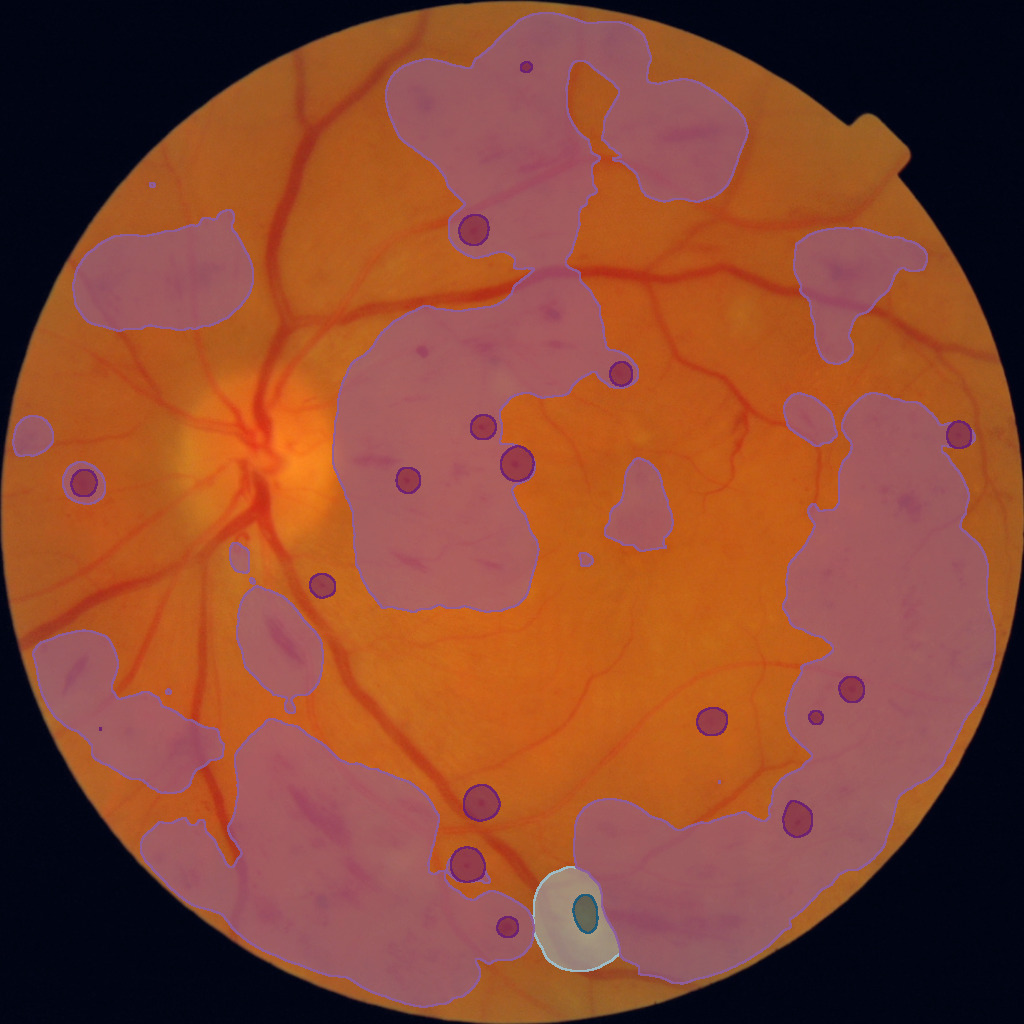} 
		& \includegraphics[width=\imgSizes]{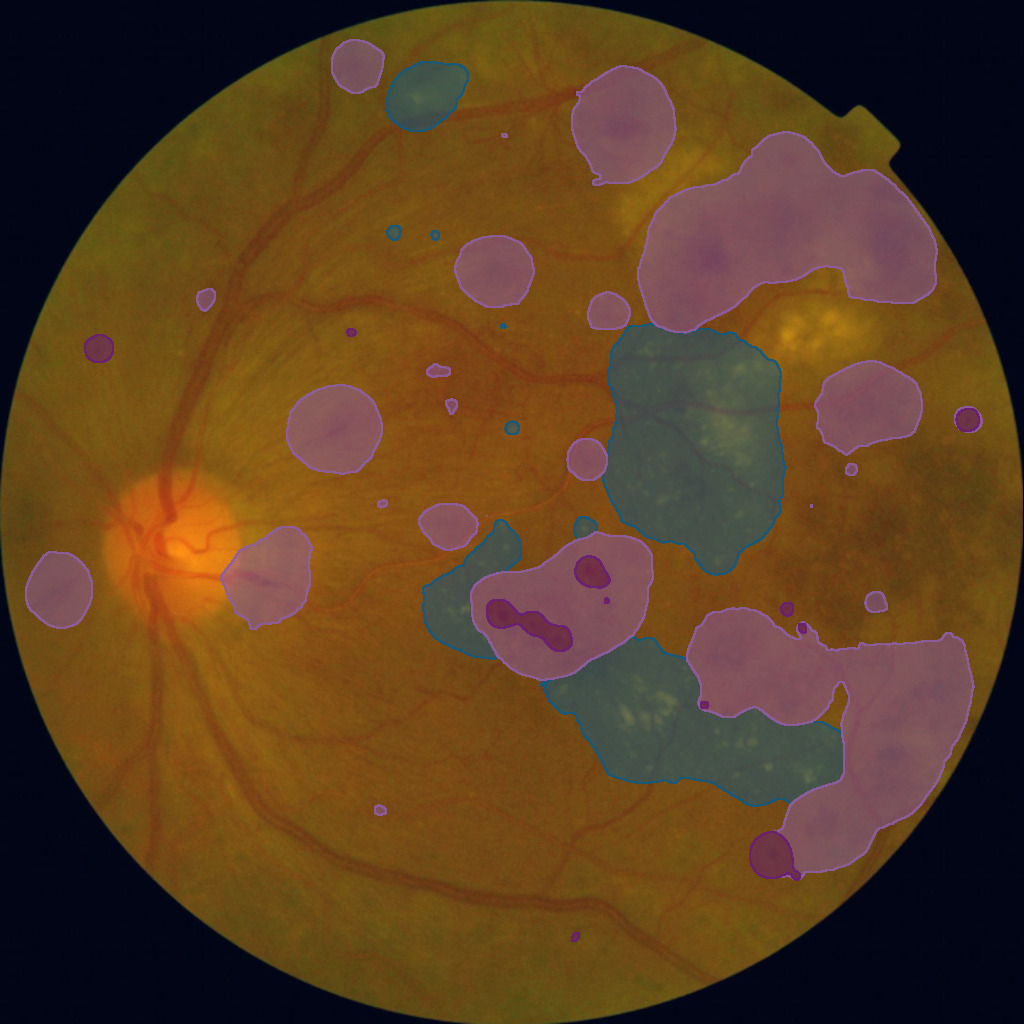} 
		& \includegraphics[width=\imgSizes]{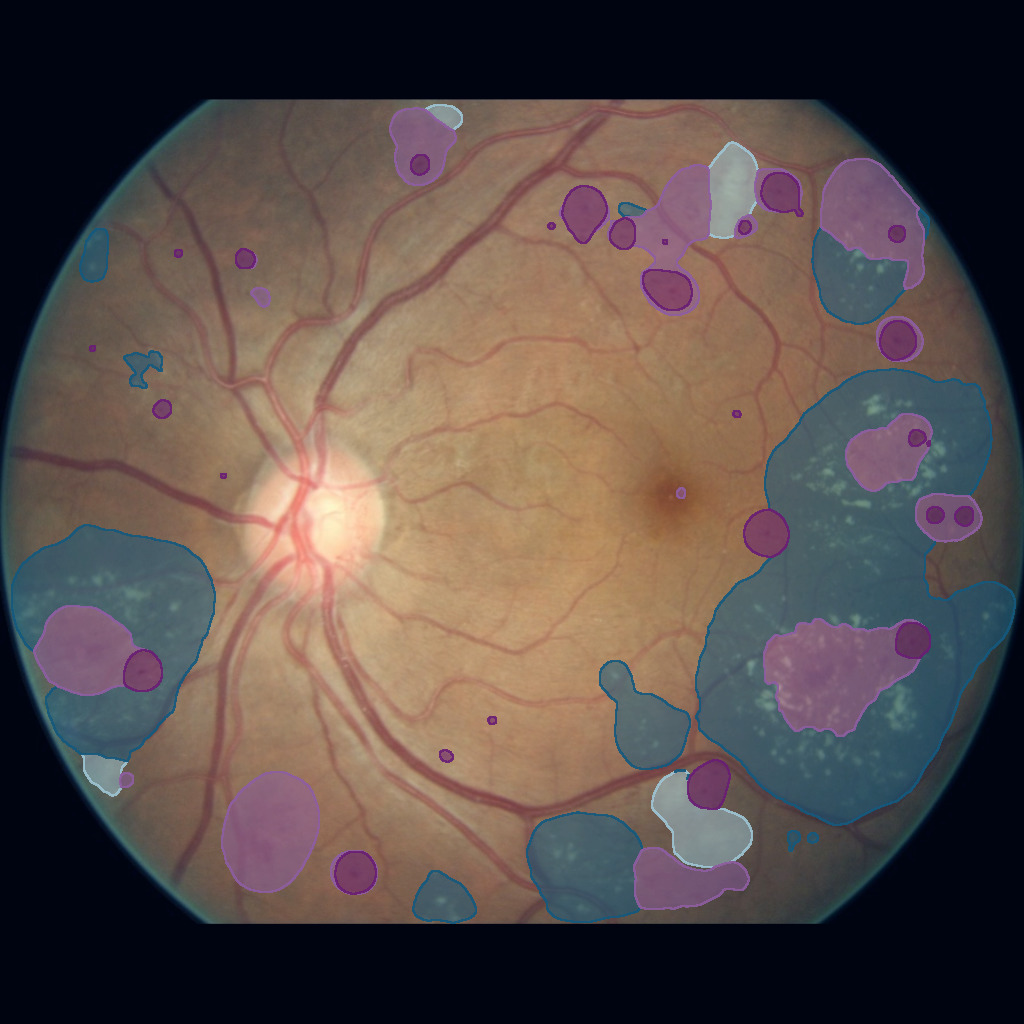} 
		& \includegraphics[width=\imgSizes]{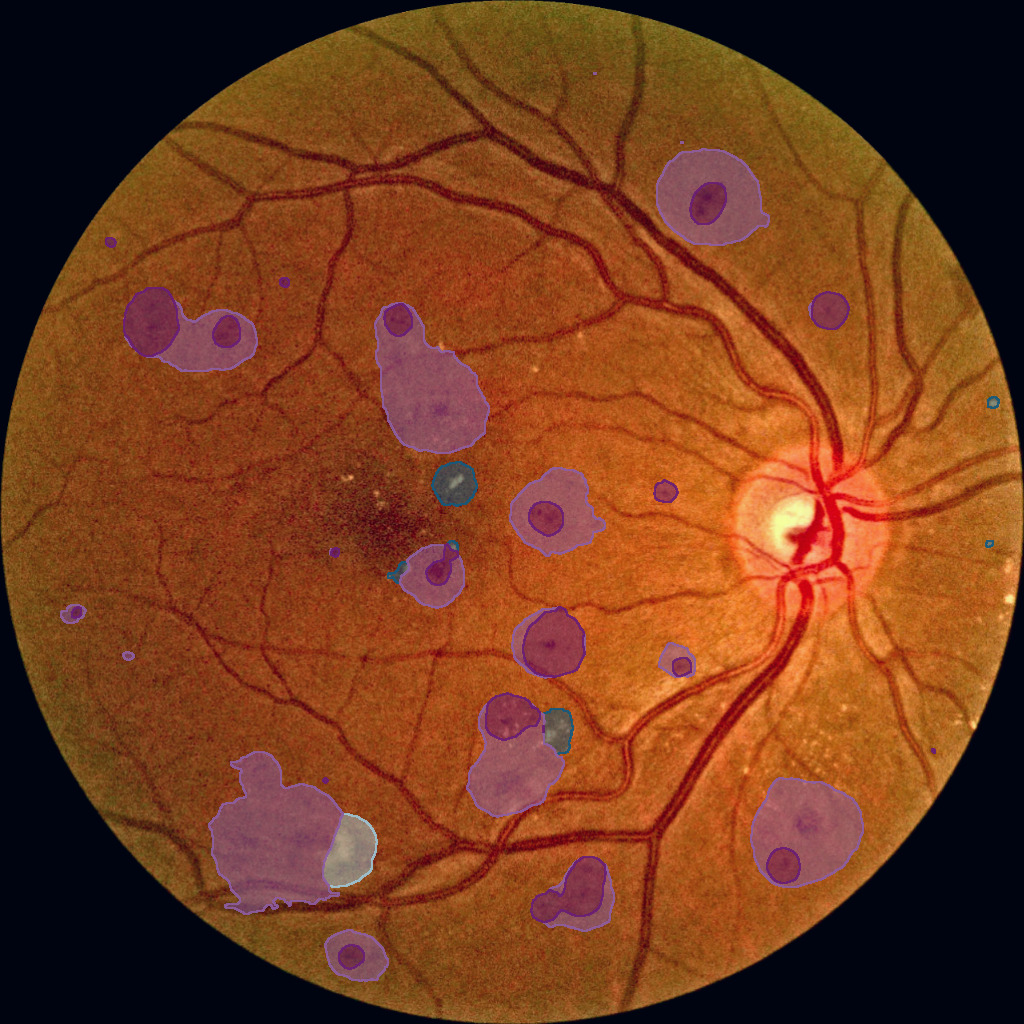} & \\
		$\MS$ 
		& \includegraphics[width=\imgSizes]{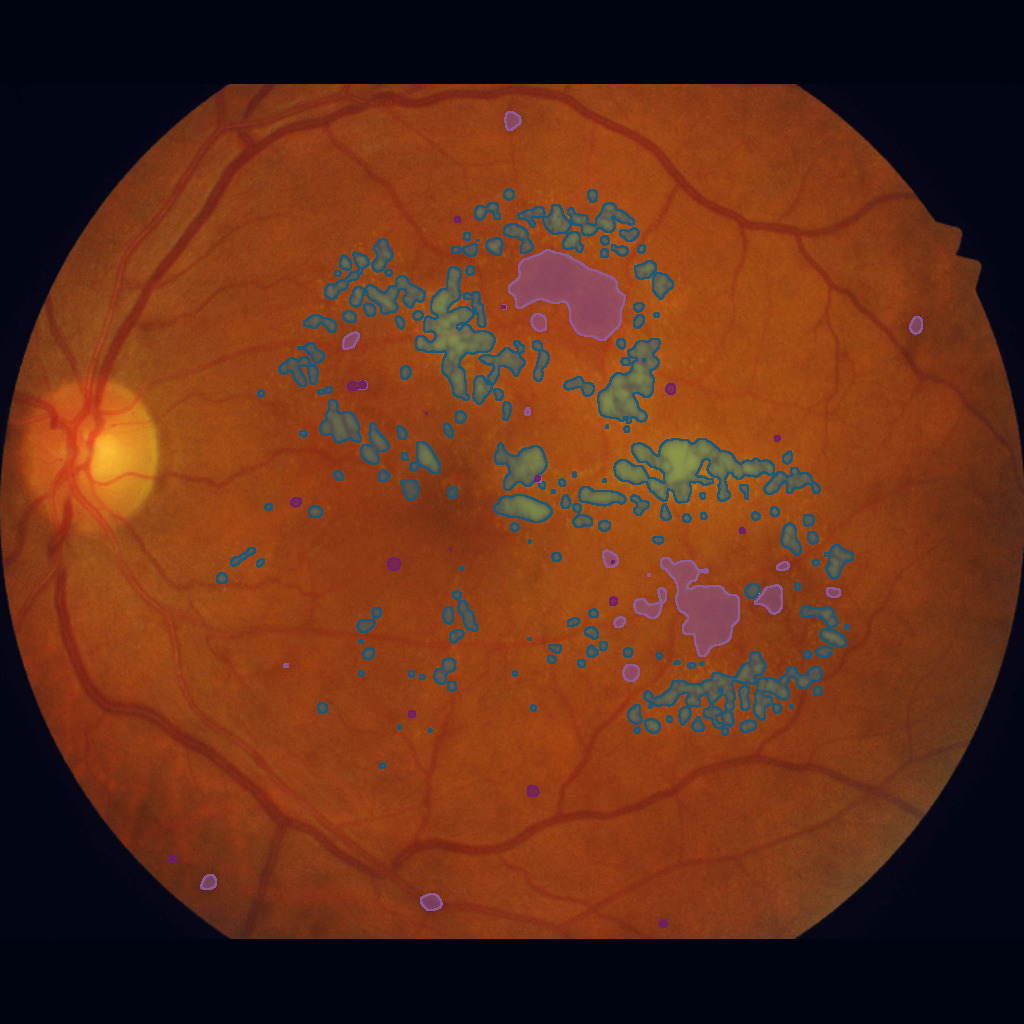} 
		& \includegraphics[width=\imgSizes]{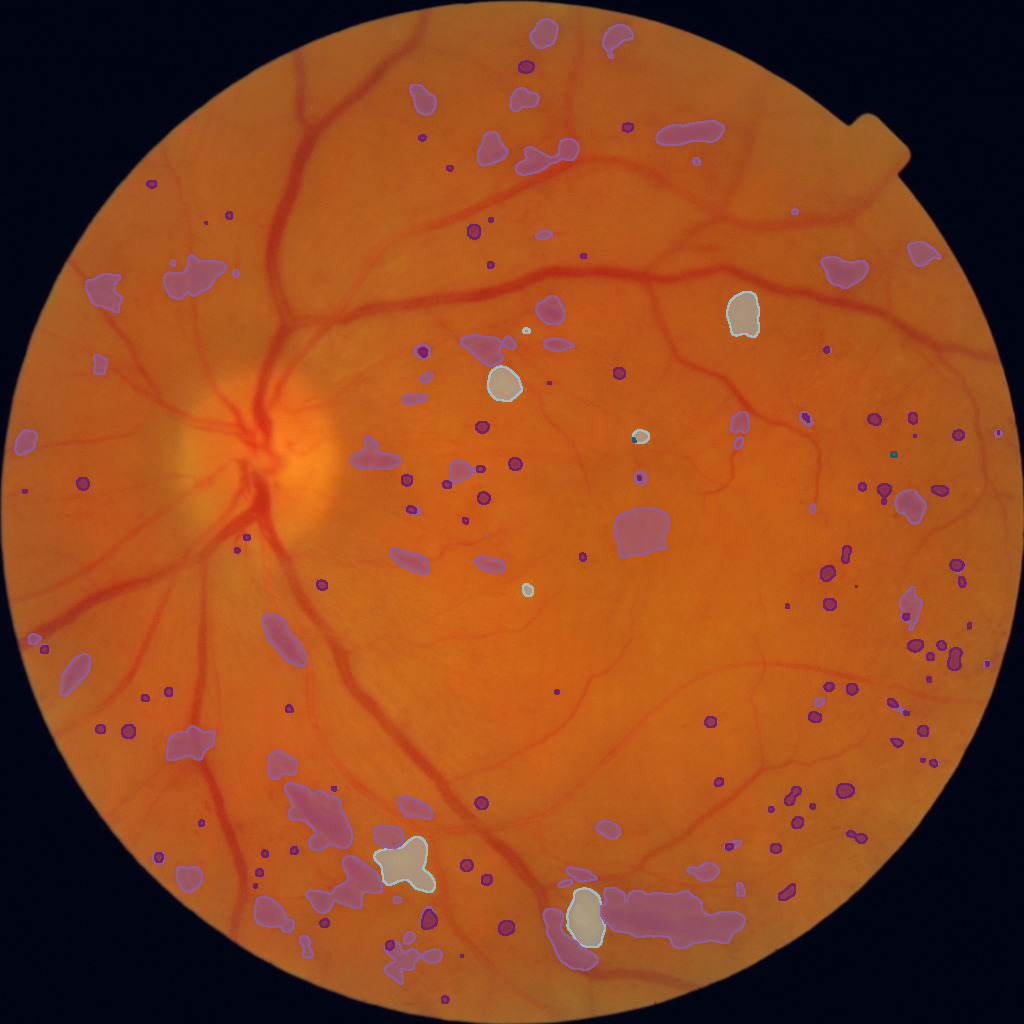} 
		& \includegraphics[width=\imgSizes]{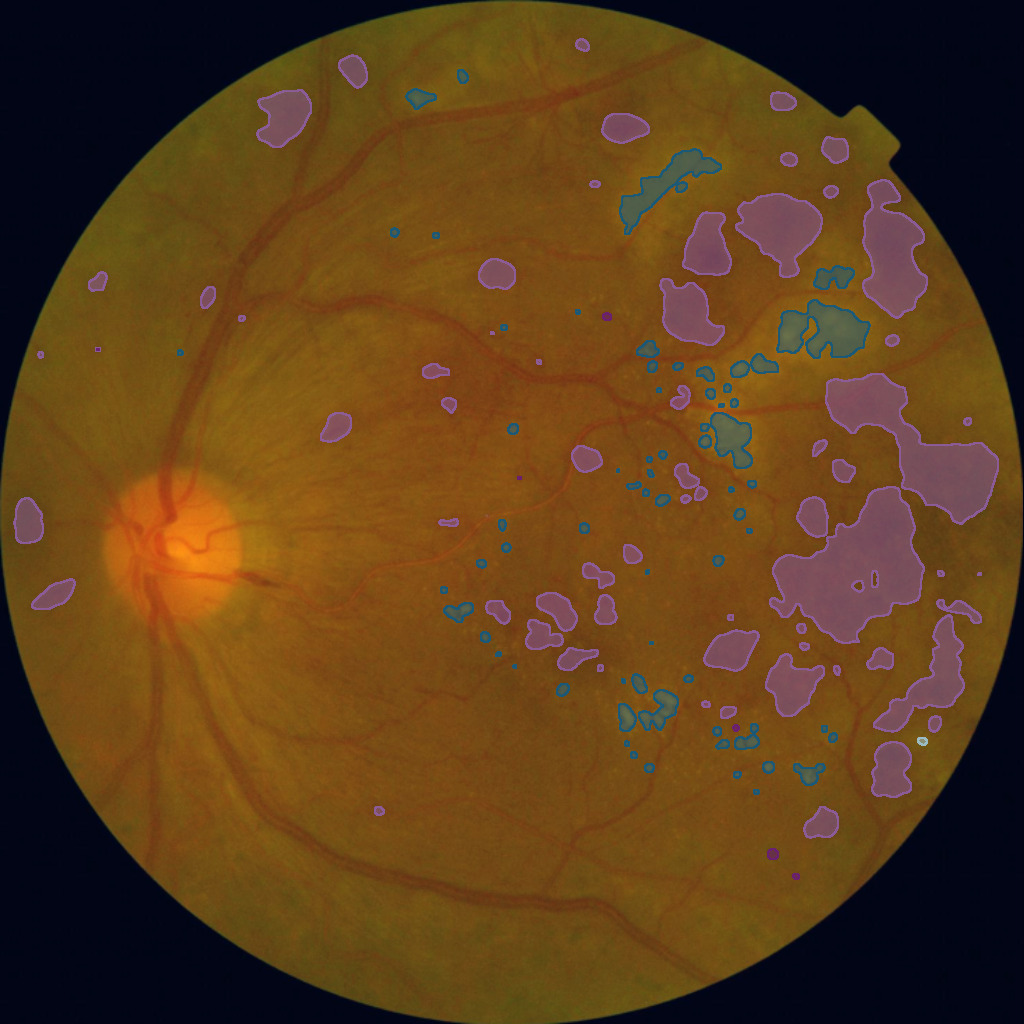} 
		& \includegraphics[width=\imgSizes]{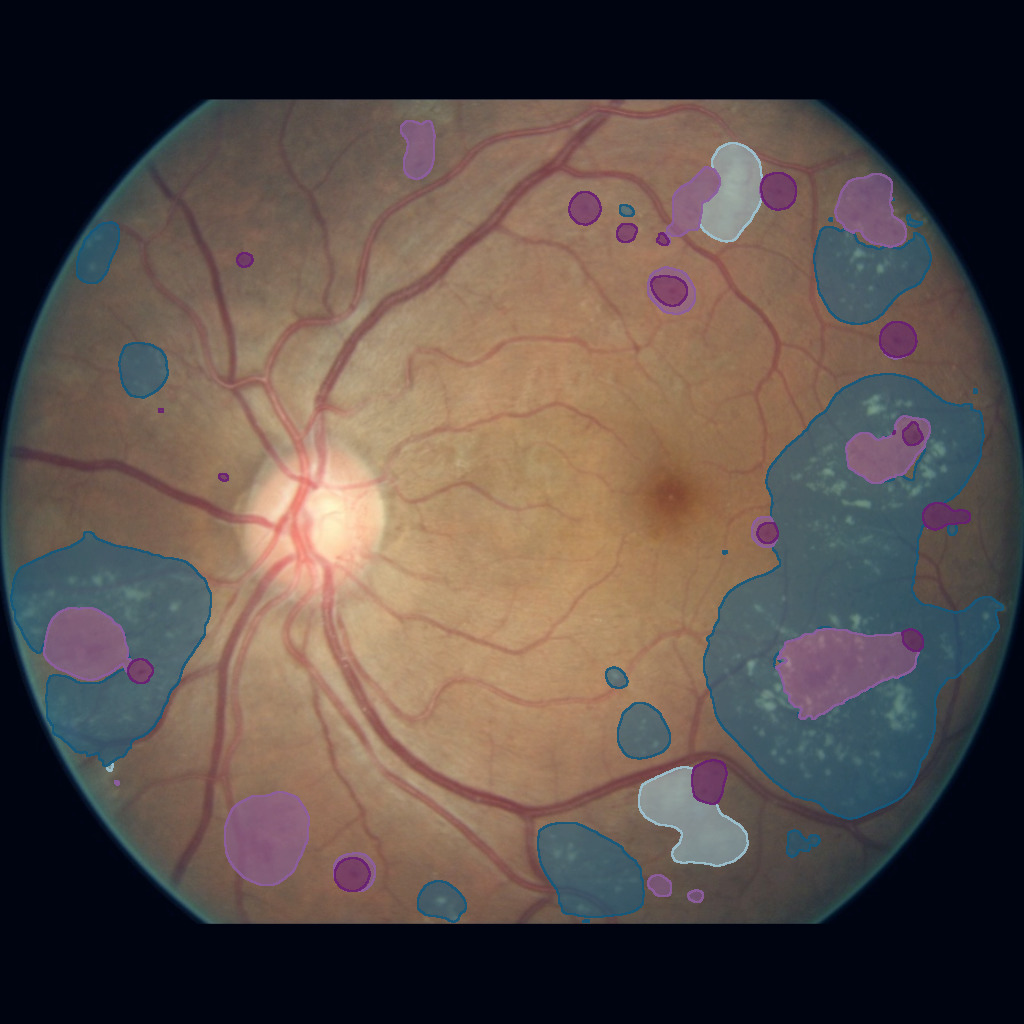} 
		& \includegraphics[width=\imgSizes]{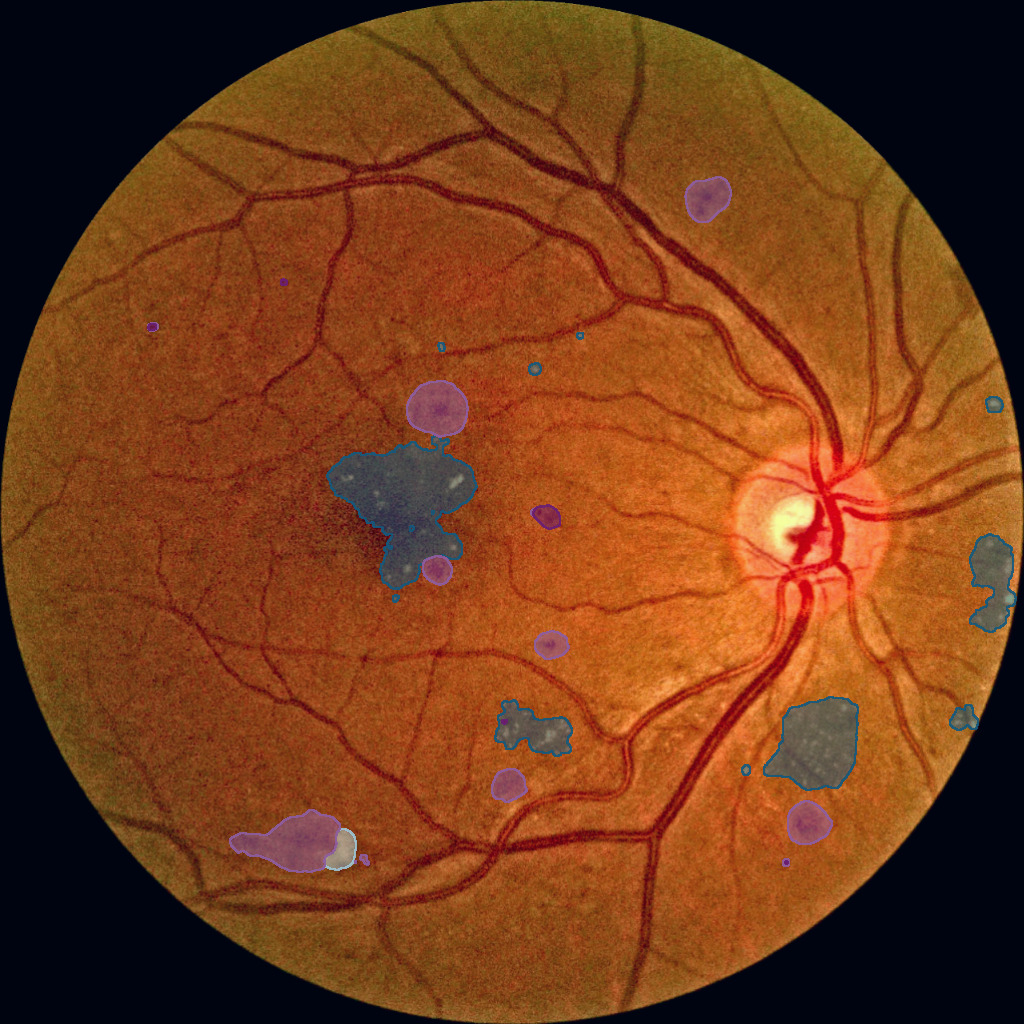} & \\

	\end{tabular}
\caption{Illustration of the differences between two specialised models (trained on the IDRiD and RET-LES datasets) and the generalist one. Note how $\MS$ changes its style based on the input image (particularly noticeable when comparing the first and fourth columns on exudates).}
\label{tab/fig:QualitativeSegExemples}
\end{table*}

In Table \ref{tab:crossDatasetResults}, we observe a counterintuitive behaviour of the generalist model $\MS$: its ability to maximize the performance on a majority of test sets (excepting solely MESSIDOR), even outperforming the ``specialised'' models. In our notation, this translates into:

\begin{equation}
\D{\M{\mathcal{S}}{j}}{\Btest{j}} \geq \D{\M{j}{j}}{\Btest{j}}, \forall j \neq {M}
\end{equation}

This observation holds in particular for the databases IDRiD and RETINAL-LESIONS, which have radically different labelling styles. Therefore, the only way for the model to maximise its performance on both test sets is to change its segmentation style on the fly. This effect is shown in Table \ref{tab/fig:QualitativeSegExemples}.
However, the model is never explicitly fed with information regarding the source of the images; therefore the only explanation behind this behaviour is that the images themselves contain a marker betraying their origin. In Section \ref{sec:sourceMarkerPerturbation}, we present a few experiments to identify what this marker could be based on. To highlight the model's ability to detect it, we build upon the idea of linear probes introduced by \cite{DBLP:conf/iclr/AlainB17}. In our case, the linear probe simply takes the features produced by the segmentation model's encoder and is trained to predict from which database they originate. We explore in Section \ref{sec:LinearProbePosition} the best positioning of the probe. Using a linear model for this purpose has a simple rationale: understanding how the segmentation model decodes the origin marker from the image is more important than using a complex classifier for the probe. Following this reasoning, during the probe's training, the segmentation model is frozen.

\subsection{Adversarial attack on the probe}
\label{sec:AdversarialAttack}
\begin{figure}[h!]
	\centering
	\includegraphics[width=.8\columnwidth]{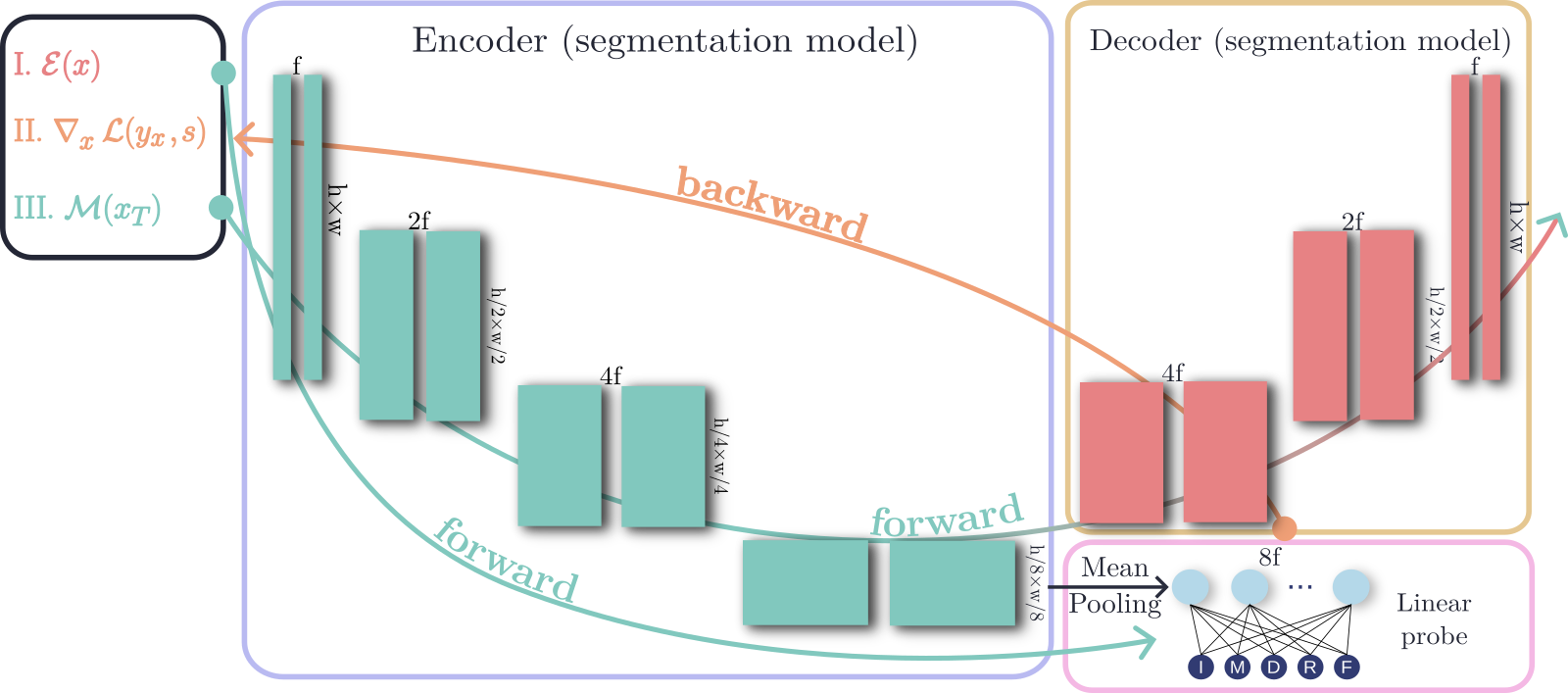}
	\caption{Graphical summary of our style conversion by adversarial attack.}
\label{fig:AdversarialAttack}
\end{figure}

Being able to detect the image's origin with an external probe serves little purpose in itself. Our main contribution relies on its accuracy and tweaks it to convert the segmentation model's style using adversarial attacks on the probe.
The concept of adversarial attacks was originally discovered by \cite{szegedyIntriguingPropertiesNeural2014} who describe them as an intriguing property of neural networks. Adversarial attacks are usually considered as a serious vulnerability of neural networks caused by their mostly linear nature and their sensitivity to gradients; however they can also be used as a form of regularisation (as in the work of \cite{goodfellowExplainingHarnessingAdversarial2015} \revisionAdd{ or more recently of \cite{croceRobustSemanticSegmentation2023} for semantic segmentation)}. Targeted adversarial attacks modify the input image in an imperceptible way (to the human eye) in order to force the classifier to predict a specific class called the target. The alteration is obtained using gradient descent in the direction that minimises the loss computed between the prediction and the target $t$. To conceive an optimal attack, \cite{goodfellowExplainingHarnessingAdversarial2015} suggest the ``Fast Gradient Sign Method'' (FGSM):
\begin{equation}
	x_{perturbed} = x - \epsilon \cdot \text{sign}(\nabla_x \mathcal{L}(y_x, t))
\end{equation}
where $x$ is the original image, $y_x$ the prediction of the classifier from $x$, $t$ the target class and $\mathcal{L}$ a loss function (usually Categorical Cross Entropy). \cite{madryDeepLearningModels2018a} further elaborates on this method by suggesting an iterative scheme called ``Projected Gradients'':
\begin{equation}
	x^{n+1} = \text{Proj}_{x + \mathcal{S}}(\text{FGSM}(x^n))
	\label{eq:ProjectedGradient}
\end{equation}
where $\mathcal{S}$ is the sphere centred on $x$ of allowed perturbations and $\text{Proj}$ is a re-normalisation operator casting the perturbed image within the radius of $\mathcal{S}$. This approach adds two additional parameters: the radius $r$ of $\mathcal{S}$ and the number of steps $N$ taken. Using an adversarial attack, we expect not only to fool the probe, but also the whole segmentation model, forcing it to adopt the style of our choice by ``overwriting" the source marker within the image. Figure \ref{fig:AdversarialAttack} illustrates this process. 
In practise, this technique is surprisingly effective, as shown in Table \ref{tab:Conversion of Styles}. \revisionAdd{Following this observation, we conducted a set of experiments to better understand what could influence the model toward one style or another, and to quantify the efficacy of our adversarial segmentation style conversion and its generalisation to unseen data and/or datasets.}

\def\imgSizesB{.5\columnwidth}
\begin{table}
	\centering
	\setlength\tabcolsep{0pt} 
	\renewcommand{\arraystretch}{0} 
	\begin{tabular}{cc}
		\toprule
		$\Btest{I}$ & $\Btest{I} \rightarrow R$ \\
		\midrule
		\includegraphics[width=\imgSizesB]{images/model_segmentation/model_ALL/Dataset.IDRID} 
		&
		\includegraphics[width=\imgSizesB]{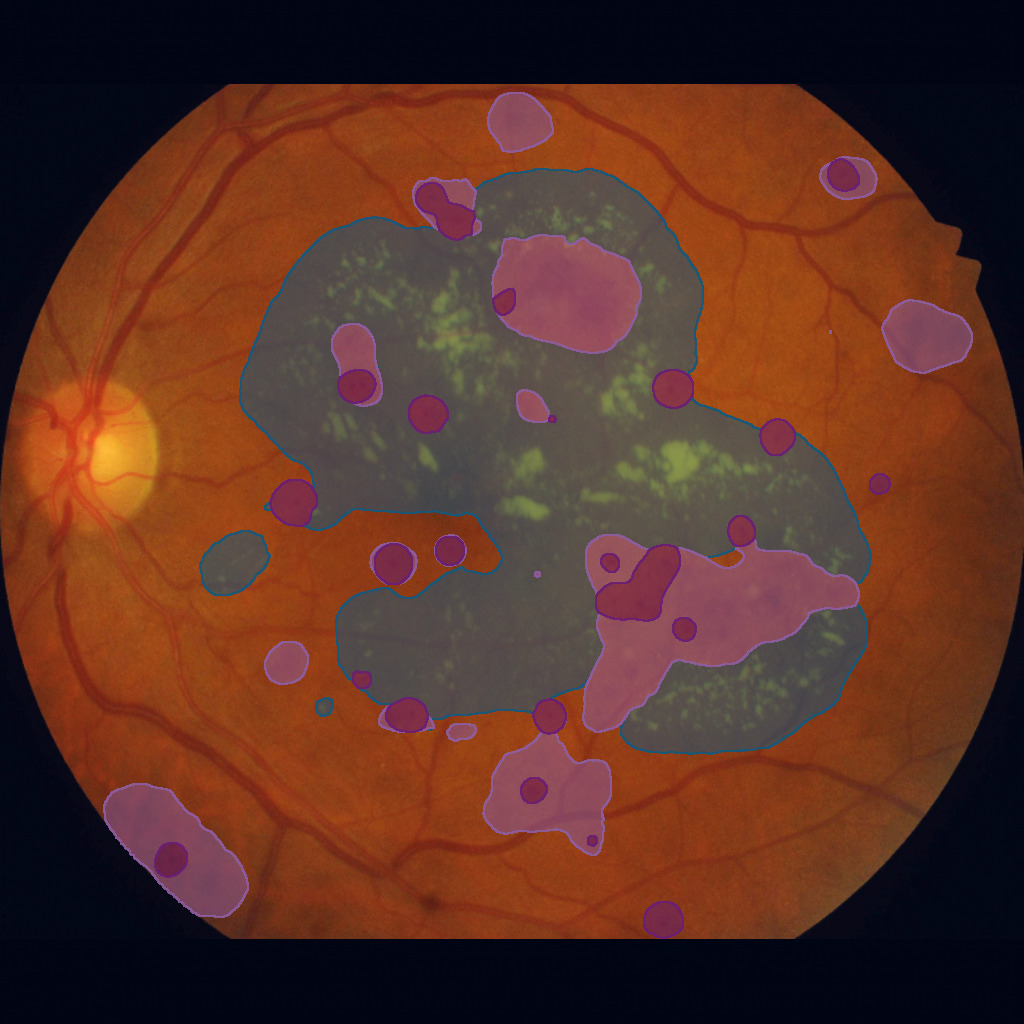} \\
		\midrule
		$\Btest{R}$ & $\Btest{R} \rightarrow I$ \\
		\includegraphics[width=\imgSizesB]{images/model_segmentation/model_ALL/Dataset.RETINAL_LESIONS} 
		&
		\includegraphics[width=\imgSizesB]{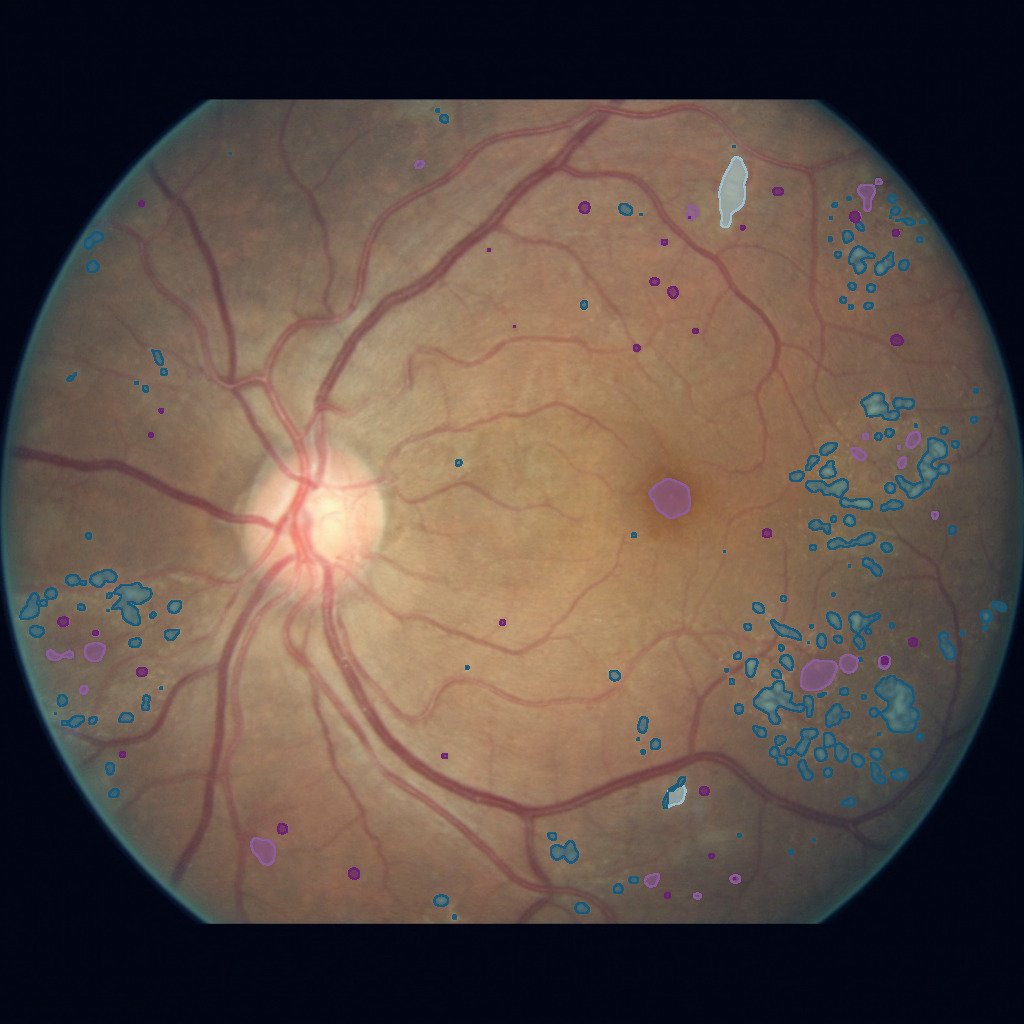} \\
		\bottomrule
	\end{tabular}
\caption{Adversarial style conversion from IDRID to RETINAL-LESIONS and conversely. All the segmentations were obtained with a single model $\MS$. The second column illustrates the adversarial conversion: it is imperceptible in the underlying fundus image but radically changes the lesion segmentation style depending on the target.}
\label{tab:Conversion of Styles}
\end{table}

\section{Experimental results}
\label{sec:experiments}

In this section, we explore in depth the results obtained with the different aspects of our methodology and extend the spectrum of its applications.


\subsection{Segmentation comparative performance}
\revisionAdd{To validate our training protocol and the choice of our architecture}, we compared the segmentation performance obtained with various architectures (and encoders per architecture). The models were trained (\revisionAdd{this included checkpointing at regular interval and selecting a model based on the best validation performance}) and tested on IDRiD following the conditions of the competition (\cite{porwalIDRiDDiabeticRetinopathy2020}).
\revisionFormat{The results are reported in Table \ref{tab:perfComparéeSegmentation}; we observe that our training procedure provides scores comparable with the best performances reported in the literature, even with different architectures. For the rest of this paper, we present the measures obtained with the UNet architecture with a ResNet-34 encoder.}

\begin{table*}
	\centering
	\begin{threeparttable}
		\begin{tabular}{l lcccc |c}
			\toprule
			\multicolumn{7}{c}{Our trained models} \\
			\midrule
			Architecture & Encoder & MA & HEM & CWS & EX & Average\\
			\midrule
			\multirow{5}{5em}{UNet \citep{ronnebergerUNetConvolutionalNetworks2015a}}
			& ResNet-18 \citep{zagoruykoWideResidualNetworks2016} & 0.4892 & 0.6407 & 0.7089 & 0.8512 & 0.6725 \\
			& ResNet-34 & 0.4958 & 0.6457 & 0.7033 & 0.8415 & 0.6716 \\
			& ResNest-50 \citep{zhangResNeStSplitAttentionNetworks2022} & 0.5041 & 0.6182 & 0.6289 & 0.821 & 0.6431 \\
			& SE ResNet-50 \citep{huSqueezeandExcitationNetworks2018} & 0.4730 & 0.6111 & 0.6803 & 0.8233 & 0.6469 \\
            & SE ResNext-50 \citep{xieAggregatedResidualTransformations2017} & 0.3965 & 0.6880  & 0.6725 & 0.8319  & 0.6472 \\
			&  \tnote{1} MIT B2 \citep{xieSegFormerSimpleEfficient} & \color{orange}\textbf{0.5123} & 0.5749 & 0.7051 & 0.8408 & 0.6583 \\
			& \tnote{1} MIT B4 & 0.5045 & 0.6473 & 0.6959 & 0.8251 & 0.6682 \\
			\midrule
			\multirow{3}{5em}{UNet++ \citep{zhouUNetNestedUNet2018}} & ResNet-18 & 0.4955 & 0.6348 & 0.7063 & \color{orange}\textbf{0.8531} & 0.6724 \\
			& ResNest-50 & 0.4900 & 0.6601 & 0.6876 & 0.8019 & 0.6599 \\
			& SE ResNet-50 & 0.4906 & 0.6141 & 0.7273 & 0.8169 & 0.6622 \\
			\midrule
			\multirow{4}{5em}{FPN \citep{seferbekovFeaturePyramidNetwork2018}} & ResNet-18 & 0.4524 & 0.6476 & 0.7260 & 0.8229 & 0.6622 \\
			& ResNest-50 & 0.4870 & \color{orange}\textbf{0.6898} & \color{orange}\textbf{0.7529} & 0.8246 & \color{orange}\textbf{0.6886} \\
			& SE ResNet-50 & 0.4576 & 0.6790 & 0.7396 & 0.8169 & 0.6733 \\
			& MobileNet V3 \citep{sandlerMobileNetV2InvertedResiduals2018} & 0.3498 & 0.5828 & 0.6348 & 0.7509 & 0.5796 \\
			\midrule
			\multirow{3}{5em}{DeepLab V3+ \citep{chenDeepLabSemanticImage2018}} & ResNet - 18 & 0.4515 & 0.6426 & 0.6967 & 0.8098 & 0.6502 \\
			& ResNet-34 & 0.4238 & 0.6073 & 0.6329 & 0.8329 & 0.6242 \\
			& SE ResNet-50 & 0.4623 & 0.6868 & 0.7049 & 0.8204 & 0.6686 \\
			\midrule
			\tnote{3} Global-Local UNet & ResNest - 50 & 0.4580 & 0.6724 & 0.7080 & 0.8263 & 0.6662 \\
			\midrule
			\multicolumn{7}{c}{Published results}\\
			\midrule
			\multicolumn{2}{l}{L-Seg \citep{guoLSegEndtoendUnified2019}} & 0.4630 & 0.6370  & 0.7110 & 0.7950 & 0.6515 \\
			
			\multicolumn{2}{l}{Deep-Bayesian \citep{garifullinDeepBayesianBaseline2021}} & 0.4840 & 0.5930 & 0.6410 & 0.8420 & 0.6400 \\
			\multicolumn{2}{l}{\tnote{2} Global-Local UNet \citep{yanLearningMutuallyLocalGlobal2019}} & \textbf{0.5250} & \textbf{0.7030} & 0.6790 & \textbf{0.8890} & 0.6990 \\
			\multicolumn{2}{l}{CARNet \citep{guoCARNetCascadeAttentive2022}} & 0.5148 & 0.6389 & 0.7215 & 0.8675 & 0.6857 \\
			\multicolumn{2}{l}{\tnote{4} Xception-UNet - Collaborative learning \citep{zhouCollaborativeLearningSemiSupervised2019}} & 0.4960 & 0.6936 & \textbf{0.7407} & 0.8872 & \textbf{0.7044} \\
			\midrule
			\multicolumn{7}{c}{IDRiD Official leaderboard}\\
			\midrule
			\multicolumn{2}{l}{Team} &  \\ 
			\midrule
			\multicolumn{2}{l}{VRT}& 0.4951 & 0.6804 & 0.6995 & 0.7127 & 0.6469 \\
			\multicolumn{2}{l}{PATech}& 0.4740 & 0.6490 & \multicolumn{1}{c}{-} & 0.8850 & \multicolumn{1}{c}{-}\\
			\multicolumn{2}{l}{iFLYTEK-MIG}& 0.5017 & 0.5588 & 0.6588 & 0.8741 & 0.6483 \\
			\multicolumn{2}{l}{SOONER}& 0.4003 & 0.5395 & 0.5369 & 0.7390 & 0.5539\\
			\bottomrule
		\end{tabular}
	\caption {Comparative performance analysis among the various tested architectures. The models were trained and evaluated on the IDRiD dataset (partitioned into two sets following the competition rules). Our highest scores are denoted in orange, while the state-of-the-art scores are highlighted in bold.}
	\label{tab:perfComparéeSegmentation}
		\begin{tablenotes}
			\item[1] ViT like encoder following the idea of the SegFormer.
			\item[2] The original model is actually composed of 4 networks, one per lesion.
			\item[3] Our re-implementation is multi-class.
			\item[4] This models combines strong and weak supervision.
		\end{tablenotes}
	\end{threeparttable}

\end{table*}

\subsection{Origin marker and sensitivity to perturbation}
\label{sec:sourceMarkerPerturbation}

The spontaneous conversion of $\Ms$'s style depending on the data fed to it was unexpected and brings into question how the model learns to do this. 
We conducted a set of experiments to assess if this conversion behaviour could be altered by simple transformations of the input images. Our initial hypothesis was that different clusters of images could have been identified by $\Ms$ in an unsupervised way based either on their resolution (despite our standardisation protocol, the databases originally have varying image sizes), on the images' colour distribution (due to the diversity of acquisition hardware used or population ethnicities) or on the compression format used for storing the images (PNG or JPEG with different levels of compression). We tested this hypothesis qualitatively by trying to alter $\Ms$'s segmentation by incorporating random image modifications. Results are shown in Figure \ref{fig:PertubationsSyle}. Overall, we did not observe a radical shift in the model's output style with these simple perturbations.

\def\colfigsExemple{.33\textwidth}

\begin{figure*}[h]
	\centering
	\begin{subfigure}[t]{\colfigsExemple}
		\includegraphics[width=\columnwidth]{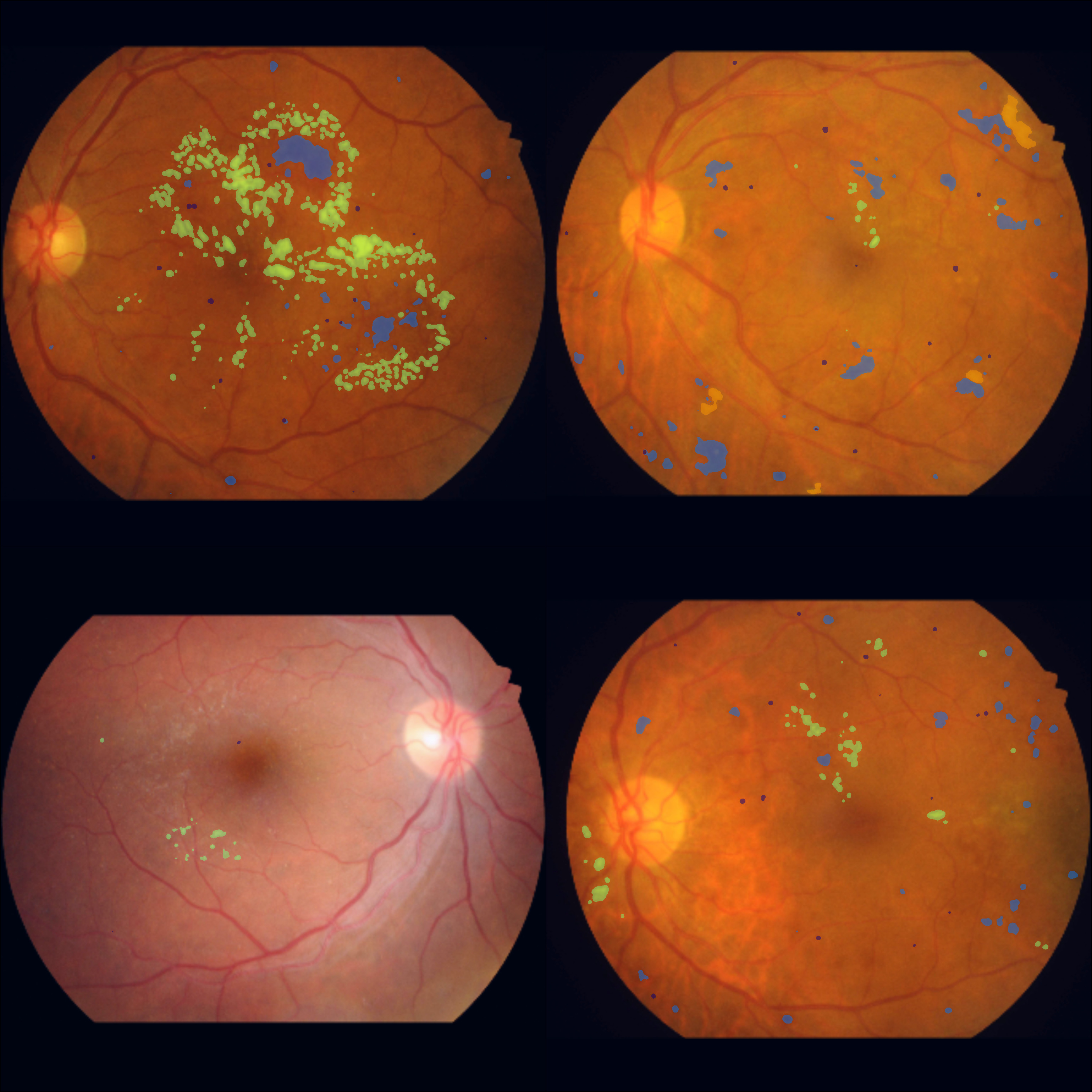}
		\subcaption{Random resampling $\pm 50\%$}
	\end{subfigure}
\begin{subfigure}[t]{\colfigsExemple}
	\includegraphics[width=\columnwidth]{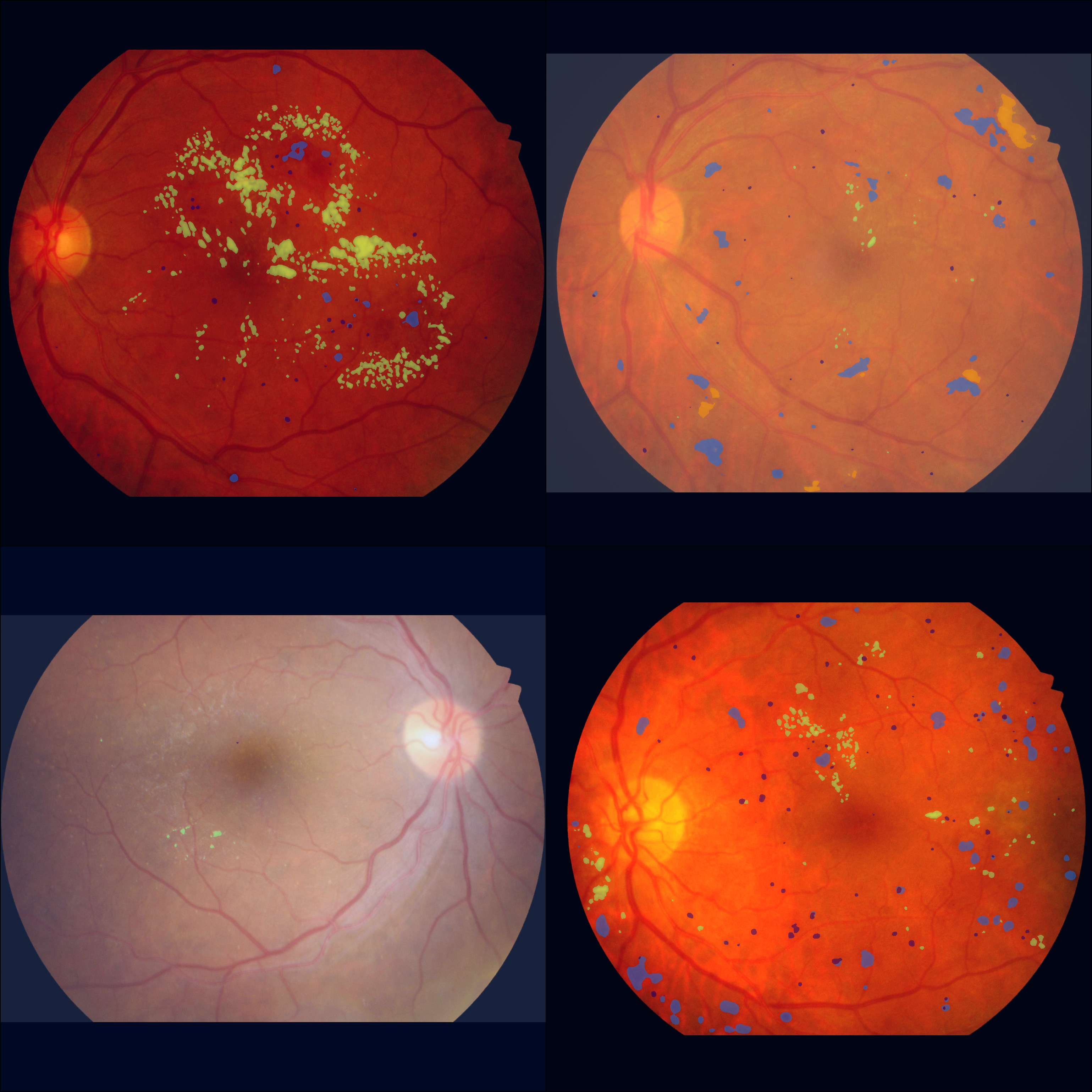}
	\subcaption{Random color jittering $\pm 20\%$ (Hue, saturation and value)}
\end{subfigure}
\begin{subfigure}[t]{\colfigsExemple}
	\includegraphics[width=\columnwidth]{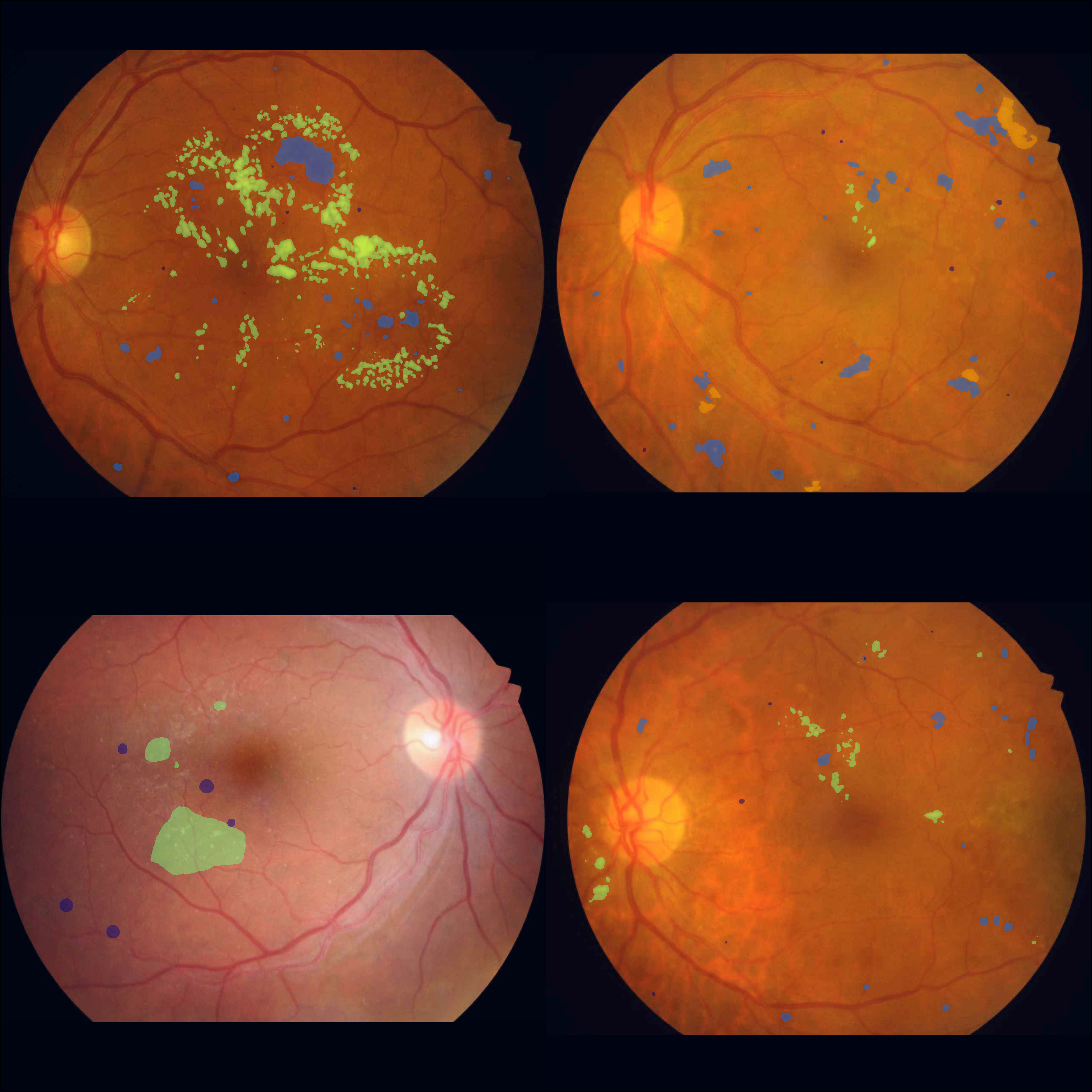}
	\subcaption{Random JPEG artefacts (compression down to $50\%$)}
\end{subfigure}
\caption{Effect of random perturbations of the input images on the segmentations by $\Ms$ (shown for four test images). Interestingly, the model appears to be robust to most perturbations. Compression artefacts may however partially fool the model toward a new style, as seen in the bottom left image in (c).} 
\label{fig:PertubationsSyle}
\end{figure*}

\subsection{Probe positioning within the \review{network}}
\label{sec:LinearProbePosition}

\begin{figure}[h]
\centering
	\includegraphics[width=.8\columnwidth]{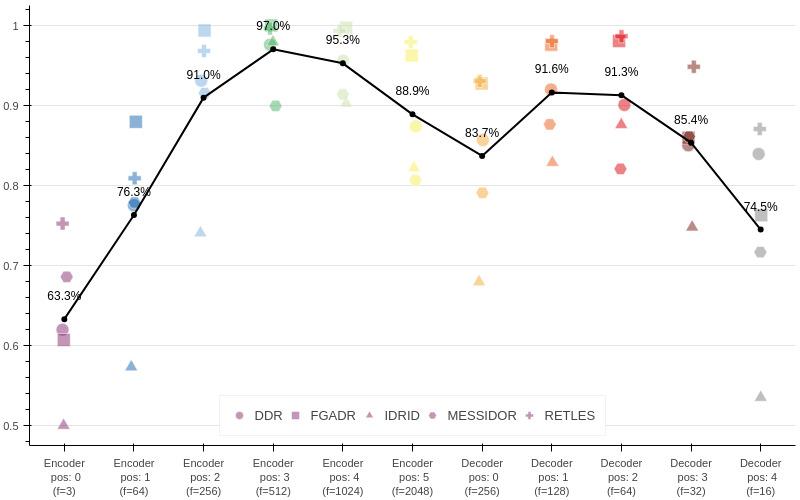}
	\caption{\review{Accuracy of the probe depending on its position in the model. As the number of channels grows with the depth, the size $f$ of the input latent vector fed to the probe increases (it is extracted by spatial average pooling of the encoder's features).}} 
	\label{fig:ProbePlacement}
\end{figure}

We studied different placements of the probe within the \revisionAdd{encoder and the decoder} of the segmentation model. Depending on the features received, the probe has more or less context to accurately predict the image's origin. Figure \ref{fig:ProbePlacement} illustrates this effect: for all the images in our validation sets, we measured the ability of the probe $\mathcal{P}^{(l)}$ to predict $\mathcal{P}^{(l)}(\Btrain{i}) \stackrel{?}{=} i$, where $l$ is the depth within the encoder. The maximum (and almost perfect) accuracy is obtained when the probe is placed \revisionAdd{at the lower levels of the encoder}. 

\subsection{Generalising conversion to external data}
So far, we have highlighted the effect of the conversion on data distributions that were seen by the segmentation models and/or the probe, i.e. coming from one of the five datasets studied. To broaden the applicability of our methodology, we introduce a supplementary dataset in our work. APTOS (for Asia Pacific Tele-Ophthalmology Society) was released in 2019 as part of a Kaggle competition \cite{aptos2019-blindness-detection}. 
It provides 3662 images from the Aravind Eye Hospital in India. Segmentation-wise, these images are unlabelled. We refer to this base as $\Btest{A}$, and used it to demonstrate the generalisation of our technique.
\revisionAdd{We conducted two experiments on these data: first, we verified the ability to fool the probe toward any of the five targets after adversarial attack. Then, we measured how close were the predictions of $\MS$ after conversion toward a style $i$ and the corresponding prediction obtained with the specialized model $\Mi{i}$}.

\subsubsection{Adversarial attack on the probe}
We evaluated the ability to fool the probe into predicting a target class from images of the APTOS dataset, i.e: 

\begin{equation}
	\mathcal{P}(\Btest{A} \rightarrow i) \stackrel{?}{=} i
\end{equation}

This experiment also served to clarify the parameters' roles in the Projected Gradient algorithm (Equation \ref{eq:ProjectedGradient}). Table \ref{tab:ProjectedGradientPerf} details these results. In addition, we use this experiment to measure the speed of the conversion. It varies from 18 images per second ($N=1$) to 1.1 i.p.s ($N=25$). \revisionAdd{In all experiments, we set $r=\frac{5}{255}$.}

\begin{table}[h]
	\centering
	\begin{tabular}{cc|ccccc}
		\toprule
		Step& \# steps & \multicolumn{5}{c}{ $\mathcal{P}(\Btest{A} \rightarrow i), i=$ } \\
		\midrule
		$\epsilon$ & $N$ & $I$ & $M$ & $D$ &$R$&$F$ \\ 
		\midrule
		$2.5\cdot 10^{-2}$ & 1 &71.4  & 53.6  & 97.6 & 95.7 & 74.5\\
		$5.0\cdot 10^{-3}$ & 5 & 100 & 99.8 & 100 & 100 & 100  \\
		$2.5\cdot 10^{-4}$ & 10 & 100 & 100 & 100 & 100 & 100  \\
		$1.0\cdot 10^{-4}$ & 25 & 100 & 100 & 100 & 100 & 100 \\
		\bottomrule
	\end{tabular}

\caption{Probe's accuracy (in \%) in predicting the target class $i$ after adversarial attack on images from Aptos. We studied the effect of step size $\epsilon$ and number of steps $N$ (with $\epsilon \times N$ kept constant).}
\label{tab:ProjectedGradientPerf}
\end{table}

\subsubsection{Segmentation style conversion}
As observed in Table \ref{tab:Conversion of Styles}, the adversarial attack does not only affect the probe, but also the whole segmentation model. Effectively, the style conversion appears to work on the Aptos images (as shown in Figure \ref{fig:ContinuousInterpolationStyle}). However, it is hard to quantitatively evaluate this effect, given that we don't have labels for Aptos, 
not to mention different groundtruth styles per image. 
As a proxy, we generate our own groundtruths using the different specialised models $\Mi{j}$, which we compare with the predictions $\MS(\Btest{A} \rightarrow i)$. 
Formally, using our notation, this is equivalent to measuring:

\begin{equation}
	\mathcal{D}(\M{i}{A}, \Ms^{(A \rightarrow j)_\star})
\end{equation}

Results are given in Table \ref{tab:CrossDatasetEvaluation}. Logically, we expected to find the highest score for $i=j$. This is verified for all datasets except FGADR and MESSIDOR. Even between very dissimilar labelling styles (such as IDRiD and RETINAL-LESIONS), the conversion appears to be quite effective.

\begin{table}[h]
	\centering
	\begin{tabular}{c|ccccc}
		\toprule
		& \multicolumn{5}{c}{$\Ms(\Btest{A} \rightarrow i), i=$}\\ 
		\midrule
		$\Mi{j}(\Btest{A}), j=$ & $I$ & $M$ & $D$ &$R$ & $F$ \\
		\midrule
		$I$& \color{orange}\textbf{0.451} & \textbf{0.400} & 0.407 & 0.258 & 0.447\\
		$M$& 0.361 & 0.358 & \color{orange}0.363 & 0.232 & 0.357\\
		$D$& 0.446 & 0.383 & \textbf{0.506} & 0.258 & \color{orange}\textbf{0.534}\\
		$R$& 0.277 & 0.284 & 0.259 & \color{orange}\textbf{0.421} & 0.286\\
		$F$& 0.360 & 0.334 & 0.343 & 0.279 & \color{orange}0.386\\
		\bottomrule
	\end{tabular}
\caption{Cross-evaluation (using mIoU metric) between the specialised models (in each row) and a single generalist one converted to different target styles. In \textbf{bold}, we indicate the maximum per column and in {\color{orange} orange} per row. }
\label{tab:CrossDatasetEvaluation}
\end{table}

\revisionAdd{\subsection{Comparison with an existing approach}}
\revisionAdd{As we mentioned in our literature review, our work is at the relatively unique intersection of semantic segmentation of retinal lesions and style adaptation from multiple domains. To our knowledge, the work of \cite{zepf2023label} is the only one that distinguishes the concept of annotation style (due to biased annotation protocols) and aleatoric uncertainty (from noisy and possibly unbiased errors). For comparison, we have therefore adopted their idea to train a Conditional Stochastic Segmentation Network (C-SSN), following the original architecture of \cite{monteiroStochasticSegmentationNetworks2020}. The principle involves modeling the probability distribution of a segmentation map conditioned on the input image and a style, following a Gaussian law whose parameters are estimated by the neural network. For comparison purposes, we have re-implemented this model using the same segmentation architecture as our model \(\mathcal{M}_{S}\). We used the cost function defined by \cite{monteiroStochasticSegmentationNetworks2020}:
\begin{align}
l &= - \text{logsumexp}_{m=1}^{M}(\sum_{i=1}^{S}(\log(p(\mathbf{y_i}| \mathbf{\eta}_i^{(m)})) + \log(M), \\& \mathbf{\eta}^{(m)}|\mathbf{x},d \sim \mathcal{N}(\mathbf{\mu(x}, d), \mathbf{\Sigma(x}, d))
\end{align}
where $M=50$ is the number of Monte-Carlo samples, $S$ the number of pixels and $\mathbf{\mu(x}, d), \mathbf{\Sigma(x}, d)$ the predicted parameters of the distribution. In our implementation, $d$ is an integer (from one to five) indicating the origin of the image $\mathbf{x}$. For further details on how the model is built, we refer to \cite{zepf2023label} and our code repository. As suggested by \cite{monteiroStochasticSegmentationNetworks2020}, we used the RMSProp optimizer. In other respects, we maintained the training configuration described in Section \ref{sec:TrainingDetails}.}

\def\colwidth{0.24\columnwidth}
\begin{figure*}[h]
    \centering
    \begin{subfigure}{\colwidth}
    \subcaption[t]{Images from TJ-DR}
    \label{fig-TJDRImage}
    \includegraphics[trim={0 1027px 0 0},clip=true,width=\textwidth]{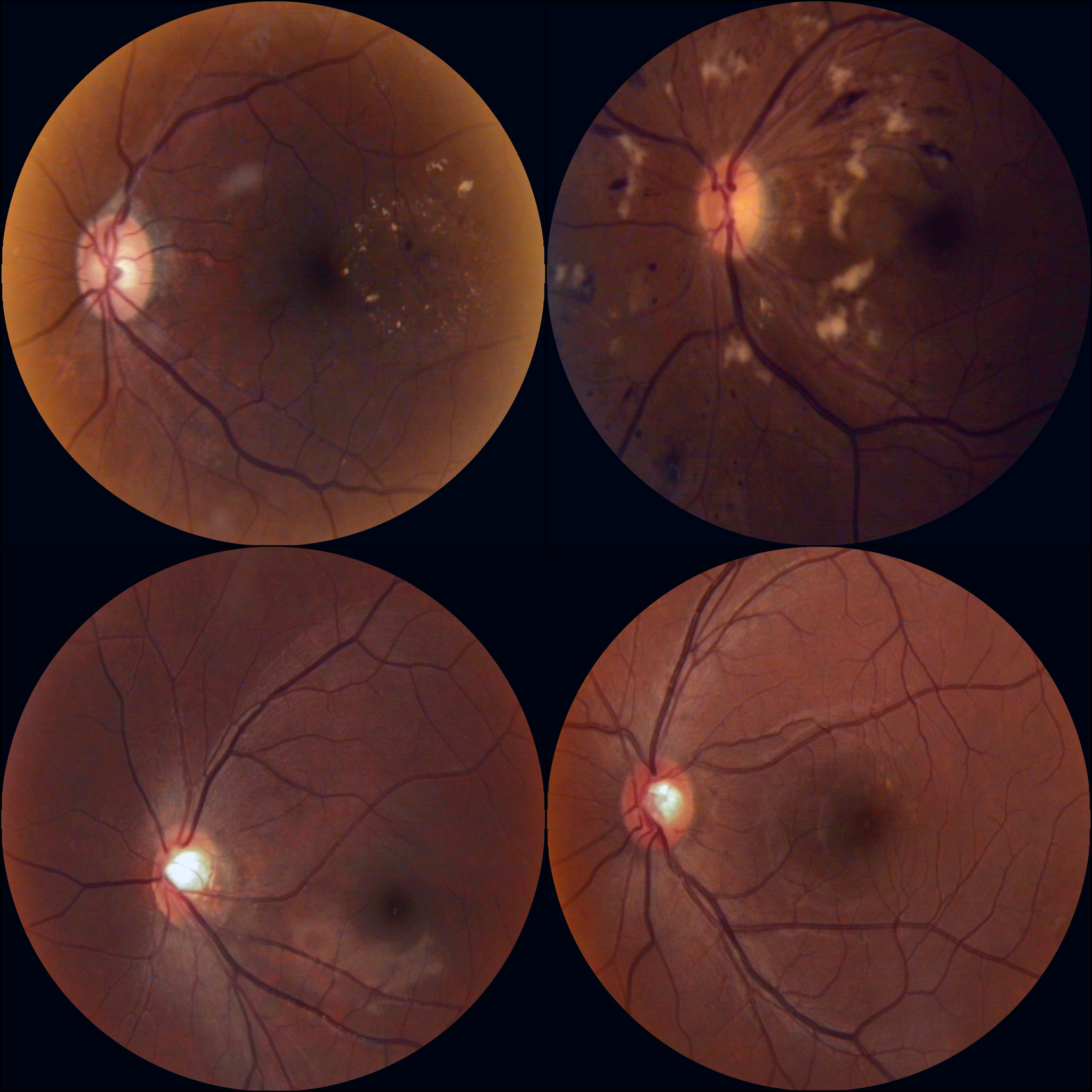}
    \end{subfigure}
    \begin{subfigure}{\colwidth}
    \subcaption[t]{C-SNN(R)}
    \label{fig-CSNNR}
    \includegraphics[trim={0 1027px 0 0},clip=true,width=\textwidth]{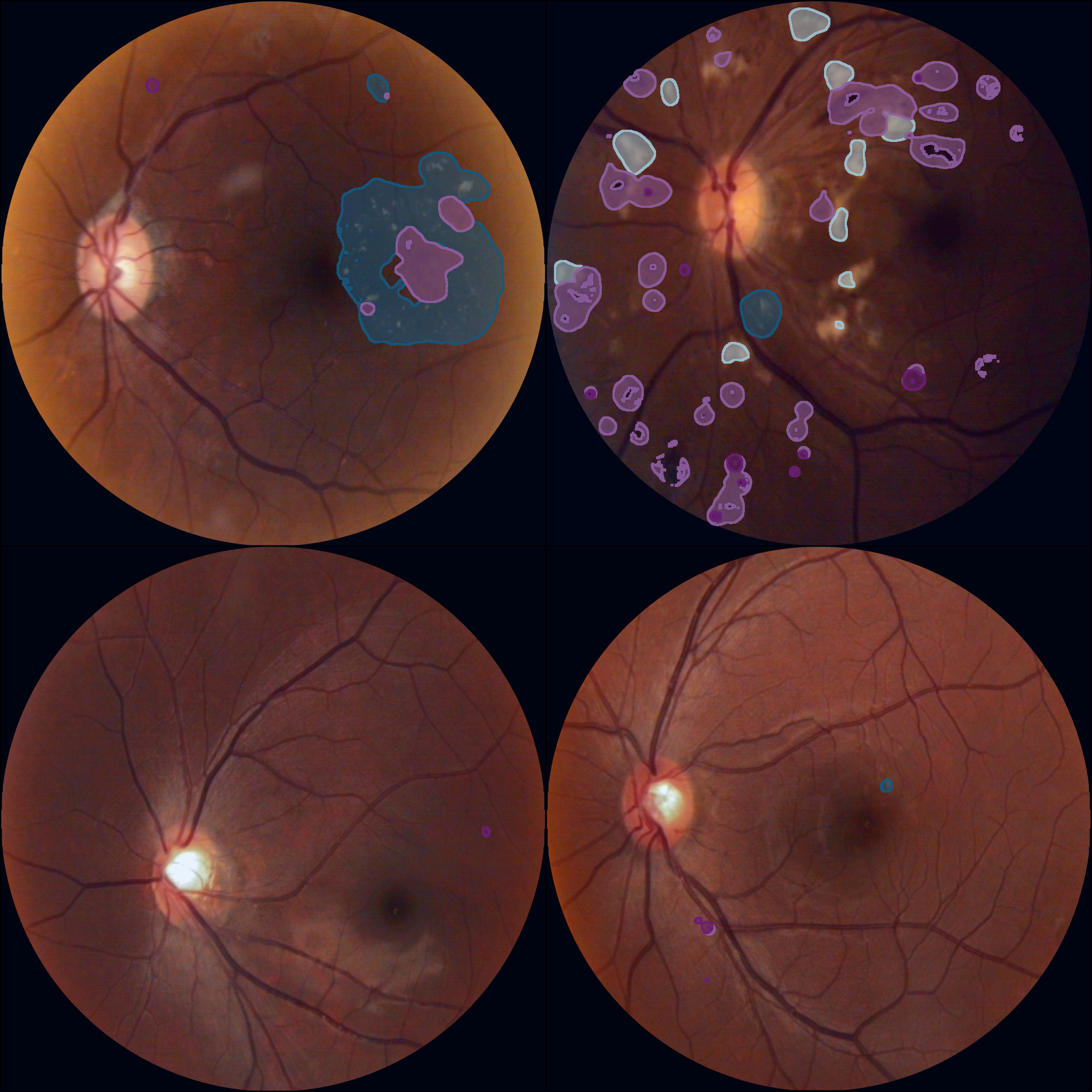}
    \end{subfigure}
    \begin{subfigure}{\colwidth}
    \subcaption[t]{C-SNN(D)}
    \label{fig-CSNND}
        \includegraphics[trim={0 1027px 0 0},clip,width=\textwidth]{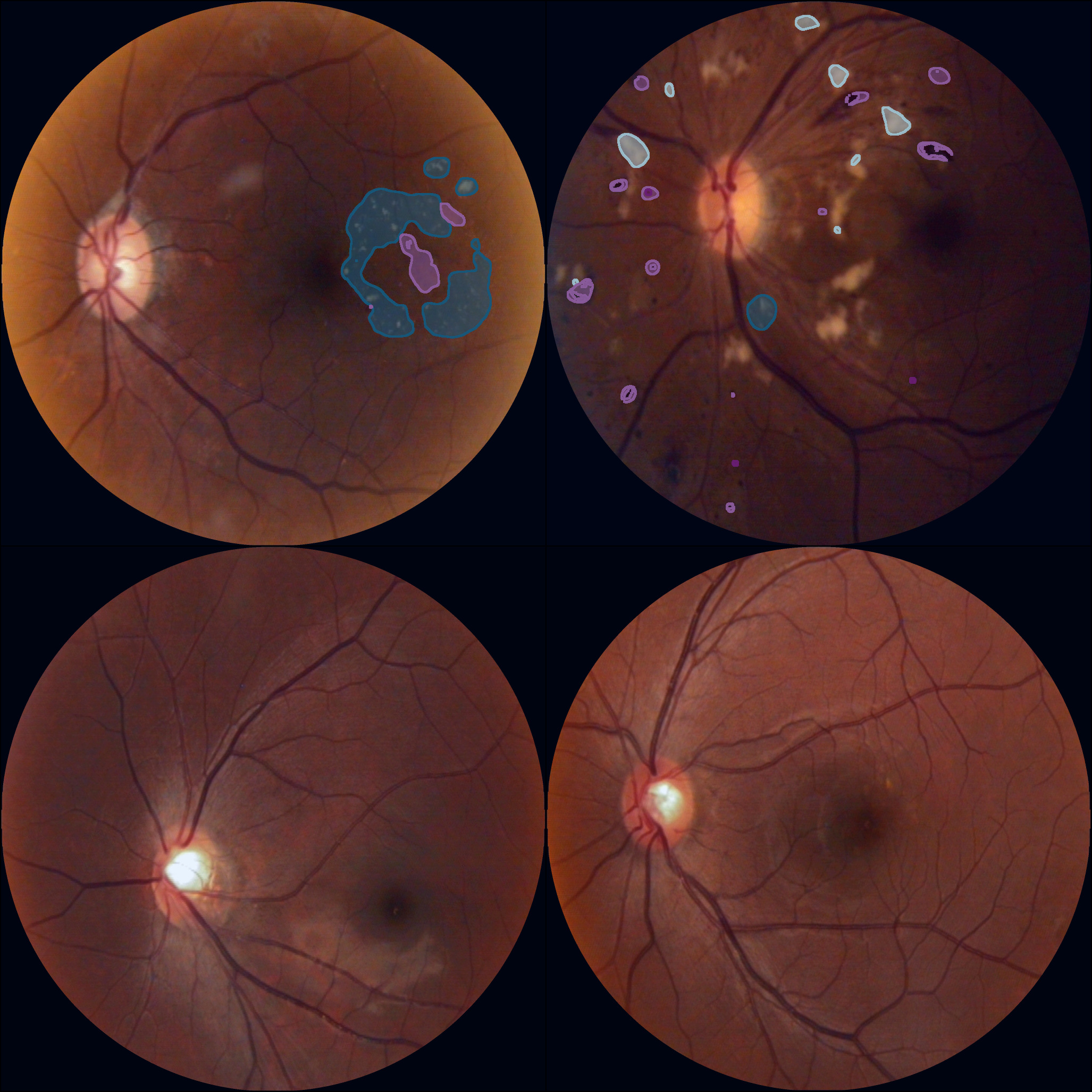}
    \end{subfigure}
     \begin{subfigure}{\colwidth}
         \subcaption[t]{C-SNN(I)}
         \label{fig-CSNNM}
        \includegraphics[trim={0 1027px 0 0},clip,width=\textwidth]{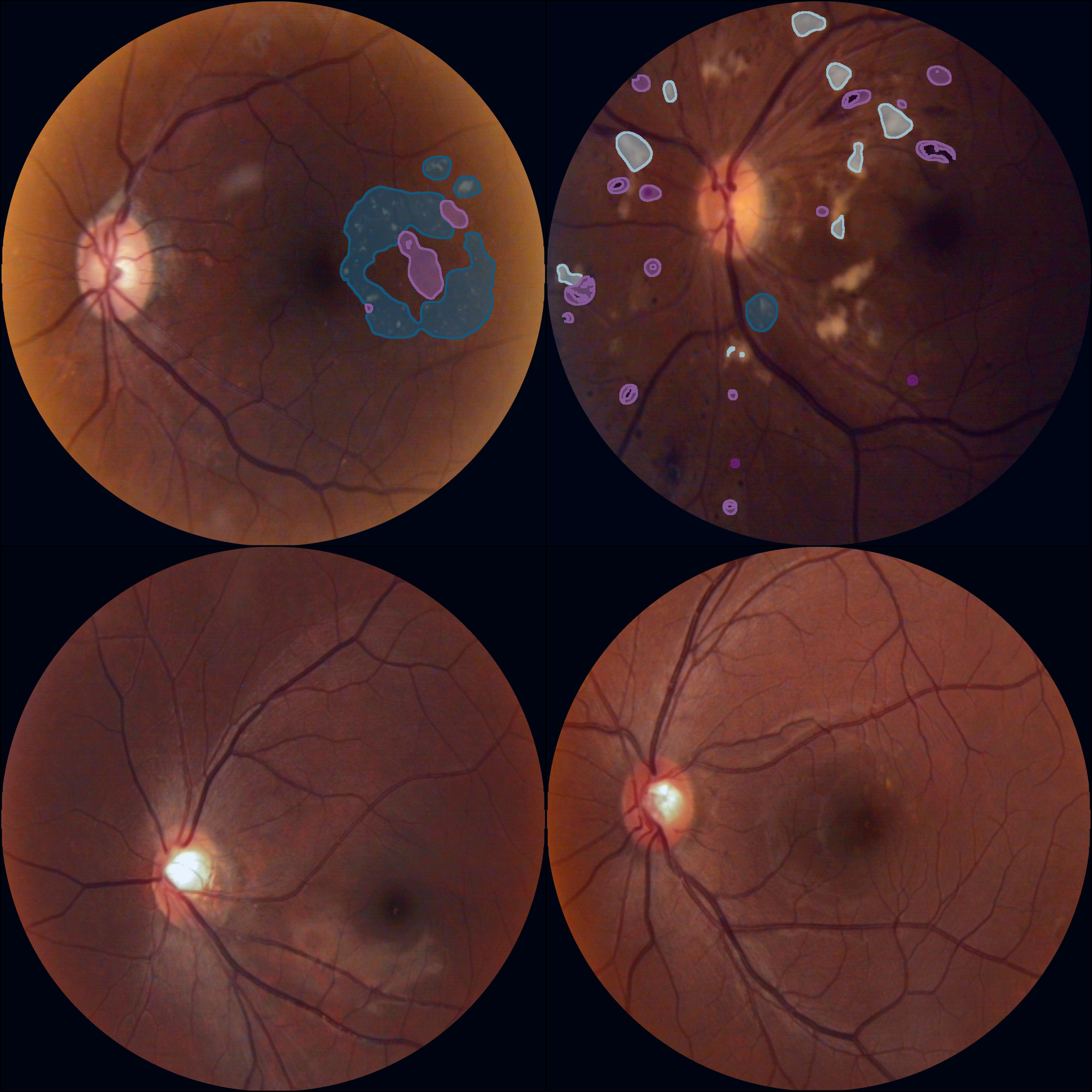}
    \end{subfigure}
    \begin{subfigure}{\colwidth}
    \subcaption[t]{Groundtruth}
    \label{fig-Groundtruth} 
    \includegraphics[trim={0 1027px 0 0},clip=true,width=\textwidth]{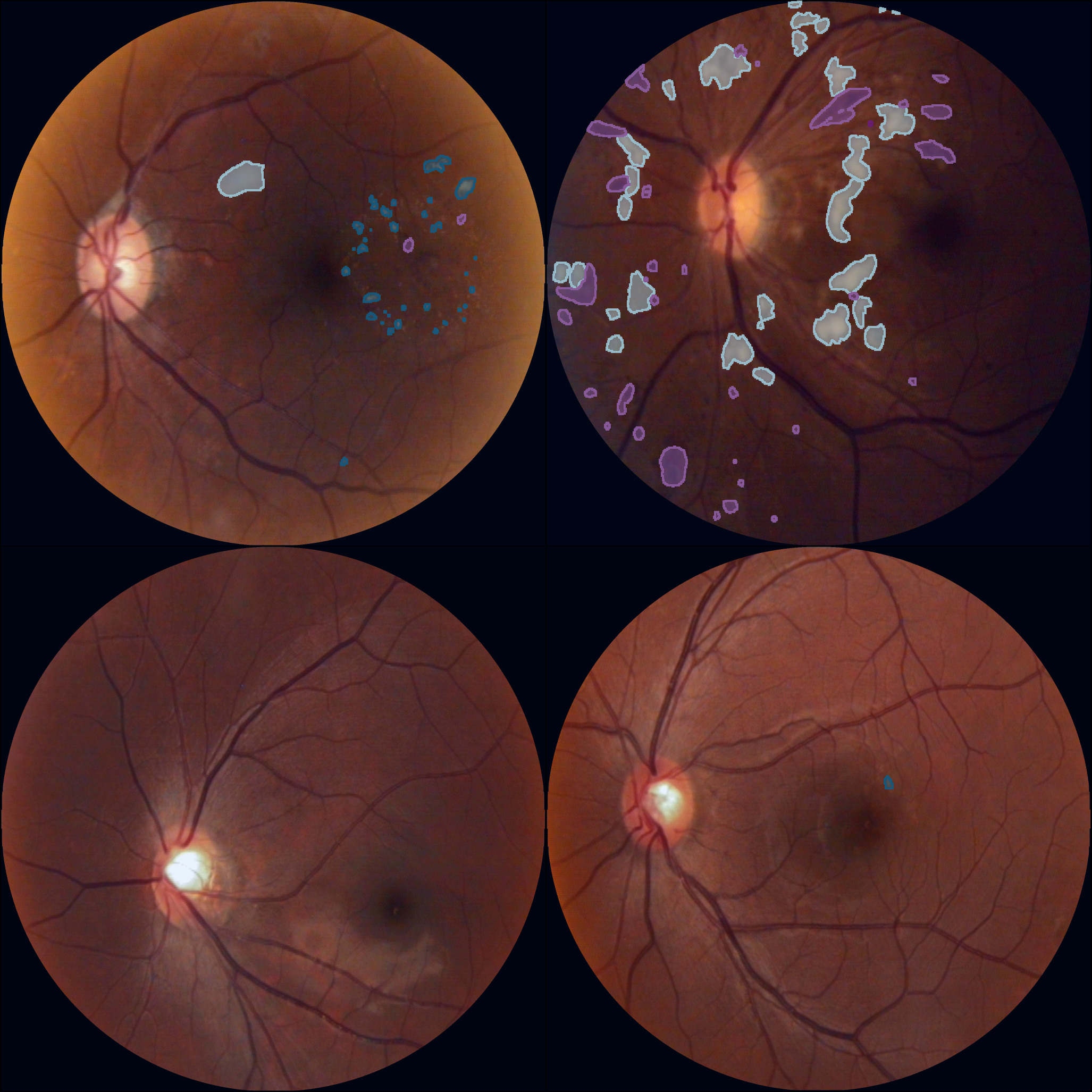}
    \end{subfigure}
    \begin{subfigure}{\colwidth}
        \subcaption[t]{$\mathcal{M}_\mathcal{S} \rightarrow R$}
        \label{fig-ProbeR}
        \includegraphics[trim={0 1027px 0 0},clip,width=\textwidth]{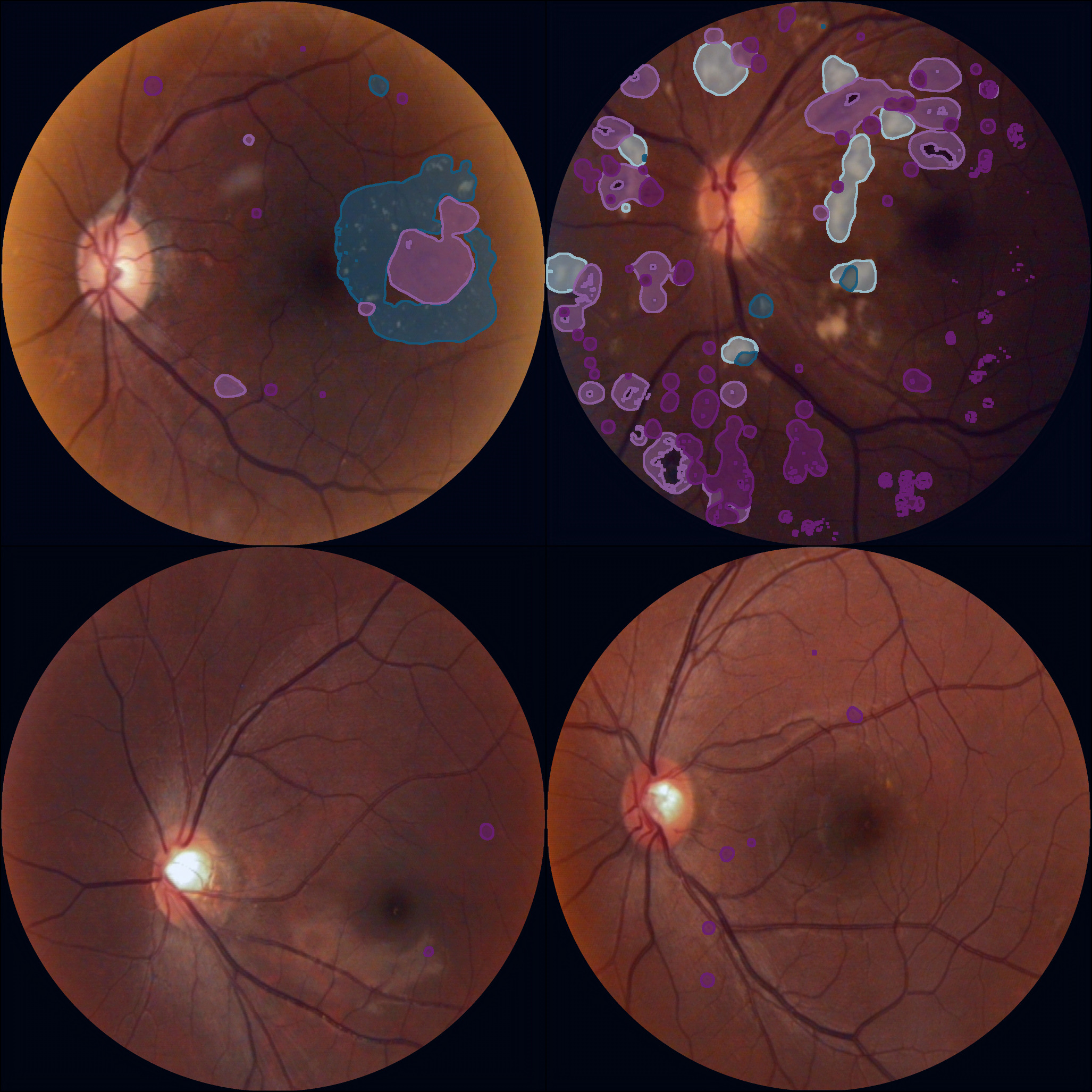}
    \end{subfigure}
    \begin{subfigure}{\colwidth}
        \subcaption[t]{$\mathcal{M}_\mathcal{S} \rightarrow D$}
        \label{fig-ProbeD}
        \includegraphics[trim={0 1027px 0 0},clip,width=\textwidth]{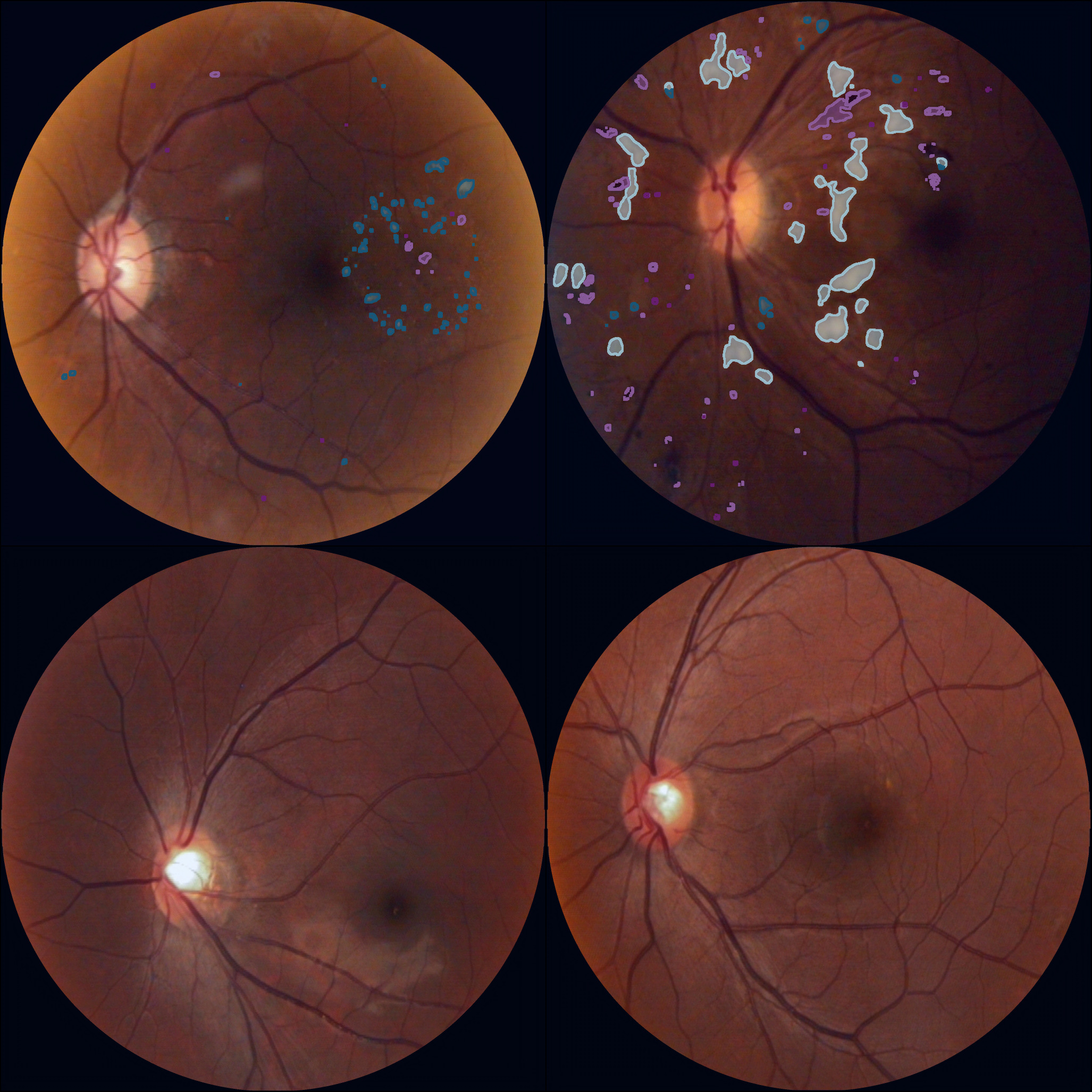}
    \end{subfigure}
     \begin{subfigure}{\colwidth}
        \subcaption[t]{$\mathcal{M}_\mathcal{S} \rightarrow I$}
        \label{fig-ProbeM}
        \includegraphics[trim={0 1027px 0 0},clip,width=\textwidth]{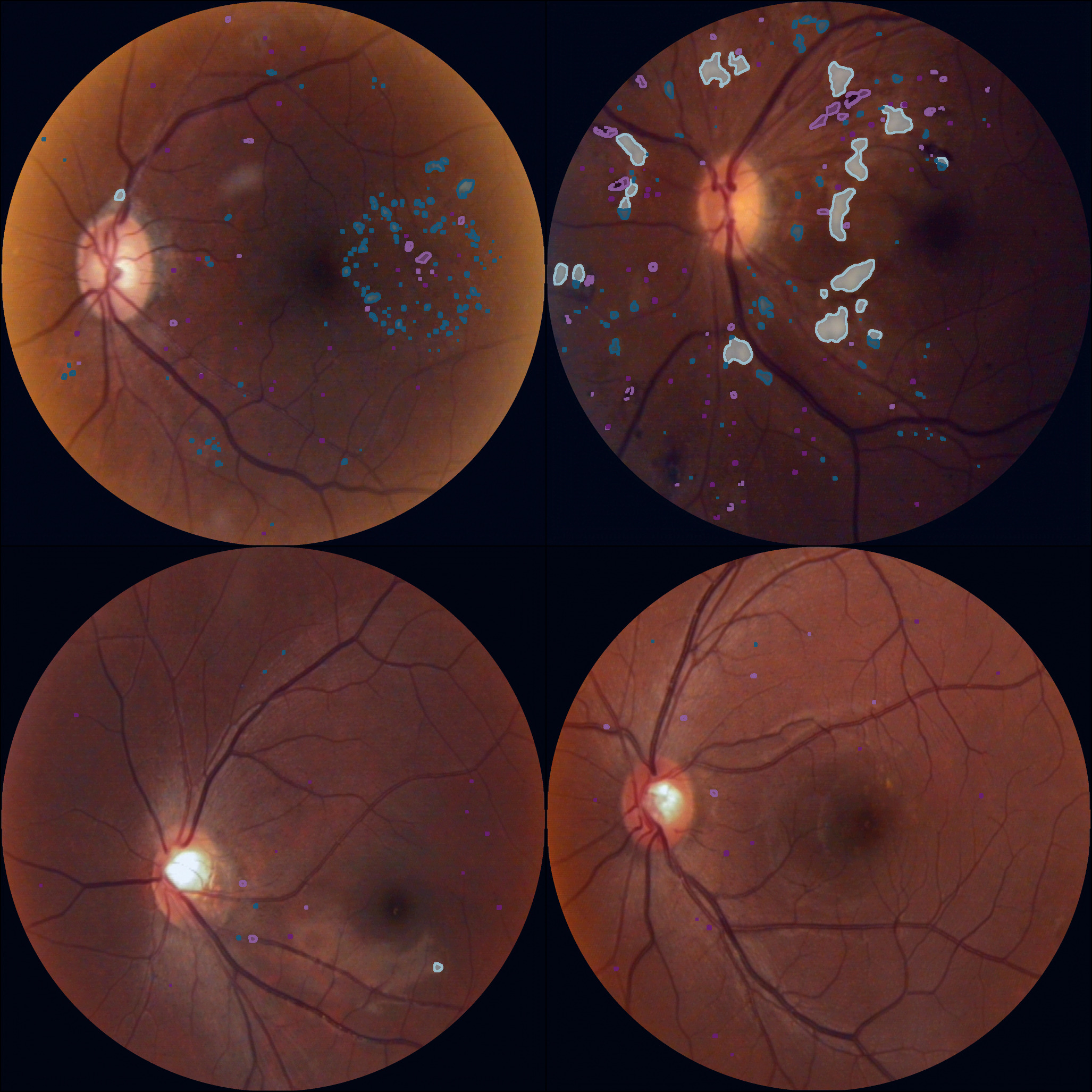}
    \end{subfigure}
    \begin{subfigure}{.32\columnwidth}
        \subcaption[t]{TJ-DR groundtruth}
        \label{fig-CloseUpGT}
        \includegraphics[trim={512px 1283px 1027px 256px},clip,width=\textwidth]{images/TJDR/StyleConvert/TJDR_GT}
    \end{subfigure}
    \begin{subfigure}{.32\columnwidth}
        \subcaption[t]{C-SNN(D)}
        \label{fig-CloseUpCSSN}
        \includegraphics[trim={512px 1283px 1027px 256px},clip,width=\textwidth]{images/TJDR/CSSN/from_TJDR_to_FundusDataset.DDR}
    \end{subfigure}
    \begin{subfigure}{.32\columnwidth}
        \subcaption[t]{$\mathcal{M}_\mathcal{S} \rightarrow D$}
        \label{fig-CloseUpOurs}
        \includegraphics[trim={512px 1283px 1027px 256px},clip,width=\textwidth]{images/TJDR/StyleConvert/from_TJDR_to_DDR}
    \end{subfigure}
    \caption{\review{Different segmentation maps obtained with the Conditional Stochastic Network (\ref{fig-CSNNR}, \ref{fig-CSNND}, \ref{fig-CSNNM}) and with our approach (\ref{fig-ProbeR}, \ref{fig-ProbeD}, \ref{fig-ProbeM}) on two images from the TJ-DR dataset. Columns 2 to 4: segmentations in the styles of RETLES (coarse), DDR (fine), and IDRID (fine, but less training data). Bottom row (\ref{fig-CloseUpGT}, \ref{fig-CloseUpCSSN} and \ref{fig-CloseUpOurs}): close-up on a group of exudates and hemorrhages on the temporal periphery of the macula.}}
    \label{fig:ComparingCSSNWithProbe}
\end{figure*}

\revisionAdd{Figure \ref{fig:ComparingCSSNWithProbe} provides examples of segmentations obtained either with \(\mathcal{M}_\mathcal{S}\) or with the C-SSN, for the same input images but with different style targets. Clearly, the C-SSN is able to measure the different style distributions conditioned to the set target, but its style conversion is never as faithful to the target as that achieved by our adversarial approach.} 

\revisionAdd{\subsection{Does adversarial conversion leads to semantic alteration?}}
\revisionAdd{By manipulating the input image through an adversarial attack, we succeed in deceiving the classification probe and thus altering the segmentation style of the dedicated network. This raises a legitimate question: what is the risk of altering the semantic content of the input image during the conversion? To verify the integrity of the image after conversion, we have implemented a set of constraints and validations:}

\begin{figure}[h]
    \centering
    \begin{subfigure}{.49\columnwidth}
        \includegraphics[width=\columnwidth]{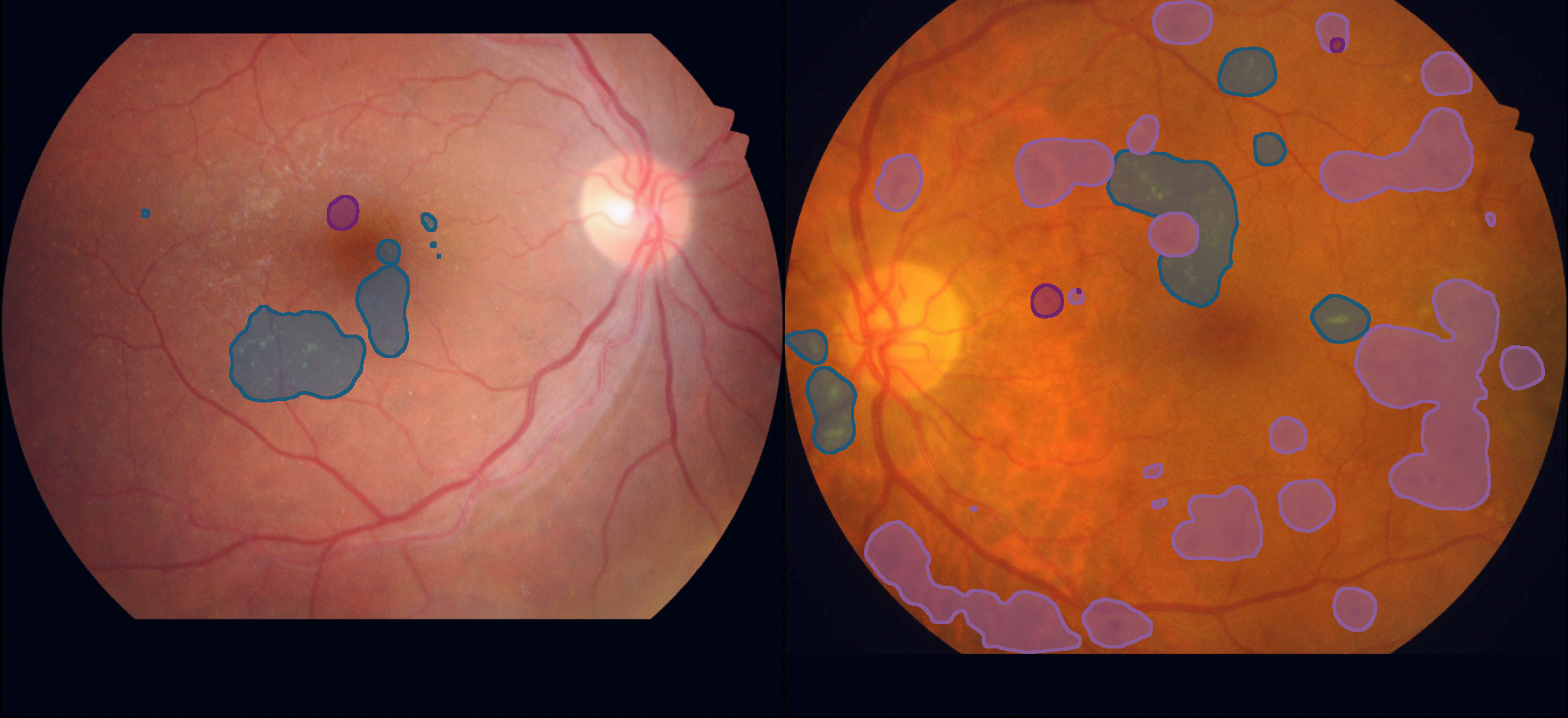}
        \subcaption[t]{$\Ms(\Btest{I} \rightarrow R)$}
    \end{subfigure}
    \begin{subfigure}{.49\columnwidth}
        \includegraphics[width=\columnwidth]{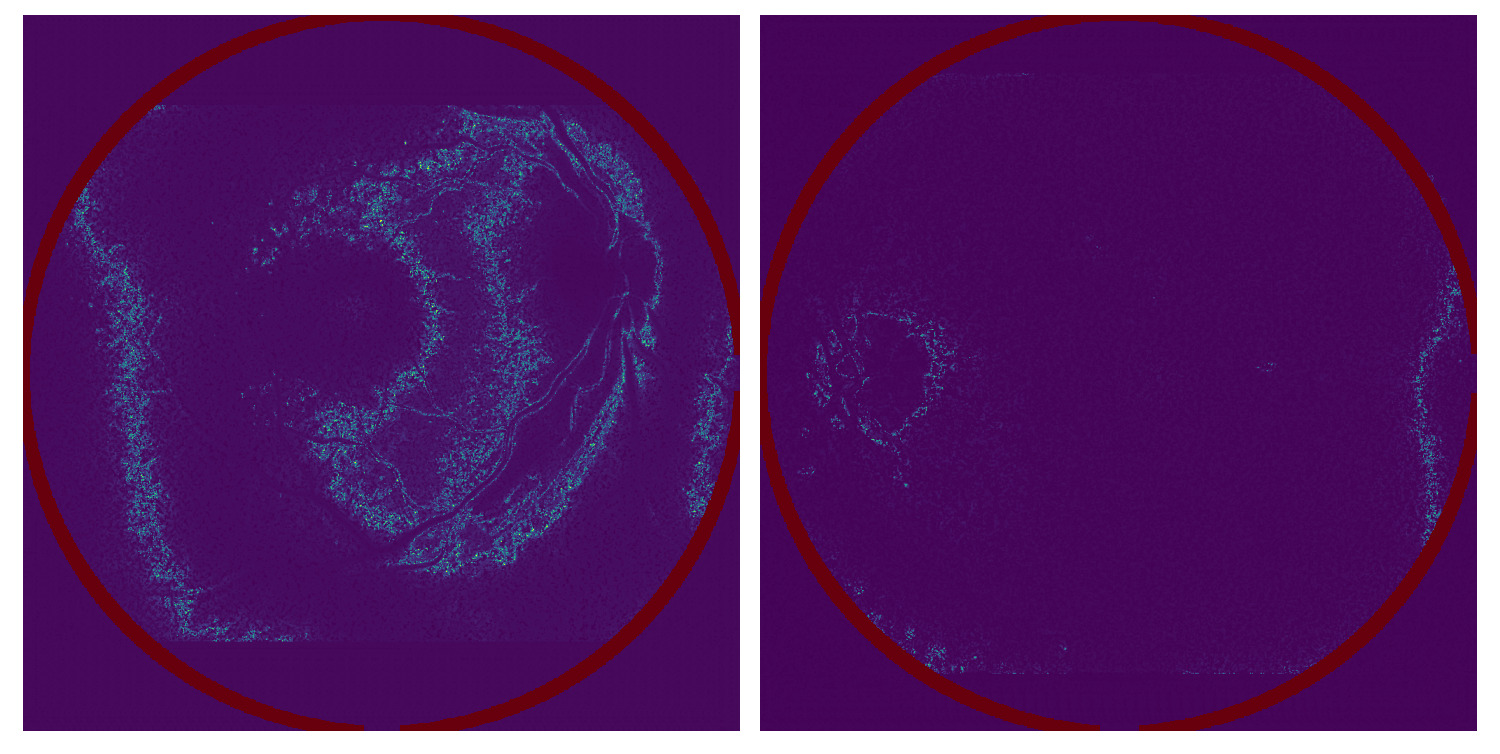}
        \subcaption[t]{$\mathcal{X}_{plotted}$}
    \end{subfigure}
    \caption{\review{Log-representation of the changes induced by the adversarial conversion on two sample images. For readability, the circular region of interest is overlaid in red on the log-residual plots.}}
    \label{fig:Xplotted}
\end{figure}

\revisionAdd{
\begin{enumerate}
    \item \textbf{Small Magnitude of Changes}: The modifications applied to the image were of minimal magnitude, carefully controlled to avoid altering the semantic information. We expressed the maximum modification $r$ of an image as a fraction of 255 (typically $\frac{5}{255}$, ensuring that the changes were at a level close to the acquisition quantification and imperceptible to human observers.
    \item \textbf{Visual Validation}: We visually inspected the original and style-converted images to confirm that there were no perceptible differences. This manual check was complemented by plotting the log-residual image, \(\mathcal{X}_{plotted}\), defined as:
    \begin{equation}
            \mathcal{X}_{plotted} = 10 \log_{10}(\frac{(x \rightarrow i)^2}{x^2})
    \end{equation}
    Figure \ref{fig:Xplotted} illustrates the result obtained.
    \item \textbf{Testing a classification model}. We trained a DR classification model (not segmentation-based) on independent databases (EyePACS + APTOS). We assessed that the grades remained unchanged before and after conversion, which should guarantee the semantic consistency of the images before/after conversion.
\end{enumerate}
}

\begin{figure}[h]
    \centering
    \includegraphics[width=.8\columnwidth]{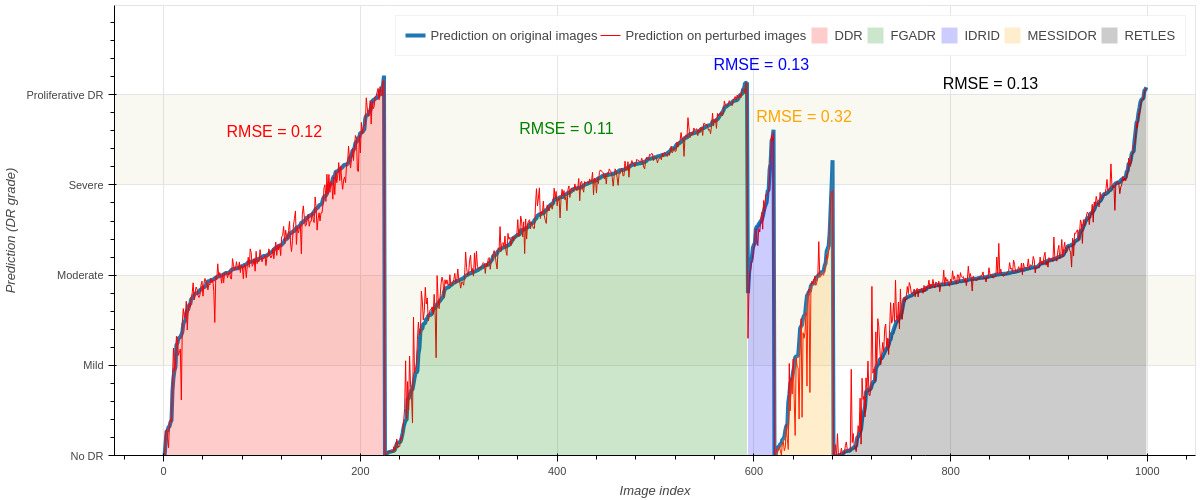}
    \caption{\review{Effect of adversarial perturbation on DR severity score predicted by a grading model trained in regression. The continuous score is measured on our five test datasets, using RETLES as the targeted style.}} 
    \label{fig:DR_Grading}
\end{figure}

\revisionAdd{Even if there is no difference to the human eye, this does not prove that the alteration maintains consistent semantic content for a neural network. Therefore, we added an experiment to validate the semantic integrity with regard to a proxy-CNN. We trained a ConvNext-Base \citep{liu2022convnet} to classify images according to the severity of diabetic retinopathy (DR), assigning classes “No DR”, “Moderate”, “Mild”, “Severe” and “Proliferative” to each image. To train the model, we combined two publicly available datasets: APTOS and EyePACS \citep{diabetic-retinopathy-detection}, for a total of 38,788 images.
To precisely quantify the effect of the image modification, the model was trained to perform regression toward the DR grade, offering the benefit of continuous prediction. This is a common practice as there is a natural ordering of the five classes. We ensured that the performance of the classification model aligned with the literature, suggesting that it was a good fit to classify our images before and after conversion. Any changes in this model’s predictions would indicate that an adversarial conversion added or removed important structures.
The continuous DR score before and after conversion for each image of the five databases is shown in Figure \ref{fig:DR_Grading}. The mean square error for each segmentation dataset varies in the range [0.11 - 0.32]. Given that a variation of 1 is needed to change the discrete diagnosis associated with an image, we conclude that the adversarial conversion does not significantly modify the semantic content of the image. Specifically, out of 1000 test images, 964 retained the same discrete grade. Upon inspection of the 36 remaining cases, the discrepancies were found where the predicted score was very close to the boundary between two discrete grades (e.g., 1.49, at the boundary between grade 1 and 2). 
}

\subsection{Continuous style-to-style interpolation}
Due to the nature of targeted adversarial attacks, our methodology only allows sampling among one of the five predefined styles, in a discrete form. We propose two simple ways to obtain continuous conversion using linear interpolation:

\begin{itemize}
	\item Building an interpolated loss in the probe's output space:
	\begin{equation}
		\mathcal{L}_{inter} = (1-\alpha) \cdot \mathcal{L}(y_{x}, i) + \alpha \cdot \mathcal{L}(y_{x}, j)
	\end{equation}
	\item Interpolation in the input space by mixing two conversions, where $x$ is typically an image from $\Btest{A}$ (or $\Btest{A} \rightarrow i$):
	\begin{equation}
		\label{eq:InputInterpolation}
		x_{inter} = (1-\alpha) \cdot x + \alpha \cdot (x \rightarrow j)
	\end{equation}
\end{itemize}

The former differs from the latter due to the non-linear nature of the Projected Gradients algorithm.
We found the second option to be more stable and to provide smoother results. Figure \ref{fig:ContinuousInterpolationStyle} illustrates the effect of the interpolation based on Equation \ref{eq:InputInterpolation}. The coefficient $\alpha$ can be sampled continuously to create a fairly smooth transition between two target annotation styles (an animation is included in the code repository). 

\def\colfigSizes{.19\textwidth}
\begin{figure*}[h]
	\centering
	\begin{subfigure}{\colfigSizes}
	\includegraphics[width=\columnwidth]{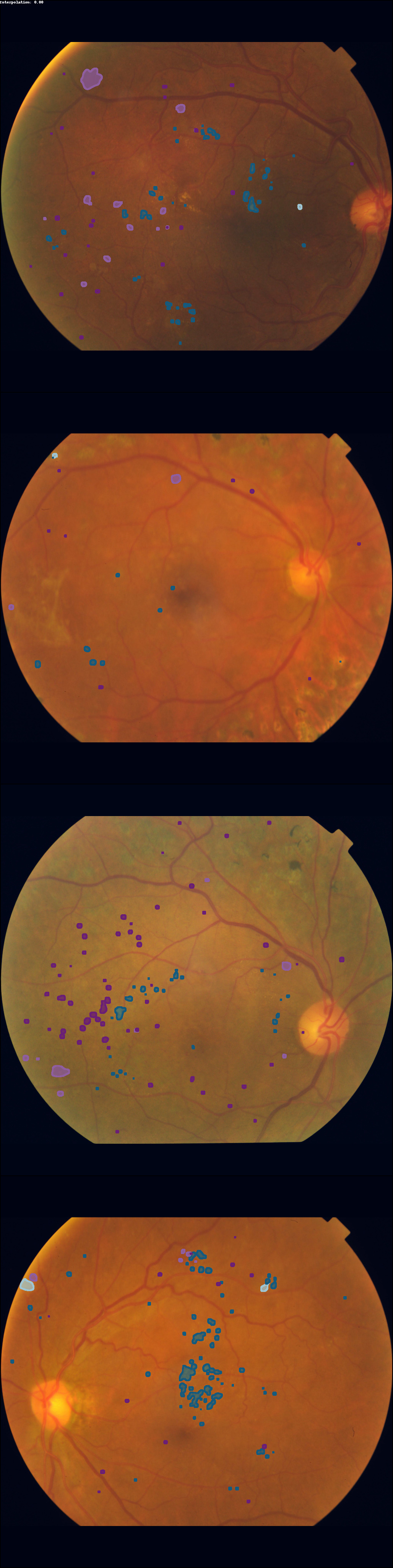}
		\subcaption{$\alpha=0$}
	\end{subfigure}
	\begin{subfigure}{\colfigSizes}
	\includegraphics[width=\columnwidth]{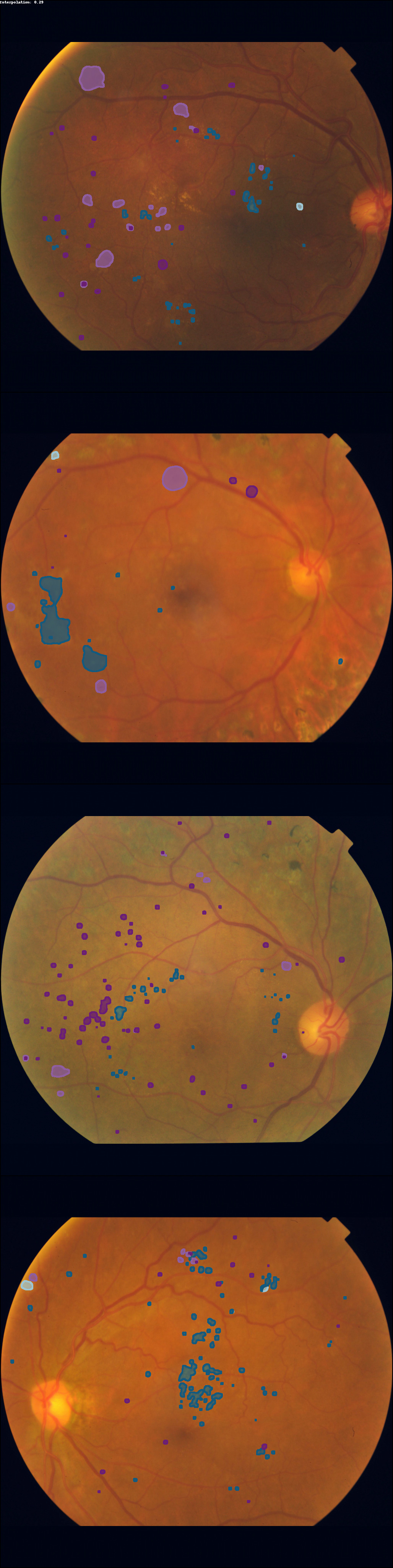}
		\subcaption{$\alpha=0.25$}
	\end{subfigure}
	\begin{subfigure}{\colfigSizes}
	\includegraphics[width=\columnwidth]{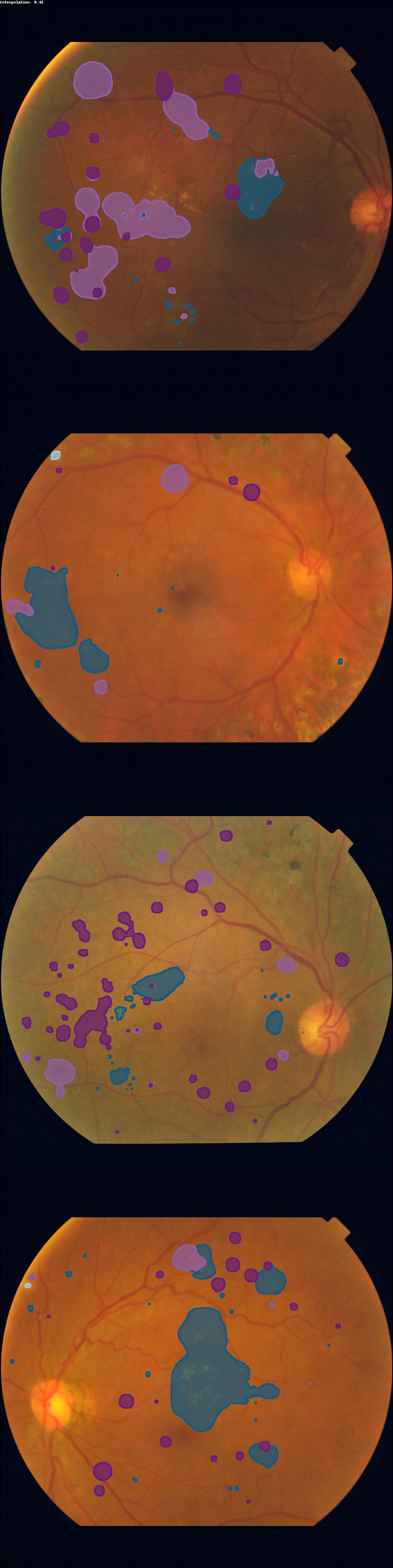}
		\subcaption{$\alpha=0.50$}
	\end{subfigure}
	\begin{subfigure}{\colfigSizes}
	\includegraphics[width=\columnwidth]{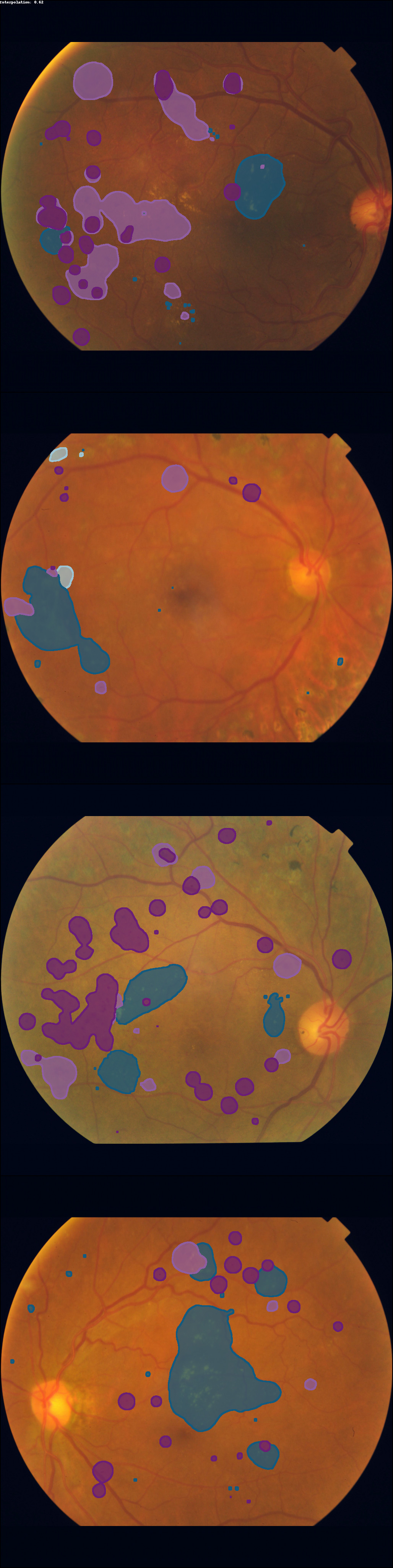}
		\subcaption{$\alpha=0.75$}
	\end{subfigure}
	\begin{subfigure}{\colfigSizes}
		\includegraphics[width=\columnwidth]{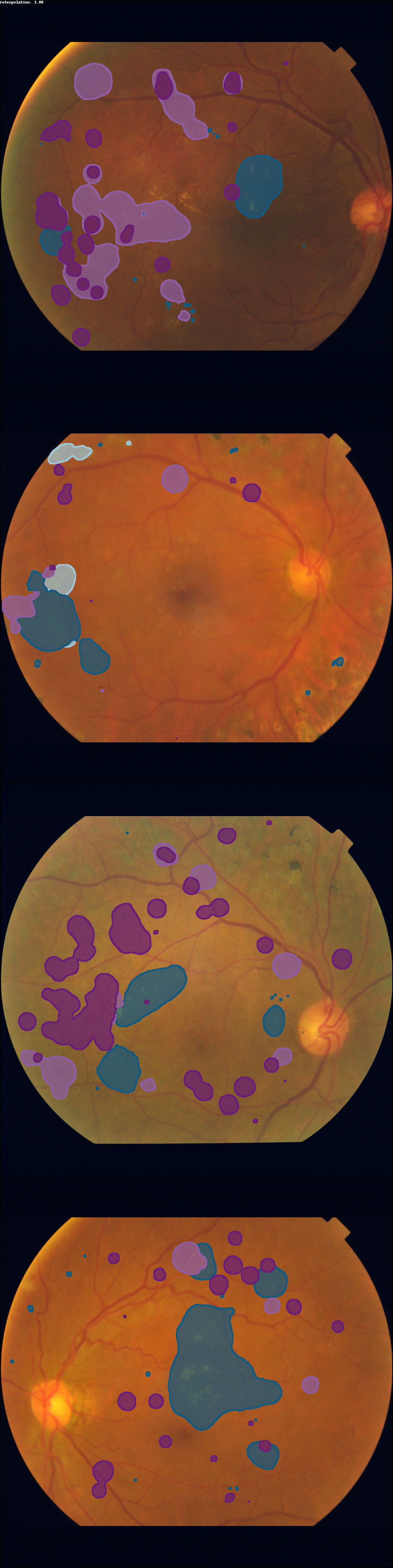}
		\subcaption{$\alpha=1.0$}
	\end{subfigure}
\caption{Continuous style conversion by linear interpolation in the input space, from fine-grained to coarse, i.e: $\Ms( (1-\alpha) \cdot \Btest{A} + \alpha \cdot (\Btest{A\rightarrow R})$. We illustrate the effect on four images sampled from the APTOS dataset (one per row). Each column corresponds to a step in the segmentation style transition.}
\label{fig:ContinuousInterpolationStyle}
\end{figure*}

\section{Applications}
\label{sec:applications}
In this section, we demonstrate \revisionAdd{three} possibles applications of our method. For the first one, we illustrate how style conversion can enhance the segmentation performance by homogenising the prediction of a model trained on low and high quality annotations. Furthermore, only a small subset of the labels need to be fined grained. 
Secondly, we \revisionAdd{demonstrate that style conversion can significantly improve the performance on external data}. Finally, we propose a method to generate uncertainty maps for a model's predicted segmentations by adapting the input space image modification used for style interpolation.

\subsection{Style distillation to improve performance \review{under unbalanced distribution}}

\begin{figure}
	\centering
	\includegraphics[width=.5\columnwidth]{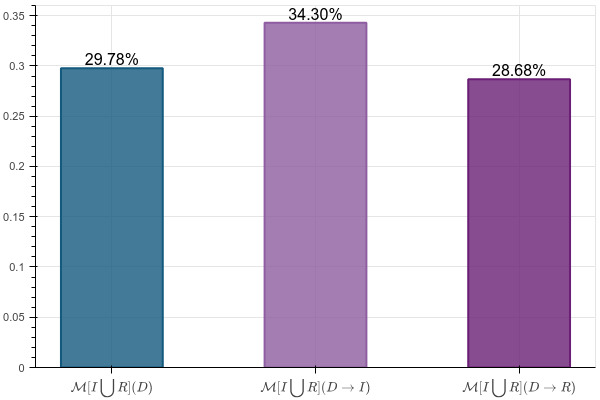}
	\caption{Performance (mIoU on $\Btest{D}$) of $\mathcal{M}[\Btrain{I} \bigcup \Btrain{R}]$ before and after conversion targeted toward $I$ or $R$ on the DDR test set.}
	\label{fig:PerformanceConversionLQHQ}
\end{figure}

In this setup, we trained a model $\mathcal{M}[\Btrain{I} \bigcup \Btrain{R}]$ using only two datasets: IDRiD and RETINAL-LESIONS. Arguably, the first one can be considered as the finest-grain dataset in term of annotations but is also the smallest with only 54 training images, whereas the second one is by far the coarsest but contains 1115 images. We tested $\mathcal{M}[\Btrain{I} \bigcup \Btrain{R}]$ on the DDR test, which has a style very visibly finer grained than RETINAL-LESIONS. 
We compared $\mathcal{M}[\Btrain{I} \bigcup \Btrain{R}](\Btest{D})$, $\mathcal{M}[\Btrain{I} \bigcup \Btrain{R}](\Btest{D} \rightarrow I)$ and $\mathcal{M}[\Btrain{I} \bigcup \Btrain{R}](\Btest{D} \rightarrow R)$. This required us to retrain a new two-class probe, but this operation only took 28 minutes on a RTX A6000. The conversion was done using interpolation in the input space as defined in Equation \ref{eq:InputInterpolation}; the parameters $\alpha, \epsilon, N$ and $r$ were adjusted qualitatively on a subset of the DDR validation set. The results are shown in Figure \ref{fig:PerformanceConversionLQHQ}; we observe an important performance gain on the DDR test set when taking IDRiD as the target style. 
Figure \ref{fig:QualitativeDDRHQLQRef} highlights the effectiveness of the conversion visually. Considering that only 4.8\% of the train set were finely labelled (the images from IDRiD), this demonstrates the ability to distillate a style even with a very limited amount of images corresponding to it. \revisionAdd{Conversely, as we can see in Figure \ref{fig:CoarseDefaultSeg}, without explicit conversion, the model segments in the style of the (vastly) predominant dataset (RETINAL-LESIONS). Yet, it still has learned IDRiD's style and can be biased toward it}. With a priori knowledge of the expected style of a given test set, we can boost the model's performance at inference time by matching the test set's style. In particular, we observe the following hierarchy:
$\Dsingle{\mathcal{M}[\Btrain{I} \bigcup \Btrain{R}](\Btest{D} \rightarrow I)} > \Dsingle{\mathcal{M}[\Btrain{I}](\Btest{D})} > \Dsingle{\mathcal{M}[\Btrain{I} \bigcup \Btrain{R}](\Btest{D})}$. In other words, adding more training data (even in large quantity) does not necessarily lead to an improved model ($\mathcal{M}[\Btrain{I} \bigcup \Btrain{R}]$ vs $\mathcal{M}[\Btrain{I}]$), mainly because of the style mismatch between the datasets but this effect can be alleviated with segmentation style conversion. \revisionAdd{In this case, we only need a small set of additional finely labelled training data to improve the model's performance.}

\def\colfigSizes{.32\textwidth}
\begin{figure*}[h]
	\centering
	\begin{subfigure}{\colfigSizes}
		\includegraphics[width=\columnwidth]{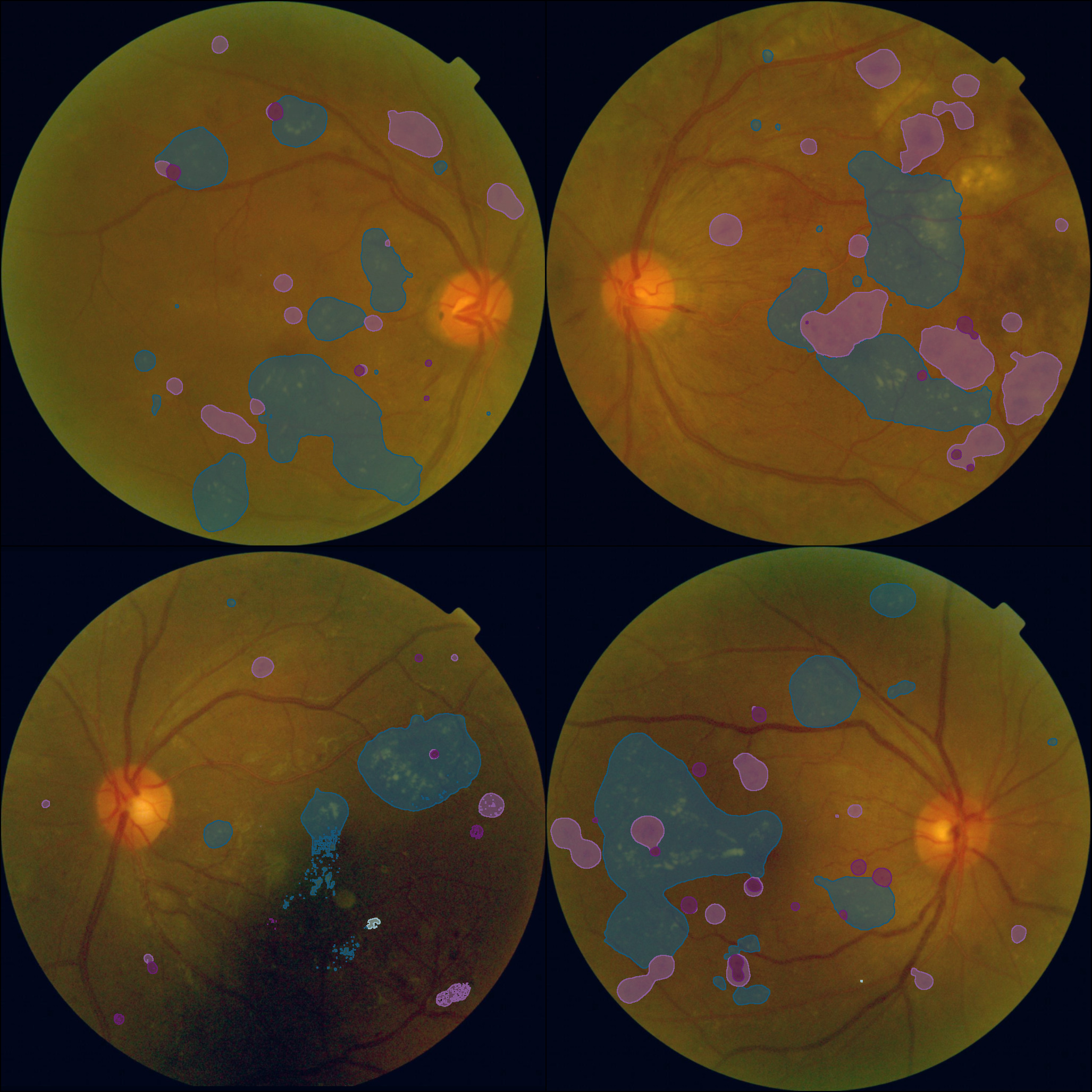}
		\subcaption{$\mathcal{M}[\Btrain{I} \bigcup \Btrain{R}](\Btest{D})$}
        \label{fig:CoarseDefaultSeg}
	\end{subfigure}
	\begin{subfigure}{\colfigSizes}
		\includegraphics[width=\columnwidth]{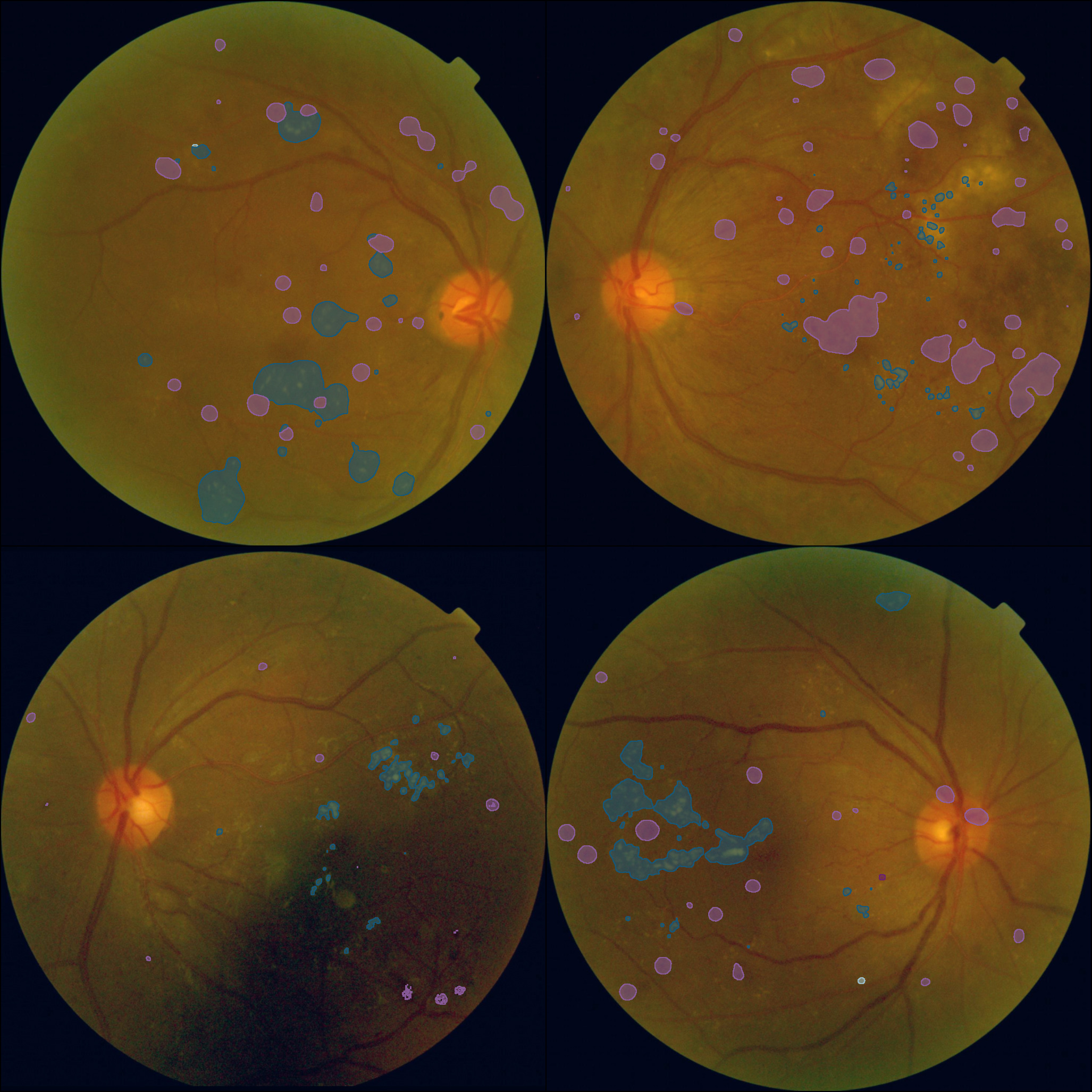}
		\subcaption{$\mathcal{M}[\Btrain{I} \bigcup \Btrain{R}(\Btest{D} \rightarrow I)$}
	\end{subfigure}
	\begin{subfigure}{\colfigSizes}
		\includegraphics[width=\columnwidth]{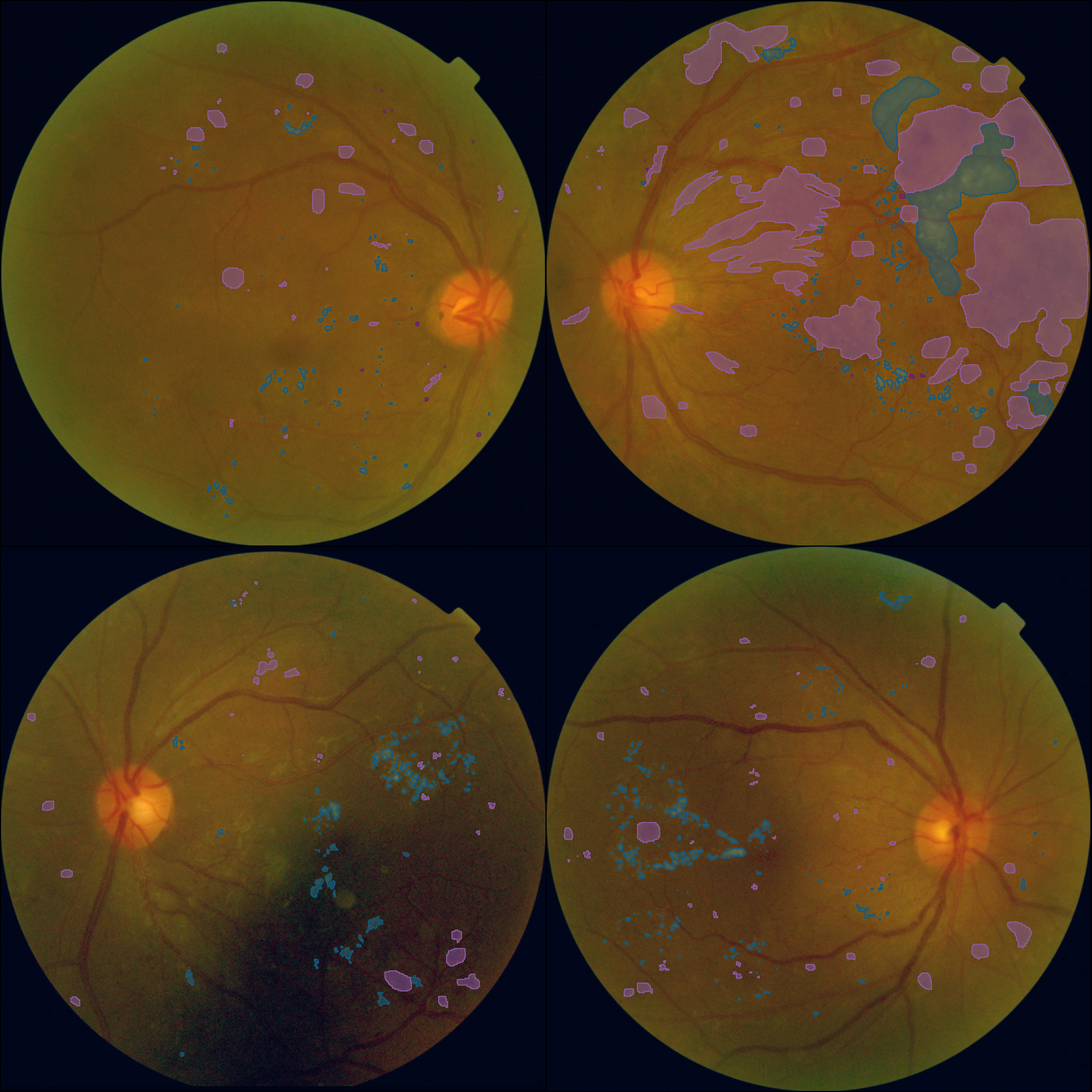}
		\subcaption{$\Btest{D}$ (Groundtruth)}
	\end{subfigure}
\caption{Adversarial conversion can be used to improve the model's performance by matching its prediction to an expected style that is different from the model's default one.}
\label{fig:QualitativeDDRHQLQRef}
\end{figure*}

\revisionAdd{\subsection{Performance improvement on external data}}
\revisionAdd{
Relying on the methodology from the previous section, we wanted to quantify the performance improvement we could achieve with our model \(\mathcal{M}_\mathcal{S}\) on external data (the TJ-DR database) using appropriate style conversion by lesion type. To do this, we first estimated the performance of the model on the TJ-DR training set, with and without conversion to the five targets available. The results are presented in Table \ref{tab:TJDR_validation_set}. 
\begin{table}[h]
    \centering
\begin{tabular}{lrrrrr}
\toprule
 & \multicolumn{5}{c}{\revisionAdd{AUC Prec/Recall curve}} \\
\revisionAdd{Model} &   \revisionAdd{CWS} & \revisionAdd{EX} & \revisionAdd{HEM} & \revisionAdd{MA} & \revisionAdd{Mean}\\
\midrule
\revisionAdd{$\mathcal{M}_{\mathcal{S}}$} & \revisionAdd{\textbf{0.515}} & \revisionAdd{0.427} & \revisionAdd{0.491} & \revisionAdd{0.252} & \revisionAdd{0.421} \\
\revisionAdd{$\mathcal{M}_{\mathcal{S}} \rightarrow I$} & \revisionAdd{0.379} & \revisionAdd{0.536} & \revisionAdd{0.430} & \revisionAdd{0.235} & \revisionAdd{0.395} \\
\revisionAdd{$\mathcal{M}_{\mathcal{S}} \rightarrow R$} & \revisionAdd{0.272} & \revisionAdd{0.343} & \revisionAdd{0.456} & \revisionAdd{\textbf{0.284}} & \revisionAdd{0.339} \\
\revisionAdd{$\mathcal{M}_{\mathcal{S}} \rightarrow D$} & \revisionAdd{0.427} & \revisionAdd{\textbf{0.537}} & \revisionAdd{0.276} & \revisionAdd{0.236} & \revisionAdd{0.369} \\
\revisionAdd{$\mathcal{M}_{\mathcal{S}} \rightarrow M$} & \revisionAdd{0.459} & \revisionAdd{0.506} & \revisionAdd{\textbf{ 0.515}} & \revisionAdd{0.269} & \revisionAdd{\textbf{0.437}} \\
\revisionAdd{$\mathcal{M}_{\mathcal{S}} \rightarrow F$} & \revisionAdd{0.316} & \revisionAdd{0.369} & \revisionAdd{0.470} & \revisionAdd{0.262} & \revisionAdd{0.354} \\
\bottomrule
\end{tabular}
    \caption{\review{Performance on the TJ-DR validation set using targeted style conversion by lesion type.}} 
    \label{tab:TJDR_validation_set}
\end{table}
}
\revisionAdd{We see that for the segmentation of cotton wool spots, the model without conversion ($\mathcal{M}_\mathcal{S}$) performs best. However, for the segmentation of other lesions, conversion is appropriate: to DDR for exudates, to MESSIDOR for hemorrhages and to RETINAL-LESIONS for microaneurysms. With these conclusions, we applied these conversions to the TJ-DR test set. The results are reported in Table \ref{tab:TJDR_test_set}. For comparison purposes, we include in Table \ref{tab:TJDR_test_set} the performances obtained without conversion as well as those of different specialist models. These new results demonstrate the clear advantage of converting the inference images toward a target style (depending on the lesion type to detect). }
\begin{table}[h]
    \centering
\begin{tabular}{lrrrrr}
\toprule
 & \multicolumn{5}{c}{\revisionAdd{AUC Prec/Recall curve}} \\
\revisionAdd{Model} &   \revisionAdd{CWS} & \revisionAdd{EX} & \revisionAdd{HEM} & \revisionAdd{MA} & \revisionAdd{Mean} \\
\midrule
\revisionAdd{Mix:} & \revisionAdd{$\mathcal{M}_{\mathcal{S}}$} & \revisionAdd{$ \rightarrow D$} & \revisionAdd{$ \rightarrow M$} & \revisionAdd{$ \rightarrow R$} \\
& \revisionAdd{0.427} & \revisionAdd{\textbf{0.545}} & \revisionAdd{\textbf{0.512}} & \revisionAdd{0.265} & \revisionAdd{\textbf{0.437}} \\
\midrule
\multicolumn{6}{c}{\revisionAdd{Baseline models}} \\
\midrule
\revisionAdd{$\mathcal{M}_{\mathcal{S}}$} & \revisionAdd{0.427} & \revisionAdd{0.381} & \revisionAdd{0.468} & \revisionAdd{0.218} & \revisionAdd{0.374} \\
\revisionAdd{$\mathcal{M}_{R}$} & \revisionAdd{0.379} & \revisionAdd{0.317} & \revisionAdd{0.379} & \revisionAdd{0.193} & \revisionAdd{0.317} \\
\revisionAdd{$\mathcal{M}_{M}$} & \revisionAdd{0.245} & \revisionAdd{0.323} & \revisionAdd{0.464} & \revisionAdd{\textbf{0.322}} & \revisionAdd{0.338} \\
\revisionAdd{$\mathcal{M}_{I}$} & \revisionAdd{0.427} & \revisionAdd{0.327} & \revisionAdd{0.434} & \revisionAdd{0.226} & \revisionAdd{0.353} \\
\revisionAdd{$\mathcal{M}_{F}$} & \revisionAdd{\textbf{0.454}} & \revisionAdd{0.381} & \revisionAdd{0.511} & \revisionAdd{0.066} & \revisionAdd{0.353} \\
\revisionAdd{$\mathcal{M}_{D}$} & \revisionAdd{0.285} & \revisionAdd{0.341} & \revisionAdd{0.426} & \revisionAdd{0.215} & \revisionAdd{0.317} \\
\bottomrule
\end{tabular}
\caption{\review{Performance on the  TJ-DR test set using targeted style conversion by lesion type. }} 
\label{tab:TJDR_test_set}
\end{table}

\subsection{Uncertainty estimation}
\begin{figure}
    \centering
    \includegraphics[width=\columnwidth]{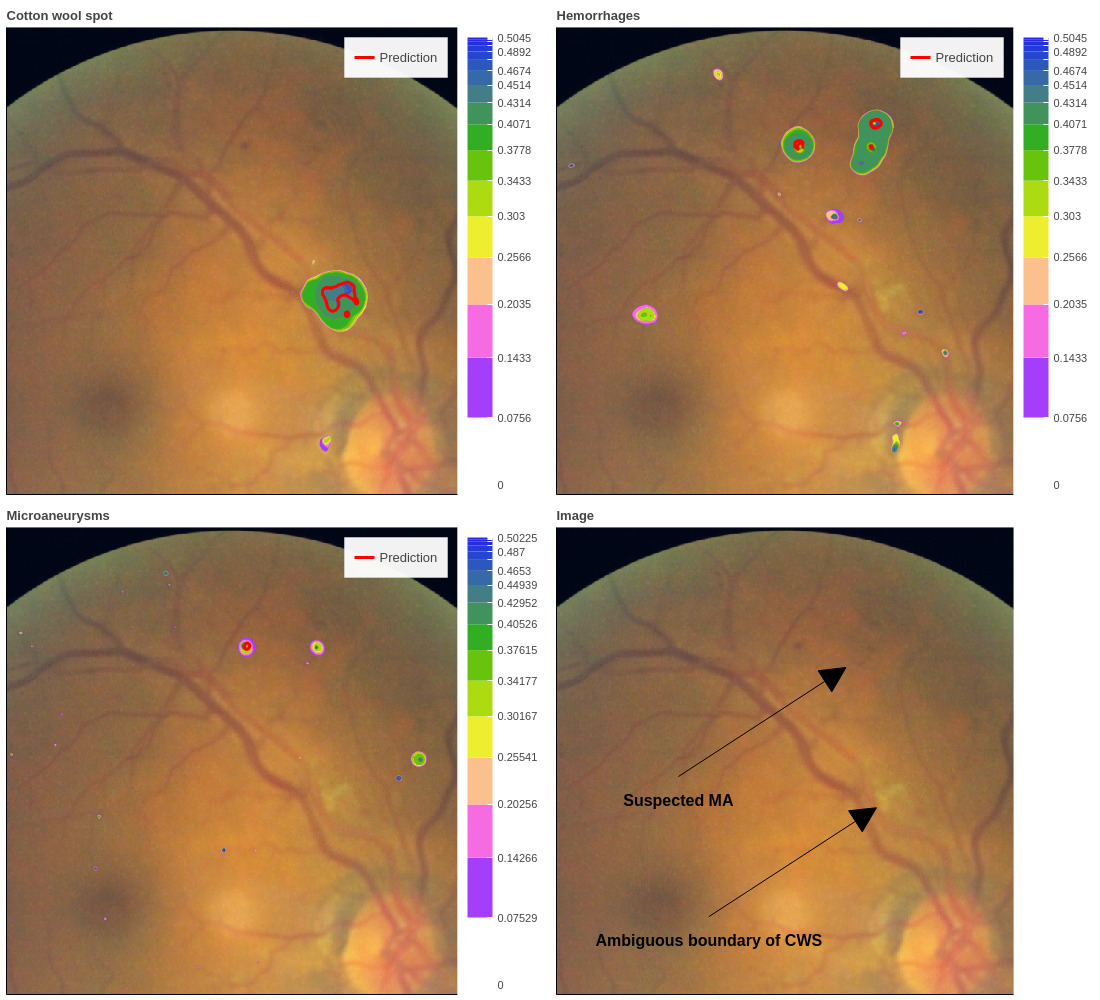}
    \caption{\review{Estimated uncertainty maps highlighting ambiguous areas of an image for each type of lesion (CWS, HEM, MA). Note that the ambiguities are not just related to predicted lesion boundaries. In the bottom right panel, the top arrow points to a suspected micraoneurysm that was not identified by the segmentation model (small uncertainty at same location in bottom left panel).}}
    \label{fig:uncertainty}
\end{figure}

Estimating the uncertainty of a model's predictions is useful to gain a better understanding of its internal behaviour. Inspired by the work of \cite{garifullinDeepBayesianBaseline2021}, we propose an estimation of the model's aleatoric uncertainty using a local perturbation-based approach. The idea is to sample $N_A$ points in the image's neighbourhood and use the predicted samples to calculate a predictive mean and standard deviation across the distribution. The sampling process reformulates Equation \ref{eq:InputInterpolation} as:

\begin{equation}
	x_{\alpha} = (1-\alpha) \cdot x + \alpha \cdot (x \rightarrow j) \text{ with } \alpha \sim \mathcal{U}(0, 1)
\end{equation}
The aleatoric uncertainty map $U_A$ is then obtained as:
\begin{equation}
	U_A = \sigma_{\alpha} (\MS(x_\alpha))
\end{equation}
where $\sigma_{\alpha}$ denotes the standard deviation taken across the $N_A$ points. In general, the computed uncertainty ($\sigma$) is large in the neighbourhoods around the lesions, which can be interpreted as revealing the different styles learned by the network, \revisionAdd{but also as highlighting the ambiguous nature of some lesions' boundaries.}
On the other hand, it can also highlight areas corresponding to potential false negatives, particularly in the case of microaneurysms. \revisionAdd{Both situations can be seen in Figure \ref{fig:uncertainty}}.

\section{Discussion}
\label{sec:discussion}

Our work has highlighted the concept of style adoption by a model throughout its training trajectory, contingent upon the chosen dataset. By combining several of these datasets, each characterized by a distinct annotation style, the model acquires the capacity to selectively adopt a style at inference time based on the input image. This suggests that it is able to trace back the origin of an image to an implicit latent variable. We demonstrated the robustness of this ability to various forms of simple perturbations. This in turn motivated our choice to train a linear identification probe based on the features computed by the segmentation model's encoder.

This probe can subsequently be subjected to manipulation through adversarial attacks, allowing a subtle alteration of the input image 
to deceive both the probe and the model. We highlighted that this perturbation also affects the segmentation model. Through a series of experiments, we illustrated the potential of this framework to sample multiple segmentations reflecting different styles, and even to interpolate continuously among them, all for a single image. Our approach has the distinct advantage of not necessitating any alteration to the model and is amenable to implementation within a conventional architecture. It only requires to train an external model (the probe), which is not resource intensive. 

\subsection{Limitations and future work}
We acknowledge several limitations of this work, which would warrant further investigation:

\begin{itemize}
	\item By assumption, we equate the notion of annotation style with that of the original database. This assumption is justified by our experience that annotation style significantly depends on the protocol and tools provided to annotators. In practice, however, there will be a certain variability among annotators even within the same database. Lacking information about individual annotators, we are compelled to assume a degree of homogeneity in annotation style within a given database. Access to annotator-specific information per image rather than per database could potentially yield a finer style conversion.
 
	\item Our style conversion is achieved through adversarial attacks, i.e., by backpropagation of gradients towards a perturbation that leads to the desired target. Deliberately, we minimize the magnitude of this perturbation, with the idea that it should not induce hallucinations of features akin to what certain GANs might produce. While this notion seems crucial in a clinical context, it complicates the hypothetical deployment of our method, as the information of added perturbation to the image is generally not storable in 8-bits (and thus in most conventional image storage formats). 
 
	\item From a clinical and diagnostic perspective, the usefulness of precise lesion segmentation (multi-styled or not), as opposed to merely detecting their presence, remains to be demonstrated. In this regard, the detailed shape of the segmented lesions (i.e. labelling style) might seem secondary.
    We argue that incorporating segmentation maps into future models should enhance our understanding of their functioning and potentially extend their applicability to other modalities, such as wide-field fundus imaging.
\end{itemize}

\section{Conclusion}
\label{sec:conclusion}

This work provides an approach for training with multiple databases despite their diverse annotation styles. Indeed, we highlight 
the substantial qualitative gain achieved through data combination. However, in adopting this approach, there is uncertainty regarding the annotation style the model will adopt during inference. Our methodology addresses this uncertainty by compelling the trained model to behave as if a new image belongs to a database with a known associated style. This principle, which we term adversarial style conversion, opens the door to several applications:
\begin{itemize}
	\item Model training can proceed conventionally, even on heterogeneous data, given that its behavior can be guaranteed a posteriori to match a known style.
	\item By training a model on multiple databases, its generalization capabilities improve, thereby offering an avenue for leveraging a larger quantity of data.
	\item Through the continuous interpolation principle between two styles, it becomes possible to generate different segmentation hypotheses. Given the substantial variability among annotators in the recognition of retinal lesions, this capability can be utilized to obtain an uncertainty estimate through Monte Carlo sampling of multiple segmentation hypotheses. However, we defer its comparison to other existing methods to future research endeavors.
\end{itemize}

We limited our experiments to fundus images and retinal lesion segmentation, the latter being our field of interest. \revisionAdd{In future work, we will explore different variants of our methodology and its generalization to multimodal domain adaptation, in particular from Ultra Wide Field images to regular fundus ones. Although our research is focused on retinal images,} we emphasise that our technique could have applications well beyond this area. The issue of  segmentation style, and in particular the combination of coarse labels and fine style distillation, has a large number of applications. Given the conceptual simplicity of our methodology, we encourage practitioners to experiment with it in others areas. 
\section*{Program Availability}

The code, trained models and the logs of the experiments will be made available from our GitHub repository: \revisionAdd{\url{https://github.com/ClementPla/MultiStyle_FundusLesionSegmentation/}}. 
\revisionAdd{To favor reproducibility of our results and to encourage further research, we have released a library standardizing the loading, preprocessing, data augmentation and train/val/test splitting of data from different fundus databases:}
\revisionAdd{\url{https://github.com/ClementPla/fundus-data-toolkit/}}.
\revisionAdd{We also provide an easy way for non-developers to use the models described in Table \ref{tab:perfComparéeSegmentation}: \url{https://github.com/ClementPla/fundus-lesions-toolkit/}}.

\section*{Declaration of Competing Interest}
The authors declare that they have no known competing financial interests or personal relationships that could have influenced the work reported in this paper. 

\section*{Acknowledgments}

\revisionAdd{We sincerely acknowledge Dr. Fares Antaki and Philippe Debann\'e  for revising this manuscript.
We also thank the Ophthalmology Research Fund from Université de Montr\'eal (FROUM), Diabetes Action Canada, the Line Chevrette Foundation and the Natural Sciences and Engineering Research Council (NSERC) for funding this project.}

\bibliographystyle{natdin}
\bibliography{main}

\end{document}